\documentclass[journal,final,twoside]{IEEEtran}

\usepackage{graphicx,algorithm,algorithmic}
\usepackage[latin1]{inputenc}
\usepackage[T1]{fontenc}
\usepackage{subfigure}
\usepackage{multirow}
\usepackage{slashbox}
\usepackage{rotating}
\usepackage{latexsym}
\usepackage{color}
\usepackage{framed}
\usepackage{amssymb}
\usepackage{amsmath}
\usepackage{float}
\usepackage{booktabs}
\usepackage{url}
\usepackage{balance}

\newcommand{\MeshAlg}{TransforMesh}

\newcommand{\CorrectMesh}{2-D compact oriented manifold}

\begin{document}

\title{Topology-Adaptive Mesh Deformation\\ for Surface Evolution,
  Morphing,\\ and Multi-View Reconstruction
  \thanks{This research was supported by the European Union's
Marie Curie program through the Research Training Network VISIONTRAIN.}}

\author{Andrei~Zaharescu,
	Edmond~Boyer,
	and Radu~Horaud
\thanks{Manuscript received; revised -.}
\thanks{Author's Affiliation: INRIA Grenoble Rh\^{o}ne-Alpes, 655 avenue de l'Europe, 38330 Montbonnot Saint-Martin, FRANCE}
\thanks{Emails: firstname.lastname@inrialpes.fr}
}


\maketitle

\begin{abstract}

Triangulated meshes have become ubiquitous discrete-surface
representations. In this paper we address the problem of how to
maintain the manifold properties of a surface while it undergoes
strong deformations that may cause topological changes.
We introduce a new self-intersection
removal algorithm, \MeshAlg{}, and we propose a mesh
evolution framework based on this algorithm. Numerous shape modelling
applications use surface evolution in order to improve shape
properties, such as appearance or accuracy. Both explicit and implicit
representations can be considered for that purpose. However, explicit
mesh representations, while allowing  for accurate surface modelling,
suffer from the inherent difficulty of  reliably dealing with
self-intersections and  topological changes such as merges and splits.  As a consequence, a
majority of methods rely on implicit  representations of surfaces,
e.g. level-sets, that naturally overcome these
issues. Nevertheless, these methods are based on volumetric
discretizations, which introduce an unwanted precision-complexity
trade-off. The method that we propose
handles  topological changes in a robust manner and removes self
intersections, thus overcoming the traditional limitations of mesh-based
approaches. To illustrate the effectiveness of \MeshAlg{}, we describe
two challenging applications, namely surface morphing
and 3-D reconstruction. 


\end{abstract}

\begin{keywords}
Surface, manifold, triangulated mesh, surface evolution, deformable objects, morphing, 3-D reconstruction.
\end{keywords}

\section{Introduction}
\PARstart{I}{n} the process of modeling shapes, several applications
resort to surface evolution to improve shape properties.  For
instance, shape surfaces are evolved so that their appearances are
improved, as when smoothing shapes, or so that they best explain given
observations as in image based modeling. The interest arises in
several fields related to shape modeling: computer vision, computer graphics, medical imaging and visualization among others. Surface
evolution is usually formulated as  an optimization process that seeks
for a surface with a minimum energy with respect to the desired
properties. To this aim, surfaces can be represented in different
ways, from implicit to explicit representations, and deformed in an
iterative way during optimization. Polygonal meshes, while being  one
of the most widely used representation when modeling shapes, are
seldom considered in such evolution schemes. The main reason for that is
the inherent difficulty to handle topological changes and
self-intersections that can occur during evolution.  

In this paper,  we introduce  an intuitive and efficient algorithm, named TransforMesh, that performs self-intersection removal of triangular meshes, allowing for topological changes, e.g. splits and merges. The method assumes as input a proper oriented mesh -- a 
2-D compact oriented manifold -- which experienced any connectivity
preserving deformation. It  computes the {\it outside surface} of the
deformed mesh. To illustrate the approach and its interests, we
propose a generic surface evolution framework based on TransforMesh
and we present two applications: mesh morphing and 
variational multi-view 3-D reconstruction.

\subsection{Literature Review}
As a result of the large interest for surface evolution in many application domains, numerous surface deformation schemes have been proposed over the last decades. They roughly fall into two main categories with respect to the representation which is considered for surfaces: Eulerian or Lagrangian.

{\bf Eulerian methods} formulate the evolution problem as time
variation  over sampled spaces, most typically fixed grids. In such a
formulation, the surface, also called the interface, is implicitly
represented. One of the most successful methods in this category, the
\textit{level  set method} \cite{Osher:2003fk,Osher:1988uq},
represents the interface as the zero level of a higher dimensional
function. A typical function used is the signed distance of the
explicit surface, discretized over the volume. At each iteration the
whole implicit function is moved. The explicit surface is recovered by
finding the 0-level set of the implicit function. A number of methods
have been proposed to extract surfaces from volumetric data
\cite{Lorensen:1987,Kobbelt:2001vl,Ju:2002ad,Ohtake:2003zp}. Such an
embedding  within an implicit function allows to automatically handle
topology changes, e.g. merges and/or splits. In addition, such methods
allow for an easy computation of geometric properties such as
curvatures and benefit from {\it viscosity solutions} -  robust
numerical schemes to deal with the evolution. These advantages explain
the popularity of level set methods in computer vision
\cite{Osher:2003kb} as well as in other fields, such as computational
fluid dynamics \cite{Sussman:1994} and computer animations of fluids \cite{Enright:2002kl}.
Nevertheless, implicit representations exhibit limitations resulting
from the grid discretization. In particular,  the precision/complexity
trade-off inherent to the grid has a significant impact on the
computational efficiency and the proposed  narrow-band solutions
\cite{Adalsteinsson:1995} or octree based implementations
\cite{Losasso:2006qy} only partially overcome this issue. In addition,
as shown by Enright {\it et al.} \cite{Enright:2002uq}, the level set method
is strongly affected by mass loss, smearing of high curvature regions
and by the inability to resolve very thin parts. Another limitation is
that level set methods are not appropriate for tracking surface
properties, such as color or texture, which can be
desirable in many image-based approaches (i.e. motion tracking). Thus,
while providing a solution for the intersection and topological issues
within surfaces, implicit representations introduce a new set of
issues for which careful solutions need to be crafted. 

{\bf Lagrangian methods} propose an approach where surfaces have
explicit representations which are  deformed  over time. Such
representations, meshes for instance,  present  numerous advantages,
among which adaptive  resolution and compact representation, as well
as the ability to directly handle non-geometric properties over the
surface, e.g. textures, without the necessity to reconstruct the
interface. On the other hand, they raise two major issues  when
evolved over time, namely self-intersections and topology changes,
which make them difficult to use in many practical scenarios. This is
why non-intersections and fixed topology were explicitly
enforced \cite{Park:2001,Hernandez:2004}. As a consequence, and in
spite of their advantages, they have often been neglected in favor of
implicit representations which provide practical solutions to such
issues. Nevertheless, solutions have been
proposed. McInerney and Terzopoulos \cite{McInerney:2000uq} introduced
topology adaptive deformable  curves and meshes, called T-snakes and
T-surfaces. However, in solving  the intersection problem, the authors
use a spatial grid, thus  imposing a fixed spatial resolution. In
addition, only offsetting motions, i.e. inflating or deflating, are
allowed. Another heuristic method  was proposed by Lauchaud et
al. \cite{Lachaud:2003kx} for mesh deformations.  Merges and splits
are performed in near boundary cases: when two surface boundaries are
closer  than a threshold and  facing each other, an artificial merge
is  introduced; a similar procedure is applied for a split, when the
two  surface boundaries are back to back. Self-intersections are  {\it
  avoided} in practice by imposing a fixed edge size. A similar method
was also proposed by Duan {\it et al.}~\cite{Duan:2004lr}.
Another extension is proposed by Brochu {\it et al.}~\cite{Brochu:2009}, with a greater focus on the the mesh optimization technique and guarantees. Alternatively,
Pons and Boissonat~\cite{Pons:2007b}  proposed a mesh approach  based on a
restricted 3-D Delaunay triangulation. A deformed mesh is obtained by
triangulating the moved vertices and assuming that the tetrahedra
categorization, i.e. inside and outside, remains after the
deformation. While being a robust and elegant solution, it
nevertheless relies on the assumption that the input mesh is
sufficiently dense such that the Delaunay triangulation will not
considerably change its layout. 

The methods proposed by Aftosmis {\it et al.}~\cite{Aftosmis:1997} and
Jung {\it et al.}~\cite{Wonnhyung-Jung:2004} are also related to our work.
The algorithm in \cite{Aftosmis:1997} recovers the outside  surface obtained from
self-intersecting meshes. The output mesh is obtained by identifying
facets, or part of facets, which are on the exterior. The algorithm
\cite{Wonnhyung-Jung:2004} uses the same idea, applied in the context
of mesh offsetting. As explained below in detail, we generalize these approaches to the more general
situations of any mesh deformation.

As a  {\bf hybrid method}, the recent work of  Wojtan et
al. \cite{Wojtan:2009cr} is representative, where the topological
changes to the mesh are handled by first identifying merging or
splitting events at a particular grid resolution, and then locally
creating new pieces of the mesh in the affected cells using a standard
isosurface creation method. The topologically simplified portions of
the mesh are stitched to the rest of the mesh at the cell
boundaries. While the authors present very convincing results, they
acknowledge some limitations, such as the restriction to a particular
grid cell size, as well as some topological concerns related to
matching exactly the extracted isosurface to the original mesh, among
others. 

In addition to the two above categories it is worth to mention also {\bf Solid modeling methods} 
that provide practical tools to represent and manipulate surface
primitives. Methods in this domain fall into two categories:
Constructive Solid Geometry (CSG) \cite{Foley:1990gf,hubbard:1990} and
Boundary Representation (B-Rep) \cite{Baumgart:1974,Braid:1978}. CSG
methods represent shapes as a combination of elementary object shapes
based on Boolean operations. Alternatively, B-Rep methods adopt the
more natural approach to represent the object boundary using vertices,
edges and facets \cite{Agrawal:1994,Rossignac:1999}. Each
representation has its advantages. While Boolean operations on CSG
objects are straightforward, a lot of computational effort is required to
render CSG objects
\cite{Rappoport:1997,Goldfeather:1986}. On the other hand, it is much
more difficult to implement  Boolean operations on boundary
representations (multi-resolution surfaces)
\cite{Biermann:2001,Nathan-Litke:2000}, whereas interactive rendering
is trivial.  
While these methods propose solution for computing Boolean operations
of surfaces, to the best of our knowledge they do not deal with any
extension needed to address  self-intersecting meshes. Generally, the
methods are more concerned with the rendering of the resulting
geometry than with the generation of correct manifolds in the case of
self-intersections. 

\subsection{Contributions}

In this paper we propose a novel 
topology-adaptive self-intersection removal
method for triangular meshes as well as an associated efficient algorithm,
\MeshAlg{}, with guaranteed convergence and numerical stability. We
generalize previous work in this area~\cite{Aftosmis:1997,Wonnhyung-Jung:2004} to any topology changes
resulting from mesh deformation,
including \textit{merges}, \textit{splits}, \textit{hole formations},
and \textit{hole losses}, e.g. Figure~\ref{fig:topological_changes}.
The main contribution is that, given an input mesh with self-intersections, the algorithm provides
a \CorrectMesh{} that represents the {\it outside skin} of the input mesh. 
Such an input mesh is typically obtained by applying arbitrary
deformations to its vertices, as is often the case with such techniques as surface
evolution, surface morphing, or multi-view 3D reconstruction.

The vast majority of the mesh-based surface deformation algorithms
available today are based on \textit{topology-preserving} methods. Alternatively, we propose a
\textit{topology-adaptive} mesh evolution method that is entirely
based on \MeshAlg{}. Such topology-adaptive scheme is
more general and hence better adapts to challenging applications such as
3D reconstruction using multiple images and non-rigid surface tracking.

Recent image-based reconstruction methods~\cite{Seitz:2006uq} make use of surface
evolution to obtain accurate 3D models.  Our
approach contributes in this field by providing an efficient
unconstrained mesh-based solution that allows for facets of all sizes
as well as for topology changes, with the goal of increasing precision
without sacrificing complexity.  The robustness and flexibility of the
proposed  framework is also validated in the context of mesh morphing,
showing several topologically challenging examples.

The remainder of this article is organized as
follows. Section~\ref{sec:background} provides some background
concepts on which our method resides.
Section~\ref{sec:meshalg} describes in detail the \MeshAlg{}
algorithm. Various aspects of the algorithm, such that the topological
changes that it can handle, convergence, numerical stability and time
complexity are detailed in
section~\ref{sec:meshdiscussion}. 
Section~\ref{sec:applications}
describes the mesh evolution algorithm based on \MeshAlg{}, as well as two sample applications: mesh morphing in section~\ref{sec:app-morphing} and 3D reconstruction is section~\ref{sec:app-reconstruction}.
Finally, we conclude in section~\ref{sec:conclusion}.  

\MeshAlg{} is available for download\footnote{\url{http://mvviewer.gforge.inria.fr/}} as an open-source software (OSS) package under the general public
licence (GPL).

\section{Background}
\label{sec:background}

Before we introduce the  \MeshAlg{} algorithm  we precise the context within which it applies. We assume an initial mesh representing the surface of a real object to be deformed into a self-intersecting input mesh from which the \MeshAlg{}  algorithm extracts an output mesh.  More precisely, we assume that the initial mesh represents  a {\it compact oriented 2-D manifold} with possibly several components and we expect the output mesh to do the same.  Consequently both initial and output meshes should satisfy the following properties: every edge belongs to exactly two flat faces; every vertex is surrounded by a single cycle of edges and faces; faces are oriented and do not intersect except at edges and vertices. The deformation that the initial mesh underwent  can then be any transformation that preserves the mesh graph structure, i.e. its connectivity.  Hence any vertex displacement field that preserves edges is acceptable. Note that this excludes displacements that fuse neighboring vertices.

The  \MeshAlg{} algorithm relies on the identification of outside or exterior faces on the deformed input mesh. An exterior face on an oriented mesh is a boundary face that delimits interior and exterior regions and that is oriented towards an exterior region, i.e. its normal points outward. To further identify regions delimited by the mesh as interior or exterior we need a rule.  Traditionally, interior and exterior regions are defined with an {\it even-odd parity} rule. Such rule simply consists of counting the number of intersections of a ray, emanating from a point, with the delimiting primitive. If this number is odd, the point belongs to an interior region, if not, the point is on the exterior. While efficient, this rule originally applies to simple primitives, e.g.  simple closed curves  in 2D and  closed surfaces in 3D, and does not correctly handle more complex primitives in particular self-intersecting primitives. In that case, the {\it winding} rule allows regions to be better differentiated by using  the primitive's orientation. This appears to be crucial when operating topological changes, such as merge and split, over regions. 

\begin{figure}[!tbp]
\centering
\includegraphics[width=6cm,angle=0]{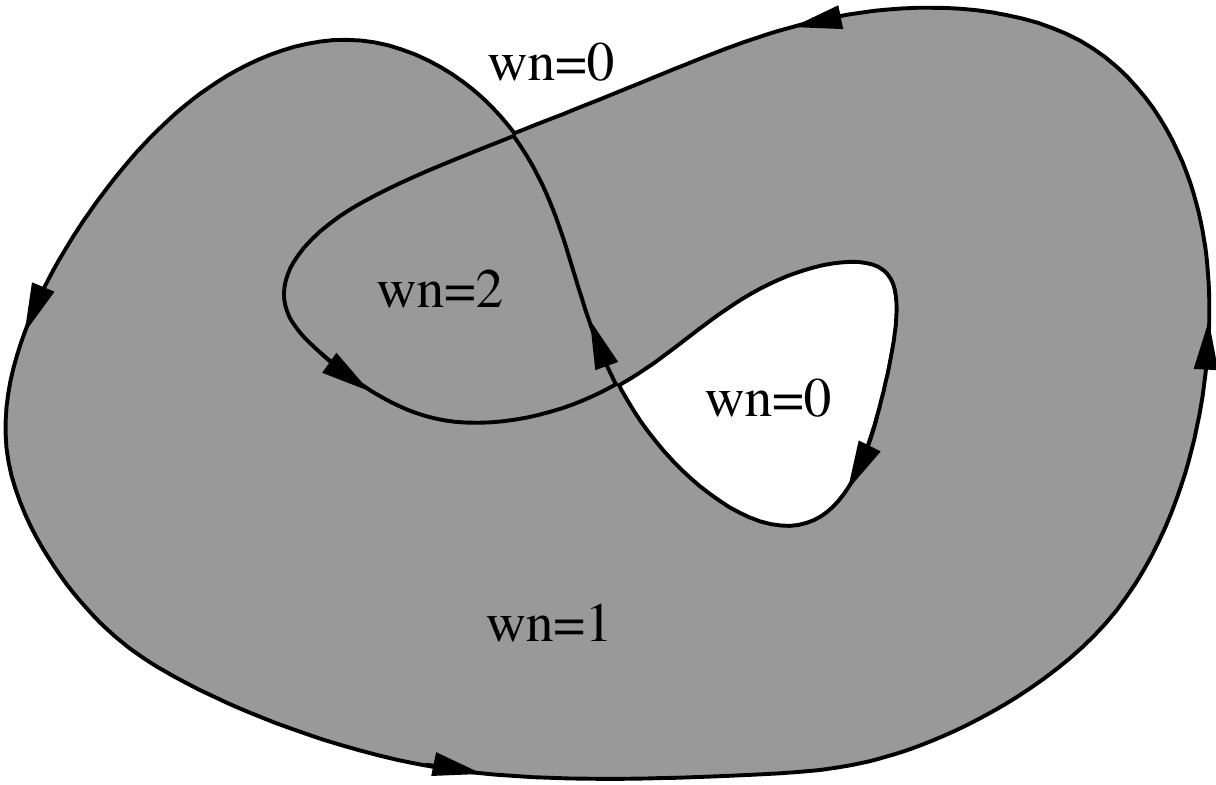}
\caption{Interior and exterior regions delimited by an oriented primitive. A point is on the exterior when it belongs to a region with a winding number $wn$ equal to $0$, on the interior otherwise.  }
\label{fig:insideoutside}
\end{figure}

The winding number of a point $p$ with respect to an oriented primitive is the number of times the primitive {\it winds} or cycles around $p$. Cycles are counted positively or negatively depending on their orientations  around the point. $p$ is then outside when its winding number is $0$, inside otherwise with positive or negative orientations depending on the sign of the winding number. Figure \ref{fig:insideoutside} depicts this principle in 2D. 

\begin{figure}[!tbp]
\centering
\includegraphics[width=7cm,angle=0]{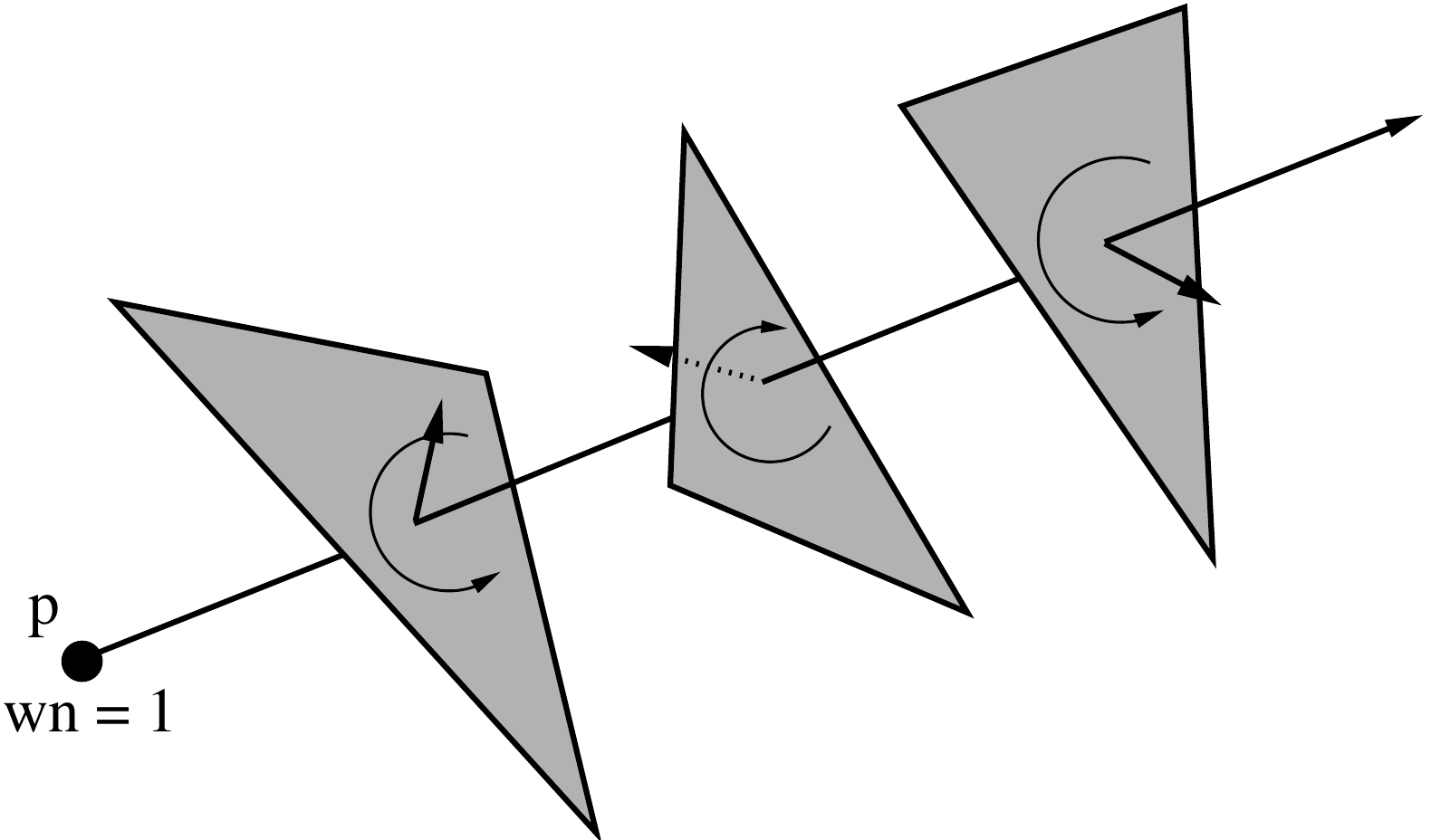}
\caption{The winding number at $p$ can be obtained by summing the dot product signs with face normals along any ray from $p$.  }
\label{fig:winding}
\end{figure}

To compute this number, two strategies can be followed. A first strategy consists in computing the total signed angle, solid angle in 3D,  made by a ray from the point under consideration to another point traveling along the primitive~\cite{carvalho95}. The sum will be equal to $0$ for a point on the exterior and a multiple of $2\pi$, $4\pi$ in 3D, for a point on the interior. Another strategy considers a ray from a point and its intersections with the primitive~\cite{computergraphics}. Each intersection is assigned a value $+1$ or $-1$ according to the sign of the dot product of the ray direction with the normal to the primitive at the intersection. If this sign is negative the value is $-1$ and $+1$ otherwise, see Figure \ref{fig:winding}. The sum of these values  will be $0$ only for a point on the exterior. We use this strategy to verify whether a face is on the exterior. We take a ray from the center of the face towards its normal direction and we sum the values  $-1$ and $+1$ obtained at the intersections with other faces along the ray. The face is on the exterior when this sum is $0$.\\

We call then  a {\it valid face }  a face  fully on the exterior without intersections with other faces and a {\it partially valid face}  a face divided by intersections into sub-parts, some of which being on the exterior. Notice that valid faces can be found inside the mesh, as independent connected components may appear inside the mesh as a result of self-intersections. Although these components are valid parts of the resulting mesh, they are usually not considered in evolution processes that do rely on criteria applying on exterior surfaces only, distance or photo-consistency for example.


\section{The \MeshAlg{} Algorithm}
\label{sec:meshalg}

\begin{figure*}[!htbp]
\begin{center}
\includegraphics[width=0.9\textwidth]{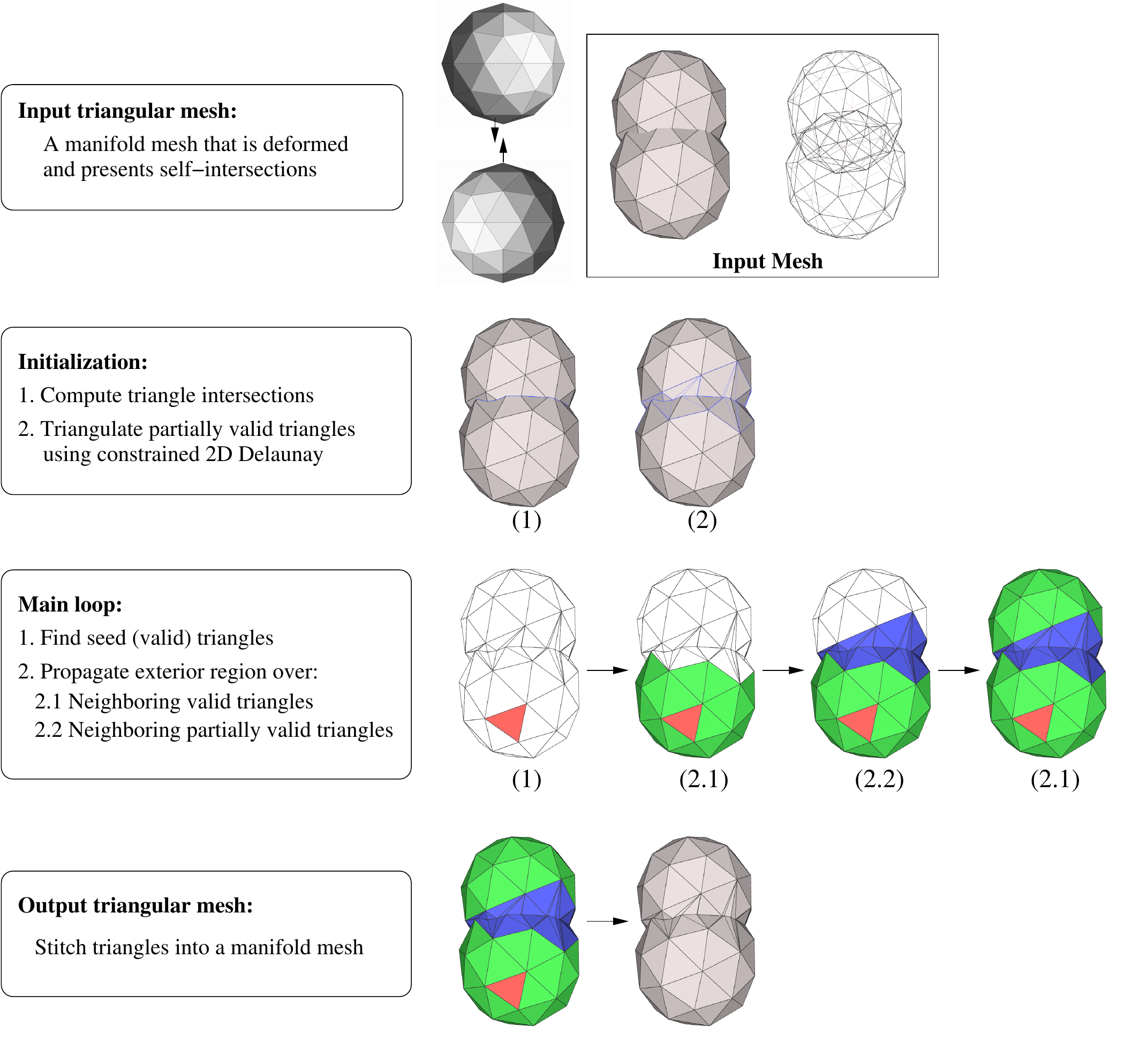}
\end{center}
\caption{Overview of \MeshAlg{}.}
\label{fig:alg_layout}
\end{figure*}

 The \MeshAlg{} algorithm removes self-intersection and adapts to topological changes in triangular meshes using an intuitive geometrically-driven solution. In essence, the approach preserves the surface consistency, i.e. 2-D manifoldness,  by detecting self-intersections and considering the subset of the original surface that is still {\it outside}. In order to identify the corresponding faces in the mesh, the method consists in first finding an initial seed face that is fully on the exterior, using the winding rule presented in the previous section,  and then propagating the exterior label over the mesh faces by means of region growing. Figure \ref{fig:alg_layout} illustrates the algorithm, the different steps are detailed in the following.

\subsection{Self Intersections}
\label{sec:alg-intersections}
The first step of the algorithm consists of identifying self-intersections, i.e. edges along which triangles of the mesh intersect. 

This information will later on be needed in the computations, since it delimits the outside regions. In the general situation, one would have to perform $O(n^2)$ checks, with $n$ the number of triangles, to verify all triangle intersections, which can become  expensive when the number of facets is large. In order to decrease the computational time, we use a bounding box  test to determine which bounding boxes (of triangles) intersect, and only for those perform a triangle intersection test. We use the fast box intersection method implemented in \cite{cgal:kmz-isiobd-06} and described in \cite{Zomorodian:2002fk}. The complexity of the method is then  $O(n \ log^3 (n))$.

\subsection{Valid Region Growing}
\label{sec:region-growing}
The second step of the algorithm consists of identifying exterior
triangles in the mesh. A valid region growing approach is used to
propagate validity labels on triangles that composed the {\it outside}
of the mesh. Alternatively, it can be viewed as a "painting"
procedure, as it was originally described in
\cite{Aftosmis:1997}. Following this idea, we present here the
sub-steps of the region-growing procedure. First, in the {\it
  Seed-triangle finding} step, valid triangles are sought as starting
triangles without intersections that reside on the exterior. In the
next {\it Valid triangle expansion}  step this information is
propagated by expanding on neighboring valid triangles  until
triangles with intersections are reached.  The {\it Partially valid
  triangle traversal}  step details then how to traverse the valid
sub-parts of intersection triangles as well as how to cross from one
intersecting triangle to the other. The local sub-parts are
triangulated using a constrained 2-D Delaunay triangulation. The
underlying idea that guides this step is to propagate the normal
information from the seed triangles using the local geometry. The algorithm seeks to maintain the orientation of the original surface. When generating the output mesh, the orientation of the valid triangles is preserved. The newly formed sub-triangles (partially valid triangles) will inherit the orientations of the parent triangles.

\paragraph{Seed-triangle finding}
A seed-triangle is defined as a non-visited valid triangle, found
using the winding rule previously introduced. In other words, a
seed-triangle is a triangle that is guaranteed to be on the
exterior. This triangle is crucial, since it constitutes the starting
point for the valid region growing. If found, the triangle will be
marked as valid; otherwise, we assume that all outside triangles are
identified and the algorithm  jumps to the next stage (section
\ref{sec:alg-stitching}). We have adopted the efficient AABB tree
implementation described in \cite{cgal:atw-aabb-09} for the
ray-to-triangles intersection test. 

\paragraph{Valid triangle expansion}
Region growing over valid triangles is simply  performed by checking  neighbors of a valid triangle and stopping on the intersections:  if the neighboring triangle  is non-visited and has no intersections, then it is marked as valid;  if  the neighboring triangle is non-visited and has intersections, then it is marked as partially valid together with the entrance segment and direction, corresponding in this case to an oriented half-edge.

\begin{figure}[!tbp]
\centering
\subfigure[Triangle intersections]{\includegraphics[height=0.42\columnwidth]{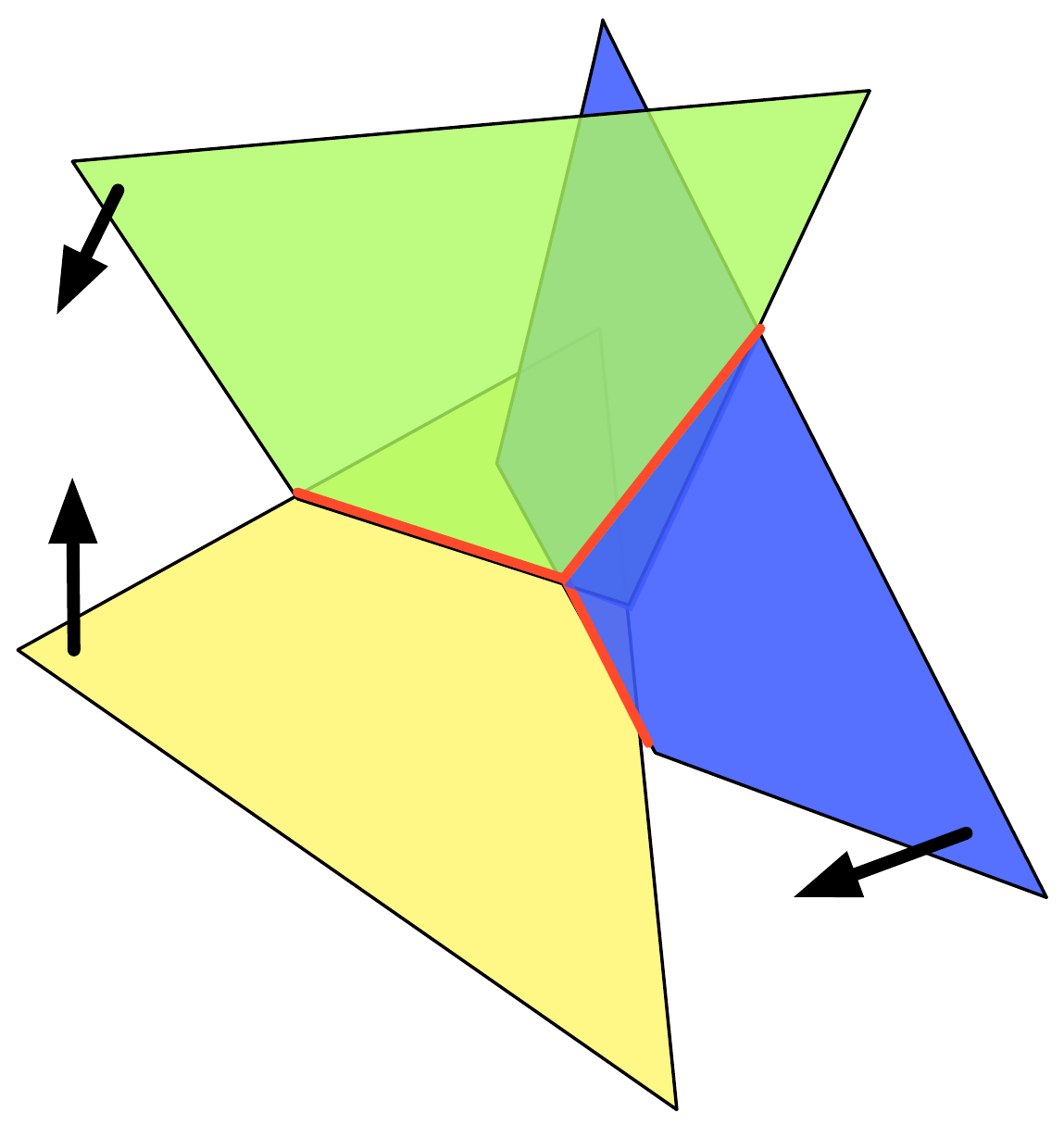}}
\subfigure[Partially valid triangle traversal]{\includegraphics[height =0.45\columnwidth]{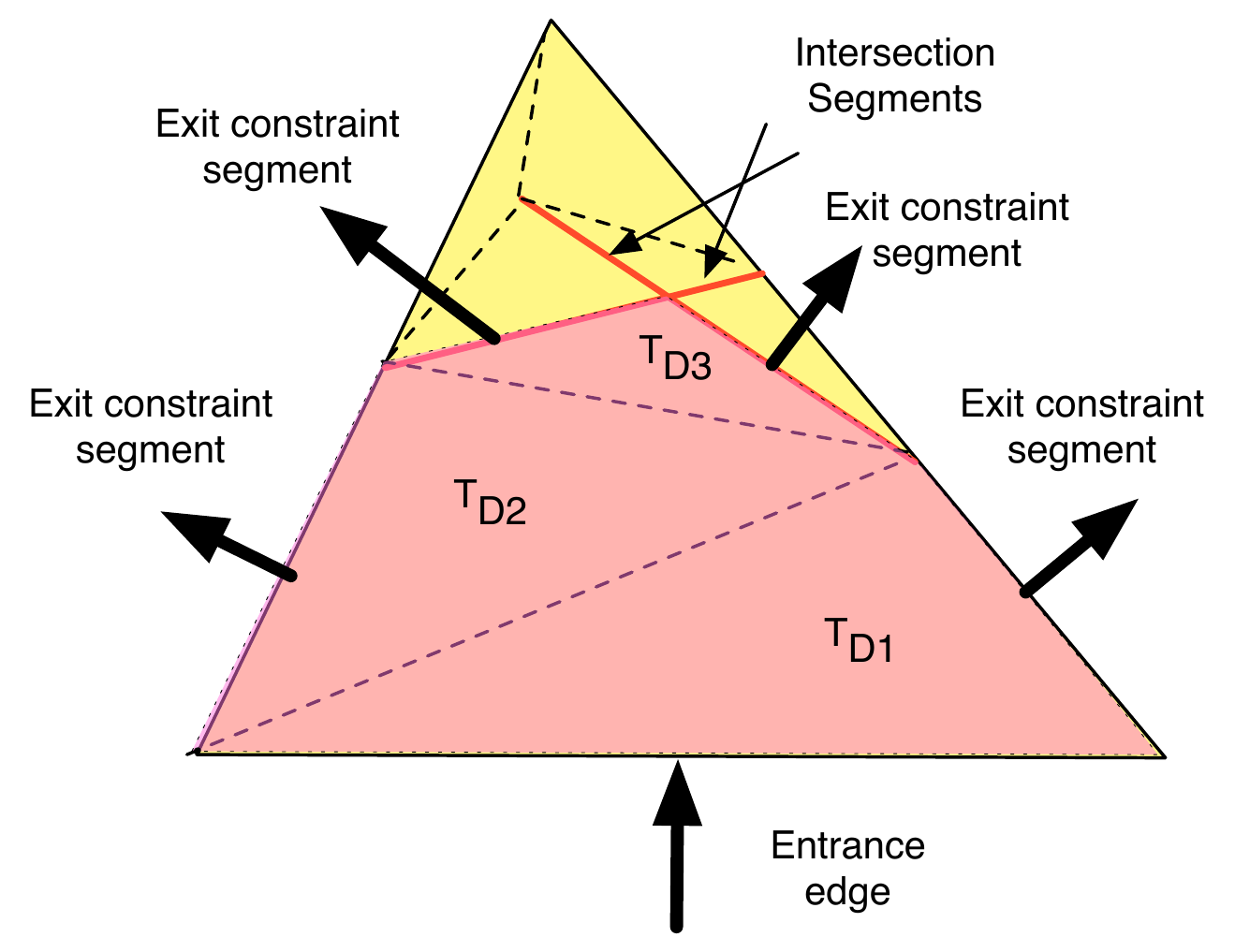}}
\caption{Partially valid triangle traversal. (a)The intersections with all other triangles are computed for each intersecting triangle. (b) close-up of the bottom triangle in (a). The local geometry is re-defined using a constrained 2-D Delaunay triangulation that ensures the presence of the original triangle edges and the intersection segments. The traversal starts at the entrance edge and stops on constraint edges thus marking $T_{D1}$, $T_{D2}$ and $T_{D3}$ as valid.}
\label{fig:delaunay}
\end{figure}

\paragraph{Partially-Valid triangle traversal}
In this step proper processing of regions containing intersections is ensured, with local geometry being generated. Let $t$ be a partially valid triangle as marked during the valid triangle expansion step. We have previously calculated all the intersection segments between this triangle and all the other triangles. Let $S_{t}=\{s_{ti}\}$ represent all the intersection segments between triangle $t$ and the other triangles. In addition, let $H_{t}=\{h_{tj}| \textrm{for }j=1..3\}$ represent the triangle half-edges. A constrained 2-D triangulation  performed in the triangle plane, using \cite{cgal:hs-chdt3-06}, ensures that all segments in both $S_{t}$ and $H_{t}$ appear in the new triangular mesh structure and that propagation can be achieved in a consistent way. A fill-like traversal is performed from the entrance half-edge to adjacent triangles, stopping on constraint edges, as depicted in Figure \ref{fig:delaunay}.

\begin{figure}[!tbp]
\centering
\subfigure[]{\includegraphics[height=1.5in]{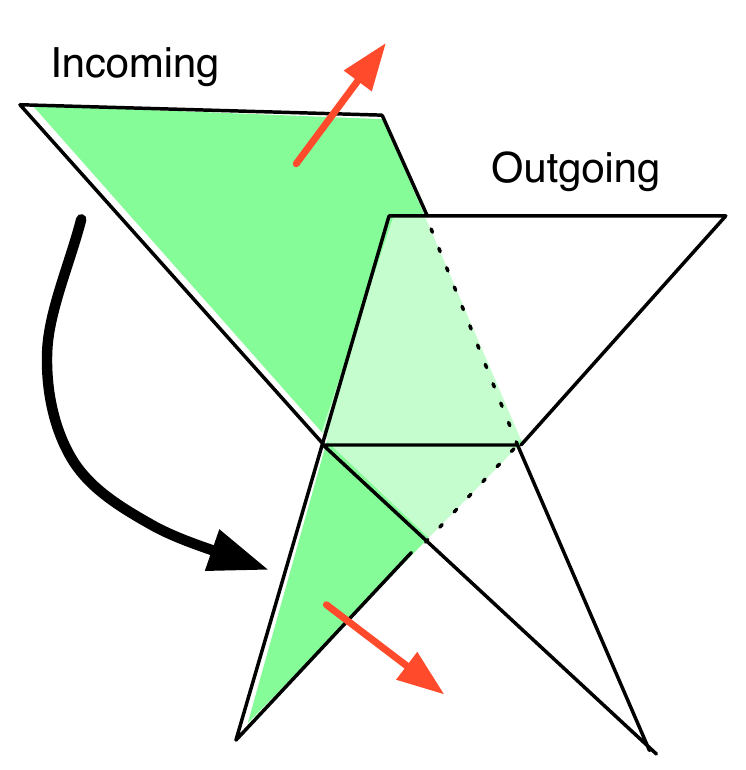}} \quad
\subfigure[]{\includegraphics[height=1.5in]{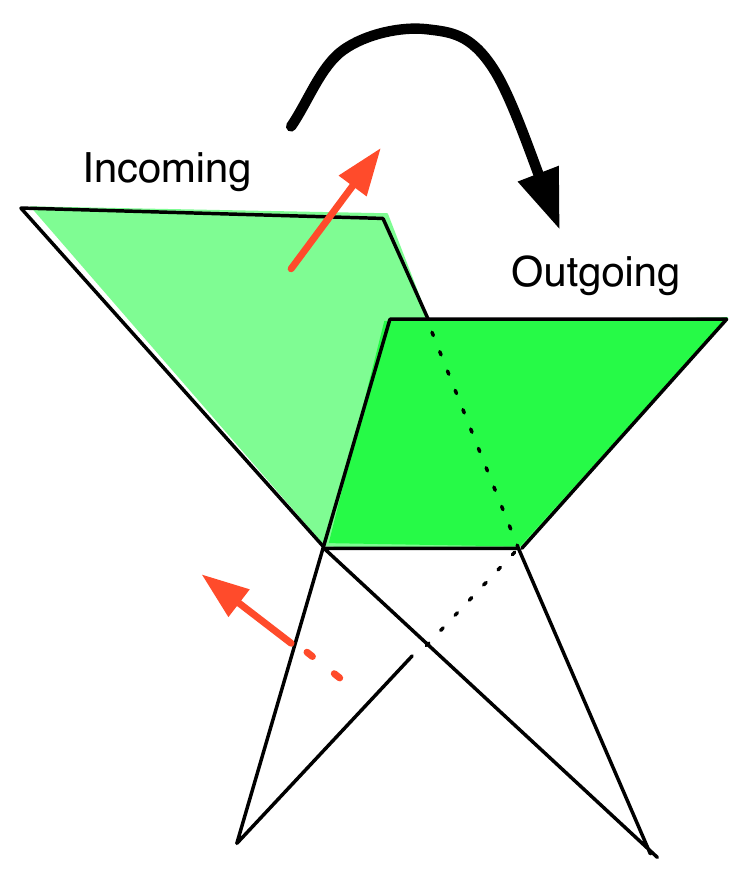}}

\caption{The two partially valid triangle-crossing cases.}
\label{fig:crossing_river}
\end{figure}

Choosing the correct side of continuation of the "fill" like region growing when crossing from a partially valid triangle to another is a crucial aspect in ensuring a natural handling of topological changes.  The correct orientation is chosen such that, if  the original normals are maintained, the two newly formed sub-triangles  would preserve the water-tightness constraint of the manifold. This condition can also be casted as follows:  the normals of the two sub-triangles  should be opposing each other when the two sub-triangles are "folded" on the common edge. A visual representation of the two cases is shown in Figure \ref{fig:crossing_river}. The triangles on the other side of the exit constraint edges will be marked  as valid appropriately, based on whether they contain any intersections or not.

Note that it is possible to visit a partially valid triangle multiple times, depending on whether there are multiple isolated (non-connected) exterior components. However, each sub-triangle formed by the local re-triangulation is only visited once. The simplest example to image is a cross, formed out of two intersecting parallelepipeds. There will be intersecting triangles appearing on both sides.

\subsection{Triangle Stiching}
\label{sec:alg-stitching}

\begin{figure}[!htpb]
\centering
\subfigure[Singular vertex]{\includegraphics[height=1.4in,width=1.1in]{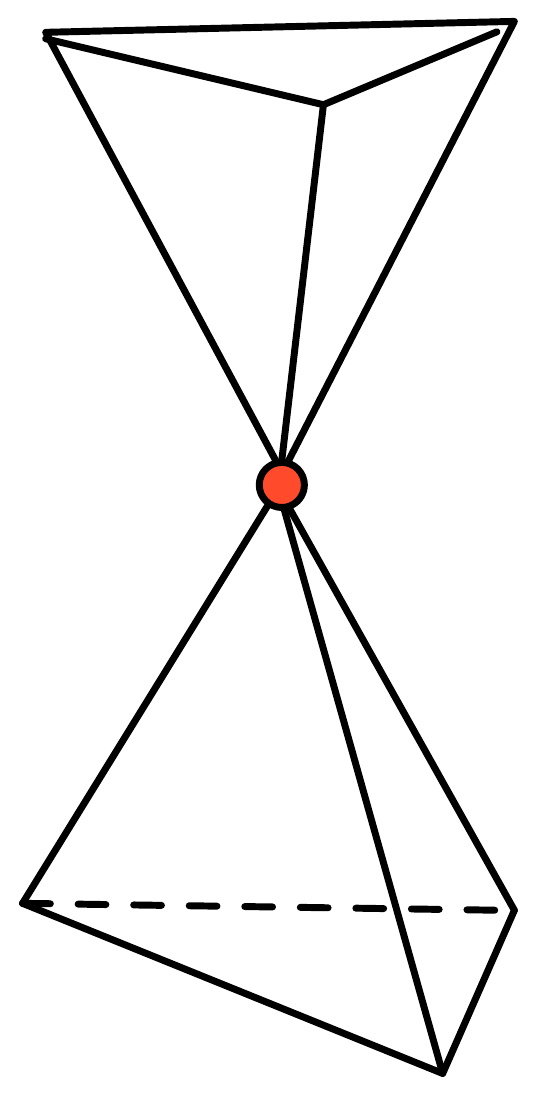}} 
\subfigure[Singular edge]{\includegraphics[height =1.6in]{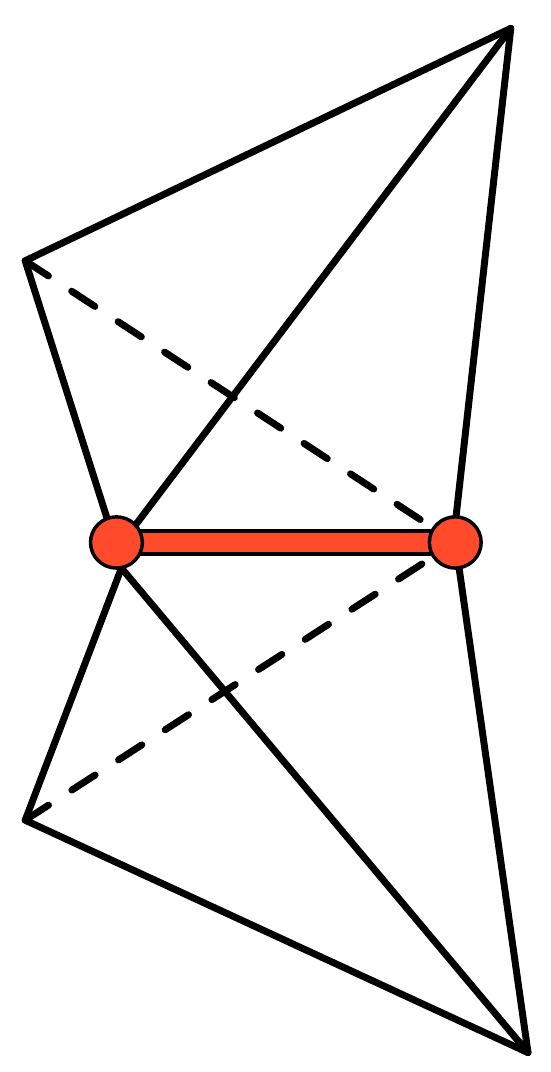}} 
\subfigure[Singular face]{\includegraphics[height =1.5in]{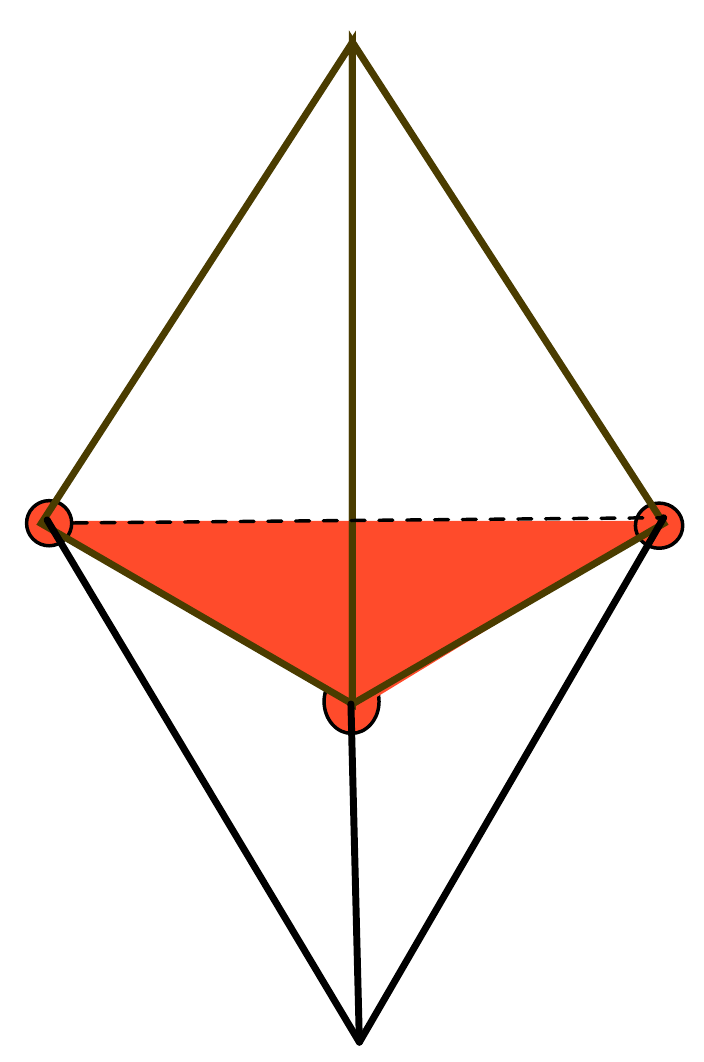}}
\caption{Special cases encountered while stitching a triangle soup.}
\label{fig:stitching_special_cases}
\end{figure}

The region growing algorithm described previously will iterate until
there are no more unmarked triangles to visit. At this stage, what
remains to be done is to stitch together the 3-D triangle soup
 in order to obtain a valid mesh which is
manifold. We adopt a method  similar in spirit to
\cite{Gueziec:2001,Shin:2004}. In most cases this is a straight
forward operation, which consists of identifying the common vertices
and edges between facets, followed by stitching. However, there are
three special cases, in which performing a simple stitching will
violate the mesh constraints and produce locally non-manifold
structures. The  special cases, shown in Figure
\ref{fig:stitching_special_cases}, arise from performing stitching in
places where the original structure should have been maintained. We
adopt the naming convention from  \cite{Gueziec:2001}, calling them
the singular vertex case, the singular edge case and the singular face
case. All cases are easily identified by performing local operations.

\paragraph{Singular vertex case} (Figure
\ref{fig:stitching_special_cases}(a)). A vertex is shared by two or
more different regions. In this case, the manifold property stating
that for each manifold point, there is a single neighborhood, does not
hold. To detect these cases, the algorithm proceeds simply by checking
that all facets incident to a vertex are within one neighborhood. The
steps are: starting from a facet of $v$, mark it visited and do the
same with its non-visited neighbors that are also incident to $v$
(neighboring triangles are chosen based on the available mesh
connectivity); the process is repeated until all the neighboring
facets are processed; if by doing so we exhausted all the neighboring
facets, vertex $v$ is non singular, otherwise it is singular, so a
copy of it is created and added to all the remaining non-visited
facets. The process is repeated until all the incident facets are
visited. 

\paragraph{Singular edge case} (Figure
\ref{fig:stitching_special_cases}(b)). An edge is shared by two or
more different regions, hence the manifold property does not
hold. Such cases are detected and repaired by the singular vertex
detection step, which will correctly identify and duplicate the two
vertices that form the singular edge. 

\paragraph{Singular triangle case} (Figure
\ref{fig:stitching_special_cases}(c)). A triangle is shared by two or
more different regions, hence the manifold property does not hold.
Such cases are detected and repaired by the singular vertex detection
step, which will correctly identify and duplicate the three vertices
that form the singular triangle. 

Given that the original input mesh does not contain any of the above
singular simplex scenarios, they rarely occur in practice. Note,
however, that there are situations where creases are formed on the
mesh, usually when inverting mesh regions,  that can degenerate into
singular cases.

\section{Algorithm Analysis}
\label{sec:meshdiscussion}

Having introduced the algorithm in the previous section, we discuss in this section some of its most important aspects, including the handling of topological changes, the guarantee to obtain a valid mesh given a valid input mesh,  the numerical stability and the time complexity.

\subsection{Topological Changes}
\begin{figure}[!htbp]
\centering
\subfigure[Merge]{\includegraphics[width=0.8in]{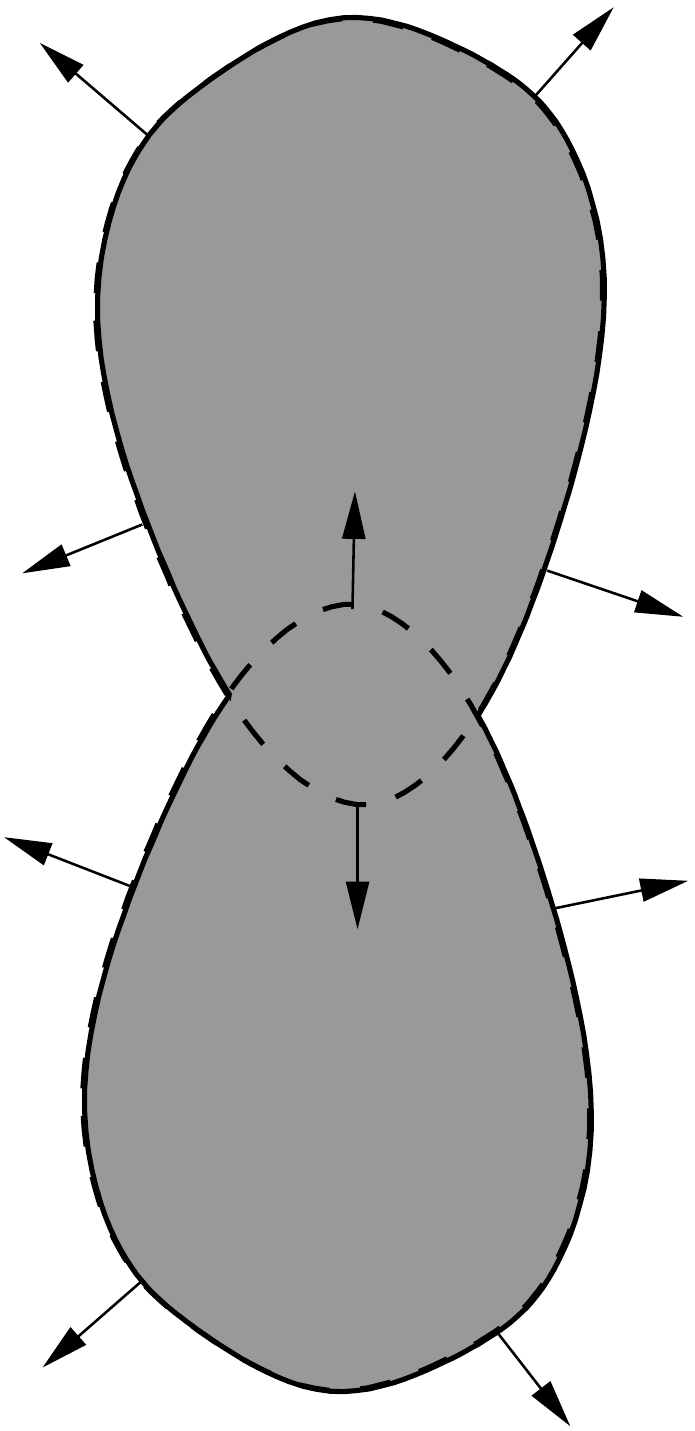}} \quad \hspace{0.4in}
\subfigure[Split]{\includegraphics[width=0.8in]{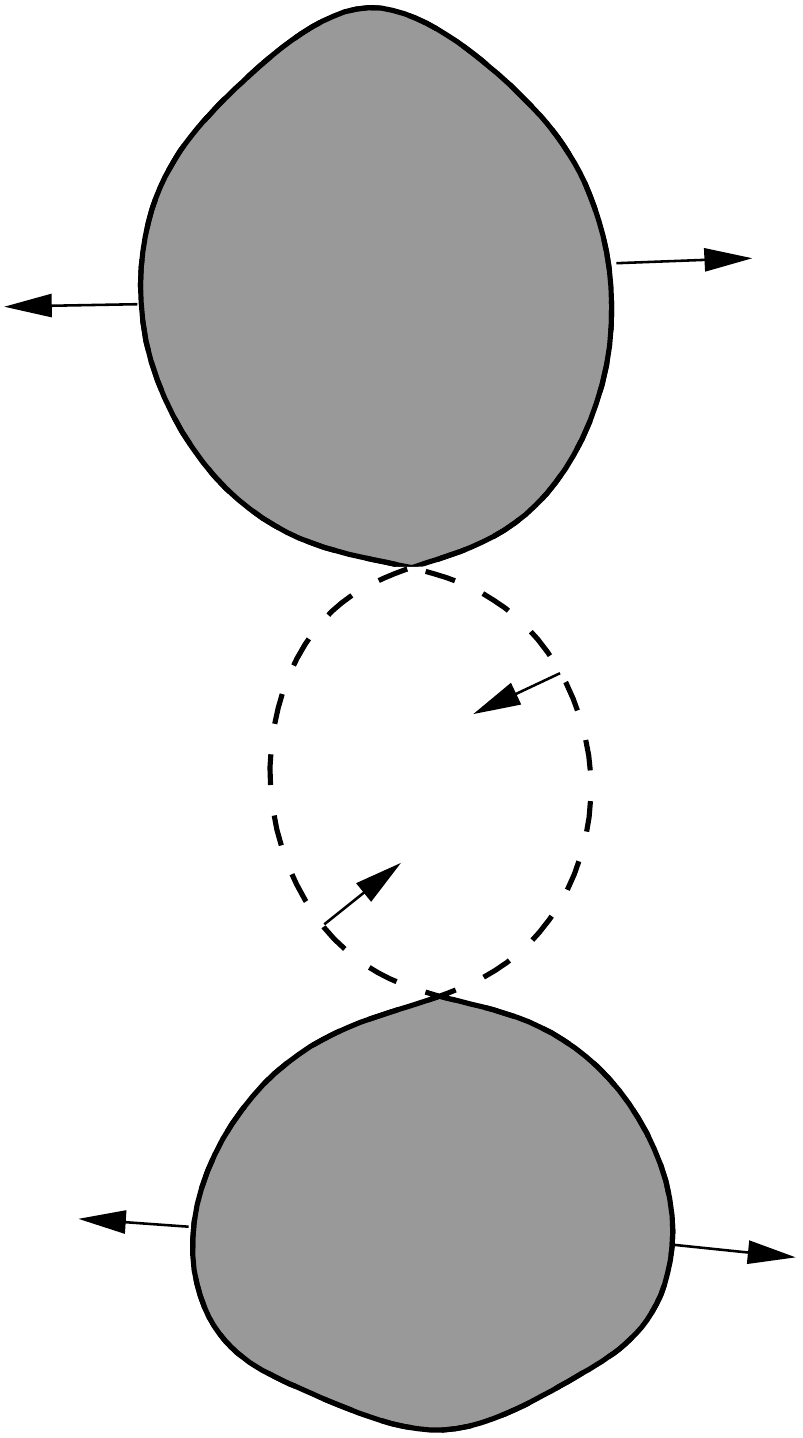}} \quad 
\subfigure[Hole formation]{\includegraphics[width=1.3in]{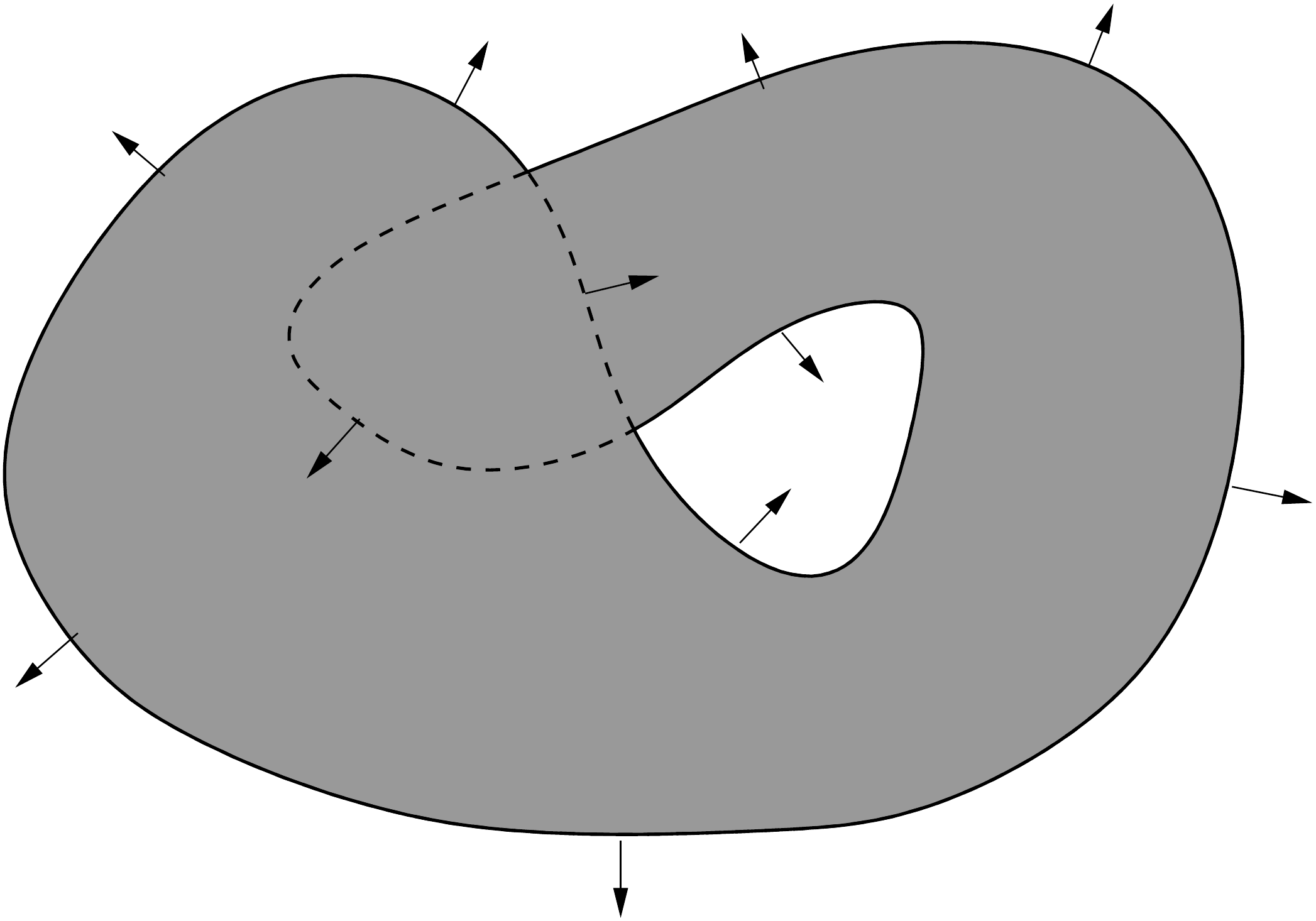}} \quad
\subfigure[Hole loss]{\includegraphics[width=1.4in]{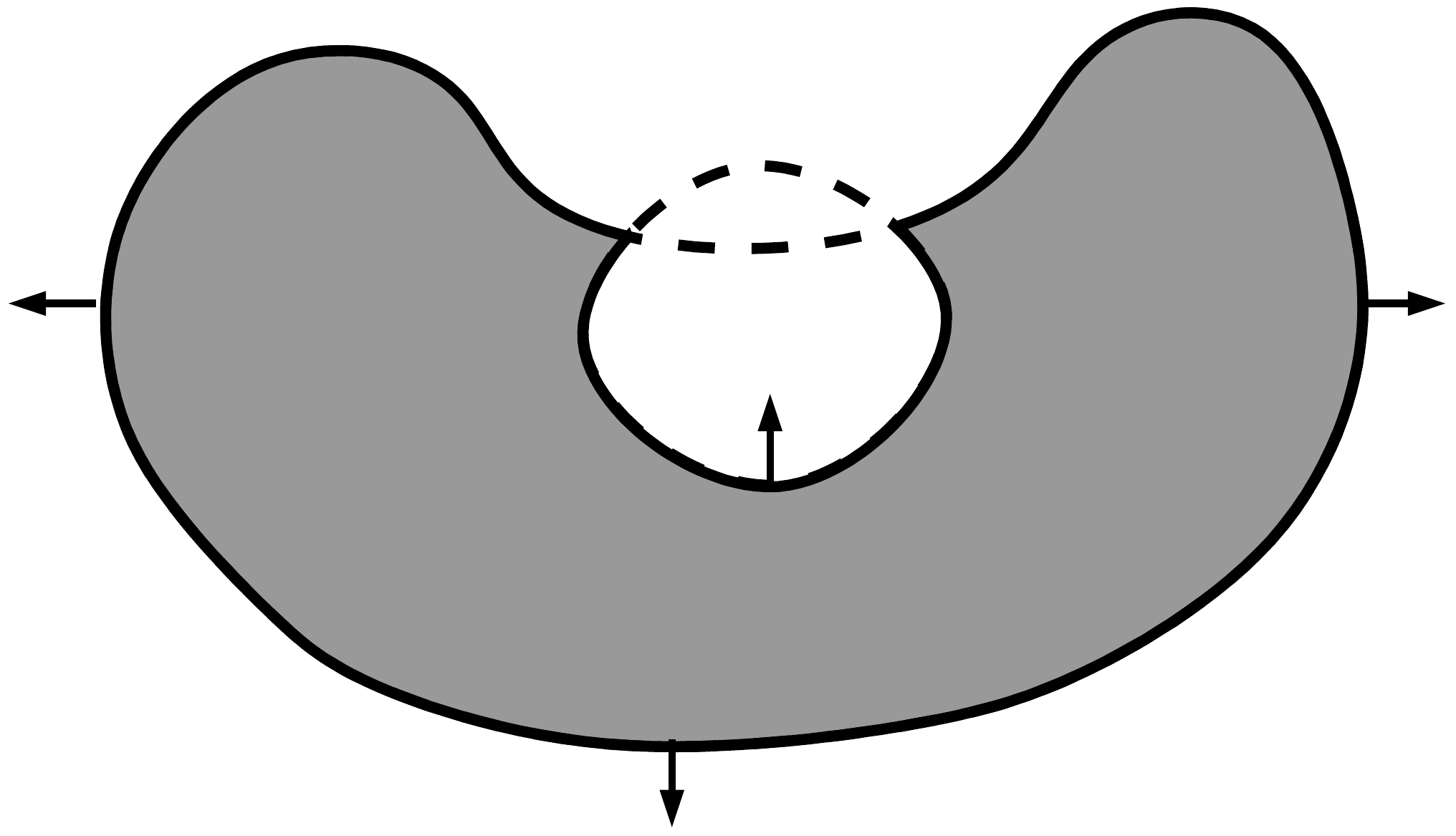}} \quad
\caption{Topological changes  (2-D simplified view). Regions delimited by non-exterior faces (dashed lines) are eliminated by the algorithm.}
\label{fig:topological_changes}
\end{figure}

A nice feature of the algorithm is that it correctly handles topological changes that result from the modification of the local geometry, i.e. faces that appear and disappear. We consider compact surfaces and in the general case, topological changes that can occur are: merge, split, hole formation and hole loss. They are depicted in Figure \ref{fig:topological_changes}.  Note that in 3D hole cases correspond to situations where a connected component is inside another connected component and that topological changes where handles appear or disappear are covered by the merge and split cases (see Figure \ref{table:morph_evolutions} for examples).  Note also that regions delimited by non-exterior faces, shown by dashed lines in Figure \ref{fig:topological_changes}, are eliminated by the algorithm.

The partially valid triangle crossing technique described earlier 
in Section \ref{sec:region-growing} and detailed in Figure \ref{fig:crossing_river}
ensures a  \textit{natural} handling of these topological changes that plagued most of the mesh approaches until now. The {\it merge} case scenario, shown in Figure \ref{fig:topological_changes}(a), coincides in spirit with the {\it union} Boolean set operation $\cup^*$. Less intuitive is the {\it split} operation, which will typically occur during a mesh evolution process, when certain parts will thin out up to the moment when some triangles from opposite sides will cross each other hence defining an inverted inside region with a negative winding number. Such a case is depicted in Figure \ref{fig:topological_changes}(b), in a mesh morphing scenario, where the initial surface has $1$ connected component and the destination $2$ connected components. 

The two other examples, hole formation and loss, are less frequent. While handled by the algorithm, as mentioned earlier, inside valid faces, e.g. Figure \ref{fig:topological_changes}(c), are usually not considered in the surface evolution processes described here.

\subsection{Guarantees}

Given that the input mesh is a \CorrectMesh\ that has been deformed by a motion field and assuming exact computations (see section \ref{sec:numerical_stability}), \MeshAlg{} will recover \CorrectMesh\ components. The number of components depends on the number of seed triangles detected. The algorithm will always \textit{finish}, because it does not revisit already traversed sub-parts. In addition, it is guaranteed to always find the \textit{exterior surface}, since it starts from a valid seed triangle, thus on the exterior, and it always rests that way, by propagating the normal information. The computed output is \textit{manifold} by construction, since it traverses a valid input manifold and accounts for the manifold violations with the degenerate cases. It is \textit{compact}, since the original input surface has no border and the algorithm does not build any, i.e. there is always a way outside a triangle intersection.

In addition, the 2-D manifold correctness is guaranteed by identifying and correcting all the possible 2-D manifold neighborhood violations when performing triangle stitching  (singular vertex, singular edge and singular facet). 

The algorithm preserves the geometry and orientation of the input mesh, with the exception of the self-intersection areas, where local triangulations redefine the geometry.

\subsection{Numerical Stability}
\label{sec:numerical_stability}

The numerical stability is critical, in order to be able to guarantee that the output is valid. It is ensured by using exact arithmetic predicates when computing intersections and when disambiguating the boundary cases. Boundary cases are defined as non-typical cases. For example, in a triangle-triangle intersection test, the typical cases are when the intersection is a 2-D segment or when there is no intersection at all, whereas the boundary cases are when the intersection is a point, a 2-D polygon or a line segment on one of the triangle edges.

The boundary cases are disambiguated using the \textit{simulation of simplicity} technique of virtual perturbations \cite{Edelsbrunner:1990}. It involves inducing a small vertex perturbation locally, which will force the boundary triangle-triangle intersections into one of the classical cases.

Mesh offsetting was used as a way to perturb the original mesh, where each vertex is moved along its normal by a small step.  While it is impossible to guarantee that a perturbation will completely eliminate the boundary cases, the algorithm detects the re-occurrence of such a boundary condition and it applies another perturbation. Please note that any other mesh perturbation can be used.

The choice of using the \textit{simulation of  simplicity} technique to handle boundary cases is motivated  by the targeted application,  mesh evolution, where such boundary situations rarely occur
and where the explicit handling of all special cases would penalize the algorithm.  \textit{Simulation of simplicity} is complementary to the approach proposed by M\"antyl\"a in \cite{Mantyla:1986sf}, where all  the possible boundary scenarios are handled explicitly, making the method more suitable for applications where exact intersections are required (i.e. boolean operations with CAD models).

\subsection{Time Complexity} 

 The overall time complexity of the algorithm depends on the number $n$ and relative sizes of facets and it is of $O(n \log^3(n))$ expected time (the average case). This complexity is dominated by the number of operations required to determine intersections. Each triangle requires $O(\log^3{n})$ tests, thanks to the fast box intersection method used, described in \cite{Zomorodian:2002fk} and implemented in  \cite{cgal:kmz-isiobd-06}. The complexity of the method is  $O(n \ log^d (n) + k)$ for the running time and $O(n)$ for the space occupied, $d$ the dimension ($3$ in the current case), and $k$ the output complexity, i.e., the number of pairwise intersections of the triangles. In practice, more than 80\% of the running time is spent computing the self-intersections. Typically, the running time for performing the self-intersections test is under $1$ second for a mesh with $50,000$ facets on a $2.6$ GHz Intel Core2Duo, with no multi-threading, and where exact arithmetic is used for triangle intersections and where the self-intersections are in the range of $100$.\\

\subsection{Comparison with a Static Strategy}
\label{sec:dis-static}

Alternatively, one could use the winding test, described in Section \ref{sec:background}, in order to replace the propagation step described in Section \ref{sec:region-growing}. Instead of growing the valid region outside it would test all the existing  triangles and sub-triangles obtained from local Delaunay triangulations and only choose the triangles that  reside on the exterior, after which it would proceed to the final triangle stitching step.
However, this static scheme would take considerably longer time, since it requires the same initial time $O(n \log^3{n})$ to compute all the triangle intersections and local Delaunay triangulations, followed by the additional time required for the valid triangle test, which is not negligible.


\subsection{Extension to Open Surfaces} 
\label{sec:open-surfaces}
The algorithm can be extended to open-surfaces, i.e. surfaces
  with holes, without significant modifications. The only part that
  does not work in the current formulation is the seed triangle
  finding, since the winding number will not necessarily reflect the
  correct in/out mesh information. In order to properly deal with this
  issue, the holes are temporarily closed just for winding numbers
  calculations, then the algorithm is run unchanged. Because the
  temporarily added hole filling triangles will not to be taken into
  account during triangle intersection tests or region growing, there
  is no need for an advanced hole filling technique that searches to
  maximize the overall surface smoothness. A simple hole filling
  technique is employed: add a center vertex around the hole contour
  and connect it with all the participating border half-edges. 

Such an extension not only allows one to deal with open surfaces, but also permits one to tackle self-intersection removal in very large meshes, where the area of interest is relatively local.

\subsection{Implementation Details} 
\label{sec:implementation}
In our implementation we used CGAL (Computational Geometry
Algorithms)  C++ library \cite{cgal:eb-06}, which provides
\textit{guaranteed}, robust and efficient implementations for various algorithms. We have
used the following CGAL modules: N-dimensional fast box intersections,
2-D constrained Delaunay triangulation, AABB trees, triangular meshes and support
for exact arithmetic kernels.  As a pre-processing step, the triangle degeneracies are eliminated (see the upcoming section).


\section{Mesh evolution and Applications}
\label{sec:applications}

In this section a mesh evolution algorithm  is introduced based on \MeshAlg. Two applications using this framework are introduced: mesh morphing and multi-view 3-D reconstruction, allowing to test various configurations.

\subsection{Mesh evolution}

A number of methods exist in the literature that deal 
with deformable surfaces,  such as Kenneth Brakke's Evolver
\footnote{\url{http://www.susqu.edu/facstaff/b/brakke/evolver/evolver.html}},
Wojtan and Turk's visco-elastic simulator \cite{Wojtan:2008eu} or the
work of Celniker and Gossard on deformable surfaces
\cite{Celniker:1991lq}. Nevertheless, the above mentioned methods are
all mesh-based topology preserving. This might be or not a desired
feature of the algorithm, depending on the target application.  It is
our goal, in the current section, to introduce an intuitive
{\it generic mesh evolution paradigm} that is topology adaptive,
based on TransforMesh. The main steps of the algorithm are presented in
Figure \ref{fig:generic_alg_layout} and detailed below. More implementation details follow.

\subsubsection{Algorithm} Within each evolution iteration,
there are four steps. Firstly, a velocity vector field  $\mathcal{\vec
  F}$ is computed for each vertex of the mesh  $\mathcal{M}$. This
step is application specific. Secondly, the mesh is deformed 
using the computed velocity vector field $\mathcal{\vec F}$ and a
small time step $t$. Thirdly, \MeshAlg{} is invoked in order to clean the
potential self-intersections and topological problems introduced by
the second step. The fourth step involves mesh optimization, with the
goal of ensuring  good mesh properties. Ideally, a mesh should consist
of triangles as close to equilateral as possible, which allows for
better computations of local mesh properties, e.g. curvatures and
normals. To this purpose, a number of sub-steps are being performed:
adaptive remeshing, vertex valence optimization and 
smoothing.  These four main steps are repeated until the mesh has reached the
desired final state, also application  specific. 

\begin{figure}[!htbp]
\begin{framed}
\begin{description} \item[ \textbf{Algorithm}: Generic Mesh Evolution with \MeshAlg{} ] \end{description}
\begin{itemize}
\item[] {While Not Finished} 
\begin{itemize}
\item[1.] { \textbf{Compute Velocity Vector Field} $\mathcal{\vec F}$ of velocities for each vertex of the mesh $\mathcal{M}$, using application specific information;}
\item[2.] { \textbf{Evolve Mesh} $\mathcal{M}$ using the vector field $\mathcal{\vec F}$ and a small time-step $t$;}
\item[3.] { \textbf{Invoke \MeshAlg{}}\ on $\mathcal{M}$ in order to clean self-intersections and topological problems;}
\item[4.] { \textbf{Mesh Optimization:}}
\begin{itemize}
\item[a)] {{\it Adaptive Remeshing}: ensures that all edges are within a safety zone interval; }
\item[b)] {{\it Vertex Valence Optimization}: improve the quality of the triangles;}
\item[c)] {{\it Smoothing}: improve the mesh based on a smooth surface prior.}
\end{itemize}
\end{itemize}
\end{itemize}
\end{framed}
\caption{The generic mesh evolution algorithm using \MeshAlg{}.}
\label{fig:generic_alg_layout}
\end{figure}

\subsubsection{Implementation details} In practice, during the second step, the mesh is deformed 
using the computed velocity vector field $\mathcal{\vec F}$ and a
small time step $t$, thresholded by a maximum movement $\alpha \cdot
e_{avg}(v)$, where $\alpha$ is a user-set threshold (typically between
0.1-0.3) and $e_{avg}(v)$ represents the local average edge length for
a vertex $v$. The adaptive remeshing step ensures that all edges are
within a safety zone interval $(e_1,e_2)$, which is user-defined. This
prevents edges from reaching sizes close to zero. In practice, this is
obtained through edge swap, edge split or edge collapse operations. 
Edge collapses are only performed if not violating the manifold constraint. Additionally, connected components where all edges are smaller than $e_1$ and that have a volume less than 
$\pi/6\: e_1^3$, are also removed. The vertex valence
optimization step performs edge swaps in an attempt to ensure an
overall vertex valence of $6$ \cite{Kobbelt:2000}. Vertex valence is
defined as the number of edges shared by a vertex.  The ideal
vertex valence of $6$ is desirable because, assuming that the manifold
is generally locally planar, it is  equivalent to obtaining $60^\circ$
for each of the sharing triangle angles, thus optimizing for
equilateral triangles. Alternatively, the vertex valence can also be
improved by performing edge swaps only if it increases the minimum
angle of either triangle adjacent to the edge. 
The Laplacian smoothing is attained by computing the discrete mesh
Laplacian  \cite{Meyer:2002}, i.e. the discrete
Laplace-Beltrami operator, $\Delta v$ for each vertex $v$ of the
mesh. Furthermore, the mesh is smoothed using $v \rightarrow v - \beta
\Delta v$.  If smoothing will artificially shrink small components and remove surface details, then the 
Laplace-Beltrami operator could be used twice, as proposed in \cite{Delingette:1992},by taking into account higher order surface information: $v \rightarrow v - \beta_1\Delta v + \beta_2\Delta \Delta v$. Alternatively, if no smoothing is necessary, $\beta$ can be set to $0$. These mesh optimization steps ensure that degenerate triangles, that is triangles with zero area, are properly handled and eliminated. Degenerate triangles can affect the output accuracy of some of the geometric calculations, such as triangle normal estimation, triangle-triangle intersection tests or Delaunay triangulations. Please note, nevertheless, that TransforMesh uses local perturbations in order to eliminate any potential left-over boundary cases.

\subsubsection{Choosing the correct time step} The currently presented mesh
evolution approach does not make any assumptions about choosing the
right time-step. This parameter is entirely application specific. The
TransforMesh algorithm does not have any information about the
temporal component. It is therefore entirely up to the user to choose
a meaningful time-step $t$ which will capture all the temporal
dynamics. The only measure proposed in the generic evolution algorithm
is to threshold the maximum vertex movement to  $\alpha\cdot
e_{avg}(v)$, in order to prevent both large jumps and to reduce the
number of intersections.

\subsubsection{Remeshing} The remeshing step is important and should
theoretically occur at each timestep, due to the fact that some
regions can become under-sampled in areas where the speed vector field
is divergent or over-sampled in areas where the speed vector field is
convergent. More importantly, intersections can generate poorly shaped
triangles, which would probably have an impact on the local numerical
process  applied to the mesh that produces the vector field.

We give below two mesh evolution examples, one for mesh morphing in section
\ref{sec:app-morphing}, demonstrating the ability of the algorithm to handle complex surface evolutions, and the other one for multi-view 3-D reconstruction in section \ref{sec:app-reconstruction}. In both cases, the
application specific information is detailed in order to compute the
vector fields $\mathcal{\vec F}_{morphing}$ and $\mathcal{\vec
  F}_{reconstruction}$, which plug directly within the generic mesh
evolution framework presented in Figure \ref{fig:generic_alg_layout}.

\subsection{Surface morphing}
\label{sec:app-morphing}

A straightforward mesh evolution application of our algorithm is surface
morphing, that is starting from a source surface $S_{A}$ and evolving
it towards a destination surface $S_{B}$. This will allow us to test
thoroughly various cases of topology changes. Surface morphing has
been widely described in the literature. We will adopt the method
proposed by Breen and Whitaker \cite{Breen:2001fk}. In the following section we will summarize the
reasoning that leads the surface evolution equation. 

\subsubsection{Methodology}

A metric that quantifies how much two surfaces overlap is defined (source surface $S_A$ and destination surface $S_B$). A natural choice of such a metric is the signed distance function $\gamma_B$ of the destination mesh $S_B$, defined as in the level set literature as being negative inside the shape $S_B$, zero on the surface, and positive on the exterior.
By considering the volume integral $\mathcal{M}_{S_B}(S_A)$ of any surface $S_A$ with respect to $\gamma_B$ (thus $S_B$), one can see that it will achieve the maximum when the two surfaces overlap. By taking the first variation of the metric $\mathcal{M}_{S_B}(S_A)$ with respect to the surface $S_A$ and a small displacement field and differentiating with respect to the vector field, one obtains the following evolution equation using a hill climbing strategy for each vertex $x$ along its normal $\mathbf{N}(x)$:

\begin{equation}
\mathcal{\vec F}_{morphing} = \frac{\partial S}{\partial t} = - \gamma_{B}(x) \mathbf{N}(x)
\label{eq:morph_final_equation}
\end{equation}

The evolution strategy described above will converge to a local minimum. Given the source surface $S_A$ and the destination surface $S_B$, $S_A$ will correctly find all the connected components of $S_B$ that are included in the original surface $S_A$. If $S_A$ represents a surface outside the destination surface $S_B$, $S_A$ will converge to an empty surface. We keep this result in mind when choosing the initial surface $S_A$.

\subsubsection{Complexity Issues and Mesh Discretization}

\begin{figure}[!ht]
  \begin{center}
    \includegraphics[width=0.7\columnwidth]{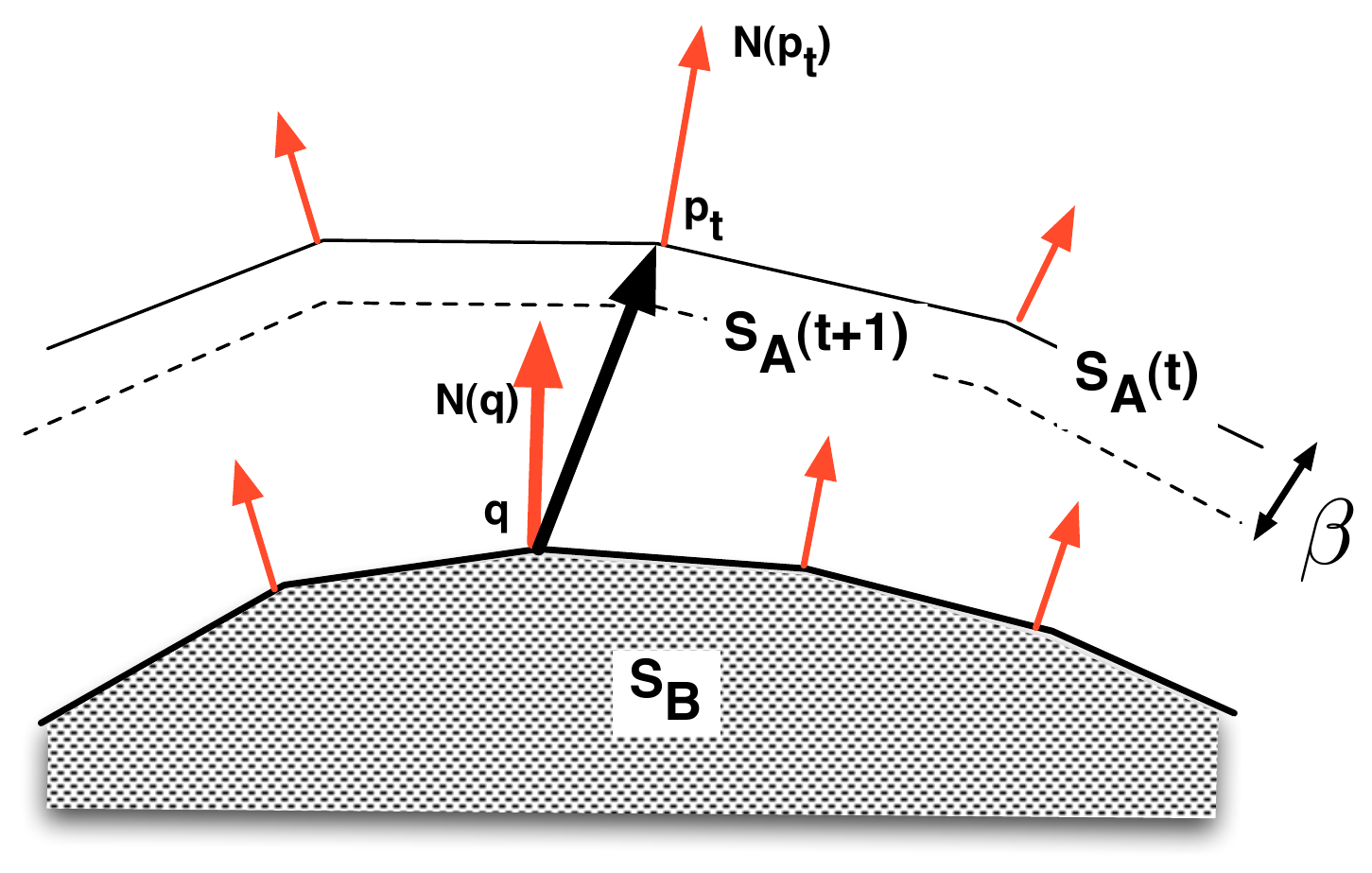}
  \end{center}
  \caption{Mesh Morphing evolution step. The surface $S_A$ evolves from time $t$ to time $t+1$ towards $S_B$. If for point $p \in S_A$, the closest point in $S_B$ is $q$, then the point $p$ will evolve along its normal with a magnitude of $(p-q) \cdot N(p)$, thresholded by a maximum user set evolution magnitude.}
  \label{fig:morph_evolution_theory}
\end{figure}

In the general case, in order to calculate an exact distance function $\gamma_B$, one would have to consider the distance from a query point to each of the facets of the mesh (representing the surface $S_B$), keeping the closest distance. This process will take $O(N_F)$, where $N_F$ represents the number of facets. This is a fairly expensive computation, which will have to be performed at each iteration throughout the evolution for every vertex.

There exists a large number of methods for computing 3-D distance fields. For a recent survey, please consult  \cite{Jones:2006sf}. As per \cite{Jones:2006sf}, the methods can be classified according to two criteria. According to the first criterion, they can be: 
\begin{itemize}
\item \textbf{Chamfer methods}, where the new distance of a voxel is computed from the distances of its neighbors by adding values from a distance template;
\item \textbf{vector methods} where each voxel stores a vector to its nearest surface point and the vector at an unprocessed voxel is computed from the vectors at its neighbors by means of a vector template and
\item \textbf{Eikonal solvers}, where the distance of a voxel is computed by a first or second order estimator from the distances of it's neighbors.
\end{itemize}
According to the second criterion, the distances can be propagated throughout the volume in a:
\begin{itemize}
\item \textbf{sweeping scheme}, when the propagation starts in one corner of the volume and proceeds in a voxel-by-voxel, row-by-row fashion to the opposite end, typically requiring multiple passes in different directions, or in a
\item \textbf{wavefront scheme}, when the distances are propagating from the initial surface in the order of increasing distances until all voxels are covered.
\end{itemize}

For our testing purposes, we propose an approximation/heuristic using the distance to the closest vertex point, as illustrated in Figure \ref{fig:morph_evolution_theory}.  If for a point $p \in S_A$, the closest point from $S_B$ is $q$, the evolution equation for point $p$ is:
\begin{equation}
\gamma_B(p) =  (q-p) \cdot N(p),
\end{equation}
where $\gamma$ was introduced in (\ref{eq:morph_final_equation}). Note that the vector magnitude will be thresholded to a maximum of $\alpha \cdot e_{avg}(p)$, as per step 2 of the generic mesh evolution algorithm, described in Figure \ref{fig:generic_alg_layout}. The distance and sign from a query point are computed on the fly, as
supposed to being stored in a distance field 3-D grid. The computation
time is reduced drastically due to the use of proper search structures. The search time for the nearest neighbor is $O(log(N_V))$, where $N_V$ represents the number of vertices.
There is an initial overhead of $O(N_V log(N_V))$ of building the search tree. In practice, we have used the implementation of \cite{hjaltason95ranking} available in CGAL. Note that if the target surface $S_B$ contains a good enough mesh resolution, this approximation is very close to the true signed distance function.
Also, if the accuracy of distance field computation is of concern, more exact implementations could be adopted \cite{Jones:2006sf}. 

In the case of sufficient sampling, the current approximation will return a vertex belonging to the closest triangle where the true projection would be. Thus, the error bound is the distance between the vertex and the projection. In practice, however, we do not use the actual distance, but its sign, in order to establish the direction of the evolution. This makes the current approximation fit for our purpose. Alternatively, one could easily verify all the incident triangles to the closest vertex to establish the true distance function, if the application requires it, keeping in mind that the sufficient sampling condition still applies.

The current heuristic only makes use of the mesh vertices of $S_B$,
together with their associated normals. This has the great advantage
of being able to be applied in the current formulation, not only to
meshes, but also to oriented 3-D points. This would allow one to morph
an initial mesh $S_A$ towards a set of oriented 3-D points $P_B$. If orientation information is not available, it can be estimated from neighboring points using principal component analysis \cite{Hoppe:1992}. Alternatively, in the context of multi-view stereo, it can be obtained via a minimization scheme \cite{Furukawa:2009pami}.

\subsubsection{Results}

In Figure~\ref{table:morph_evolutions} we present results obtained
with five test cases, entitled "Genus 3", "Thoruses", "Knots In",
"Knots Out" and "Open Plane". As it can be observed, the algorithm successfully deals
with merge and split operations as well as handling multiple connected
components. In addition, the "Open Mesh" example illustrates the algorithm extension for open surfaces, introduced in Section \ref{sec:open-surfaces}. The average computation time per iteration on a 2.6 GHZ
Intel Core2Duo processor varies between 0.07 to 1.6 seconds, depending
on the number of facets and on the number of intersections. More
detailed statistics are presented in Table~\ref{table:morph_stats}.  

In terms of parameter settings with respect to the generalized mesh
evolution framework depicted in Figure \ref{fig:generic_alg_layout}
within which we casted the current mesh morphing algorithm, we
considered $t=1$ for the timestep, $\alpha=0.2$ the average edge size
$e_{avg}$ for maximum movement amplitude and $\beta=0.1$ for the
smoothing term. Additionally, the original meshes had a constant mesh
resolution. Hence, we set the edge thresholds to $e_1= 0.7\cdot
e_{avg}$ and $e_2=1.5\cdot e_{avg}$.

\newcommand{\includegraphicsVerifier}[1]{\includegraphics[width=0.13\textwidth, height=0.10\textwidth, clip, trim=0.5cm 1.5cm 0.5cm 1.5cm]{#1}}

\begin{small}
\begin{figure*}[!htbp]
\centering
\begin{tabular}{ccccccc}


\raisebox{1em}{\begin{sideways} Genus 3 \end{sideways}} &
\includegraphicsVerifier{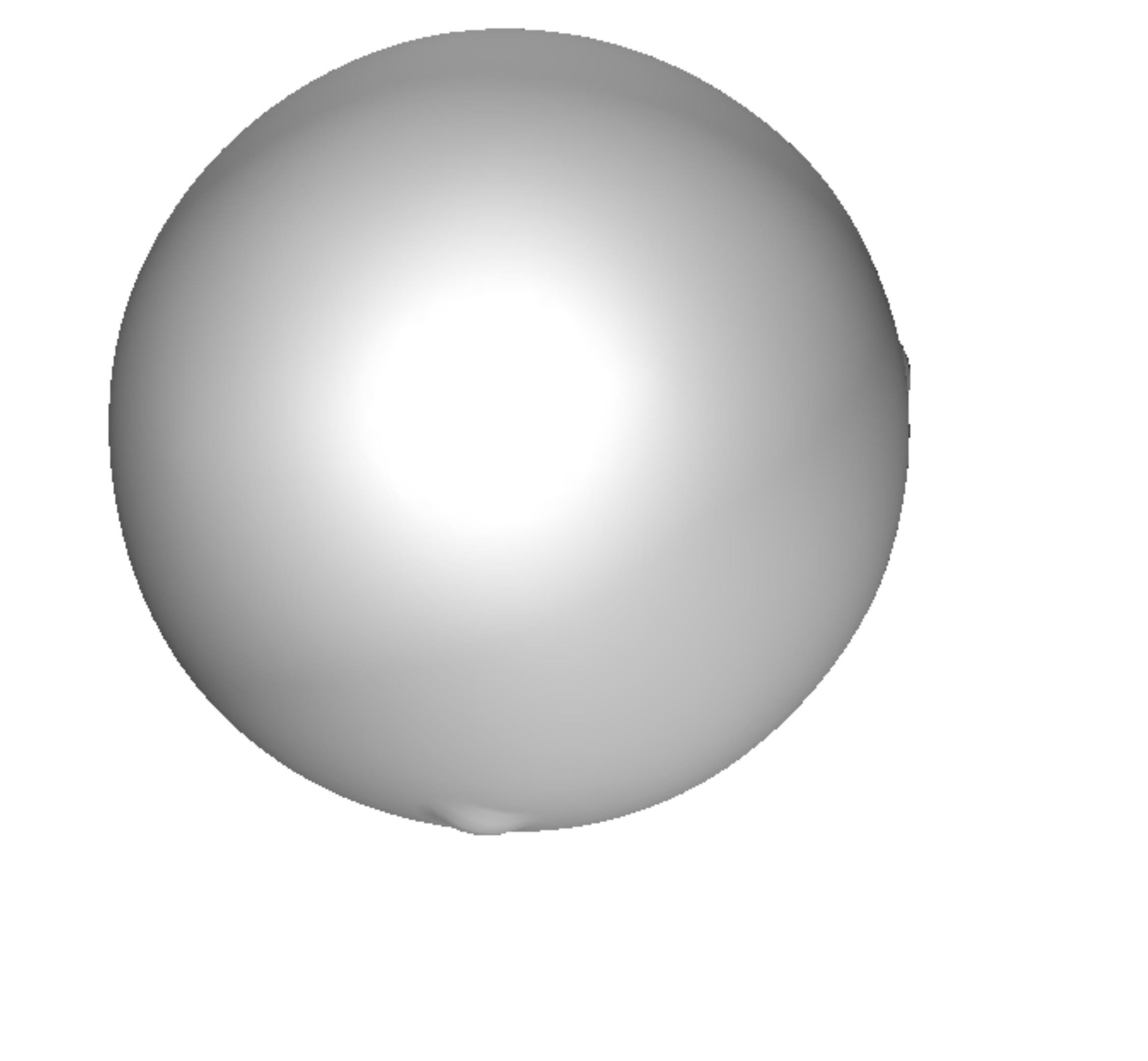} &
\includegraphicsVerifier{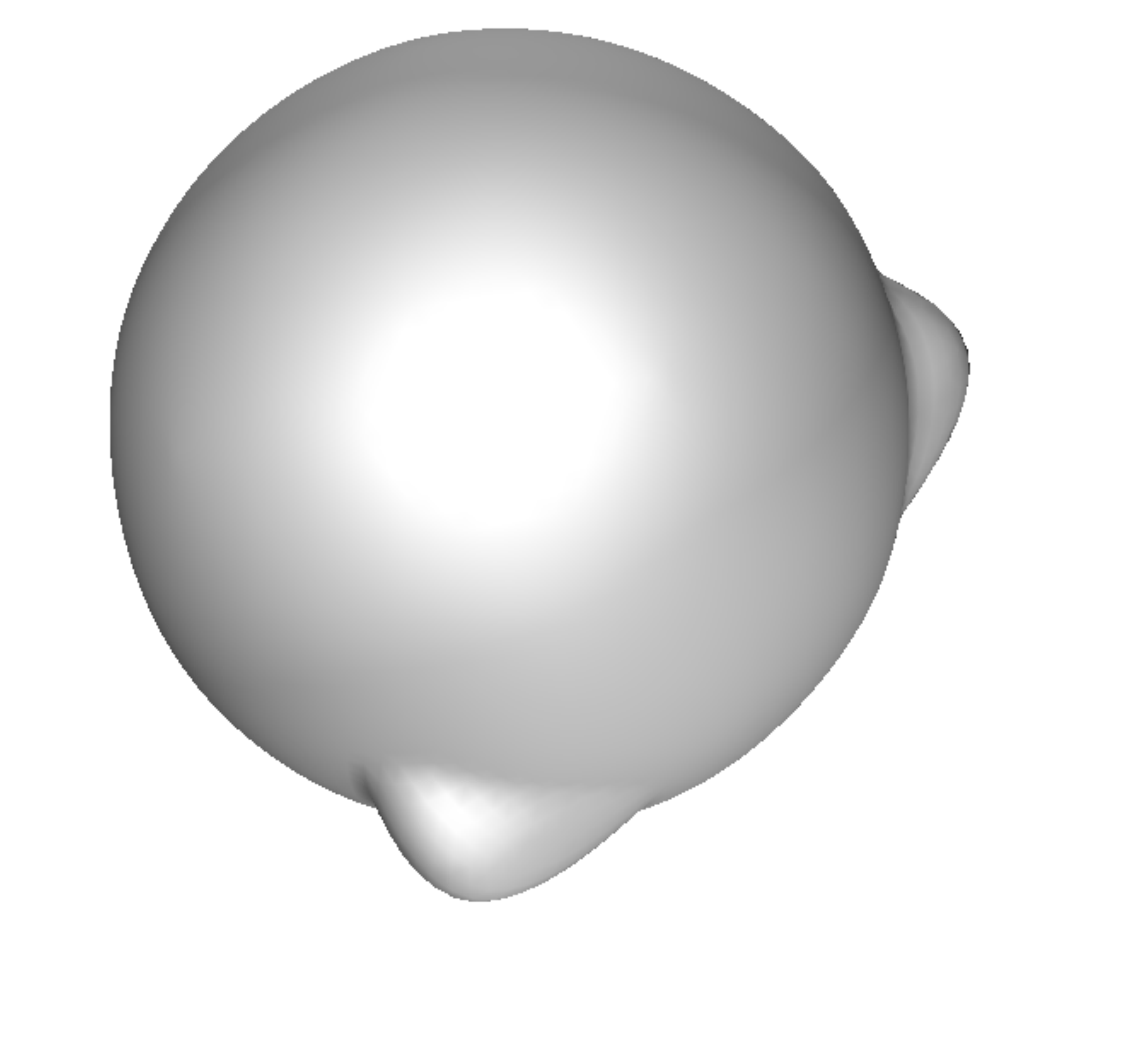} &
\includegraphicsVerifier{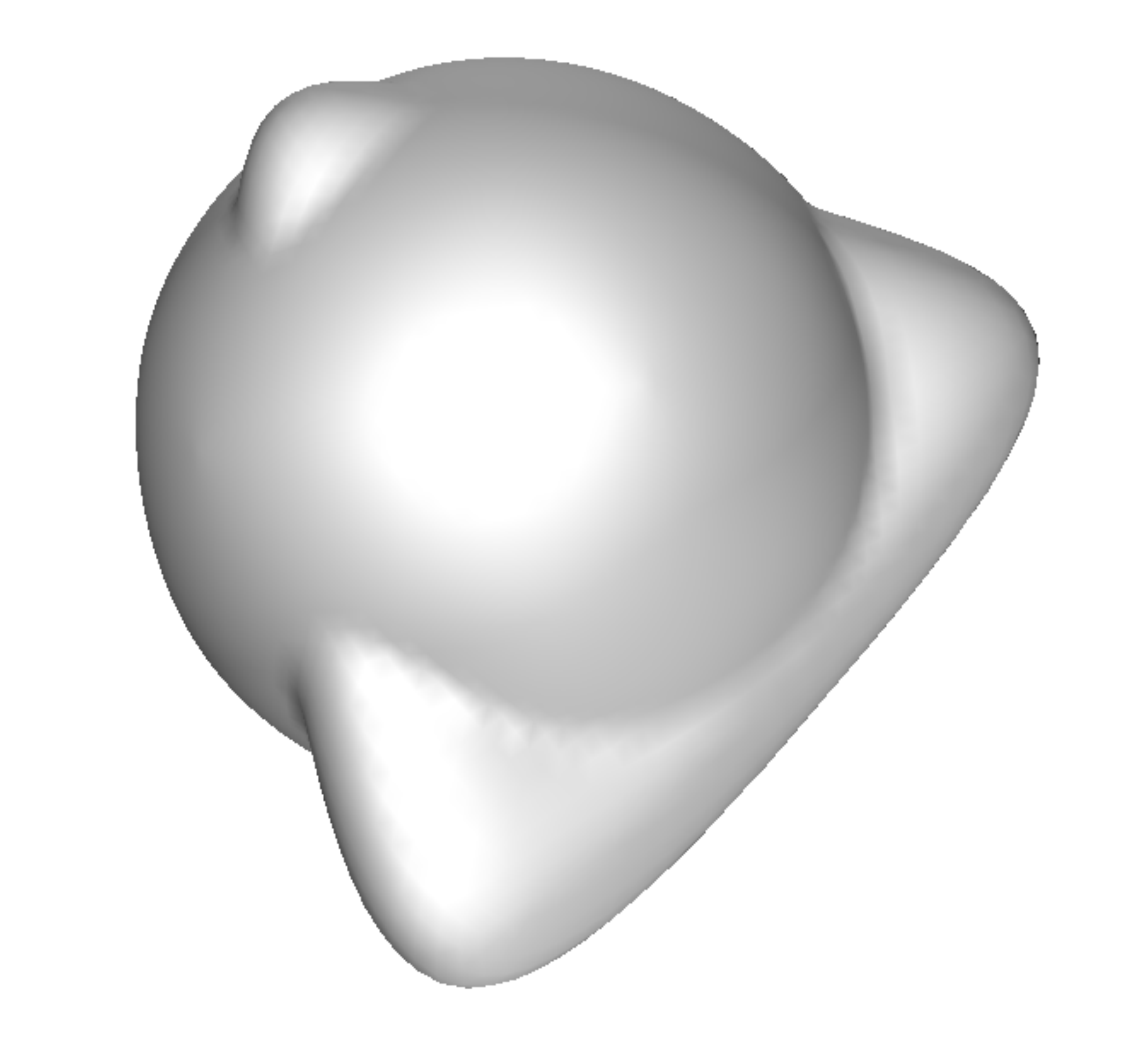} &
\includegraphicsVerifier{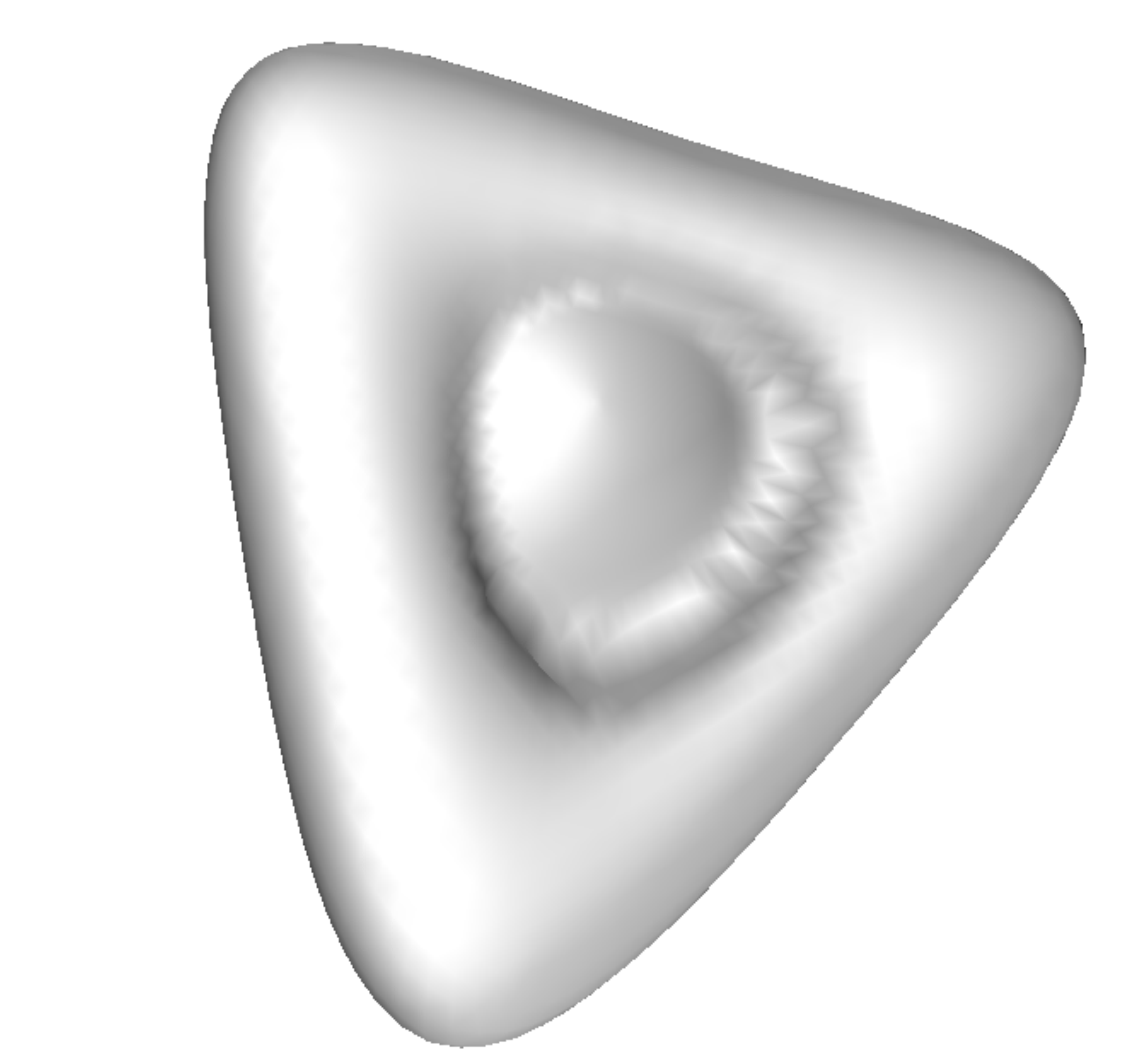} &
\includegraphicsVerifier{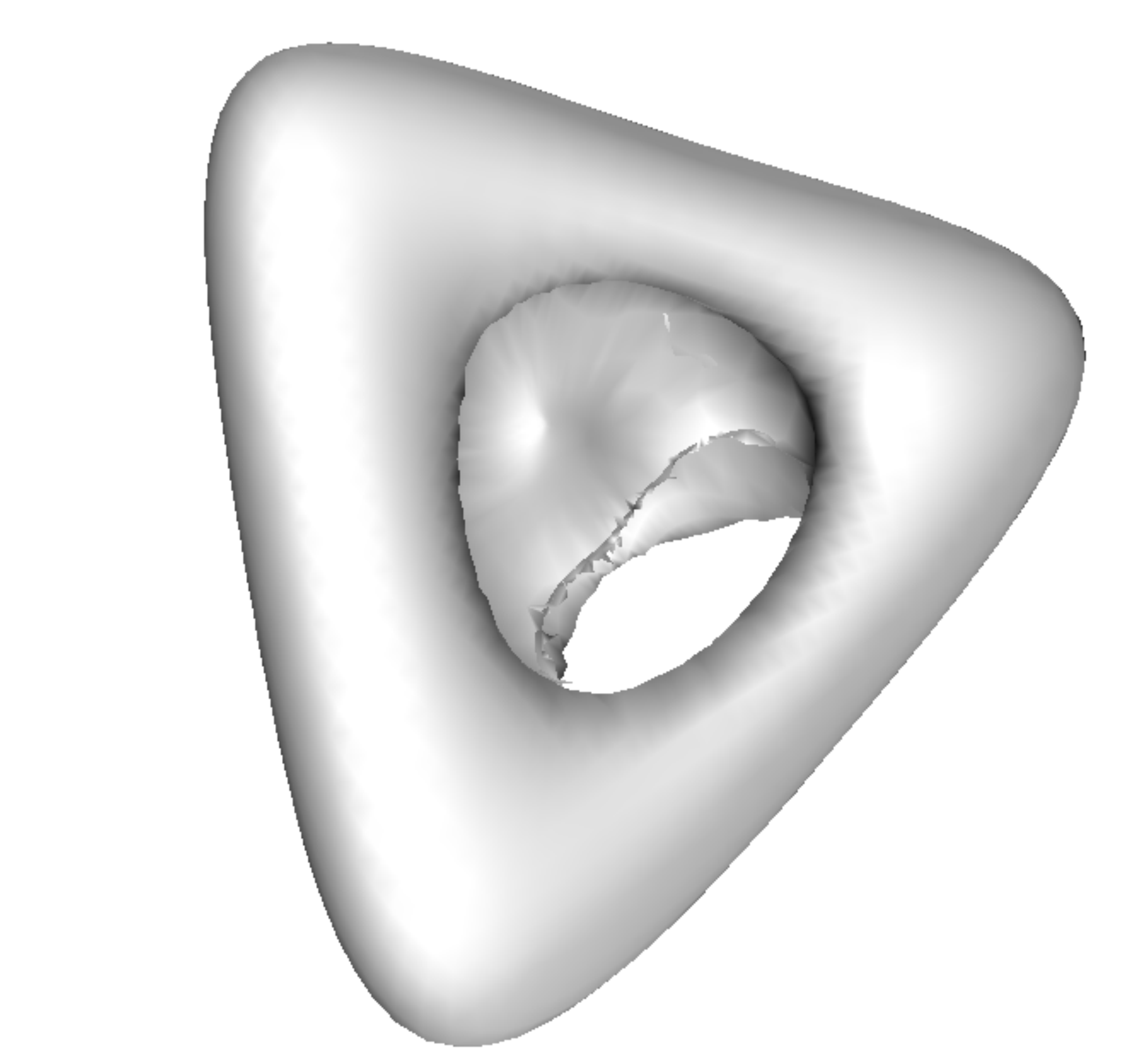} &
\includegraphicsVerifier{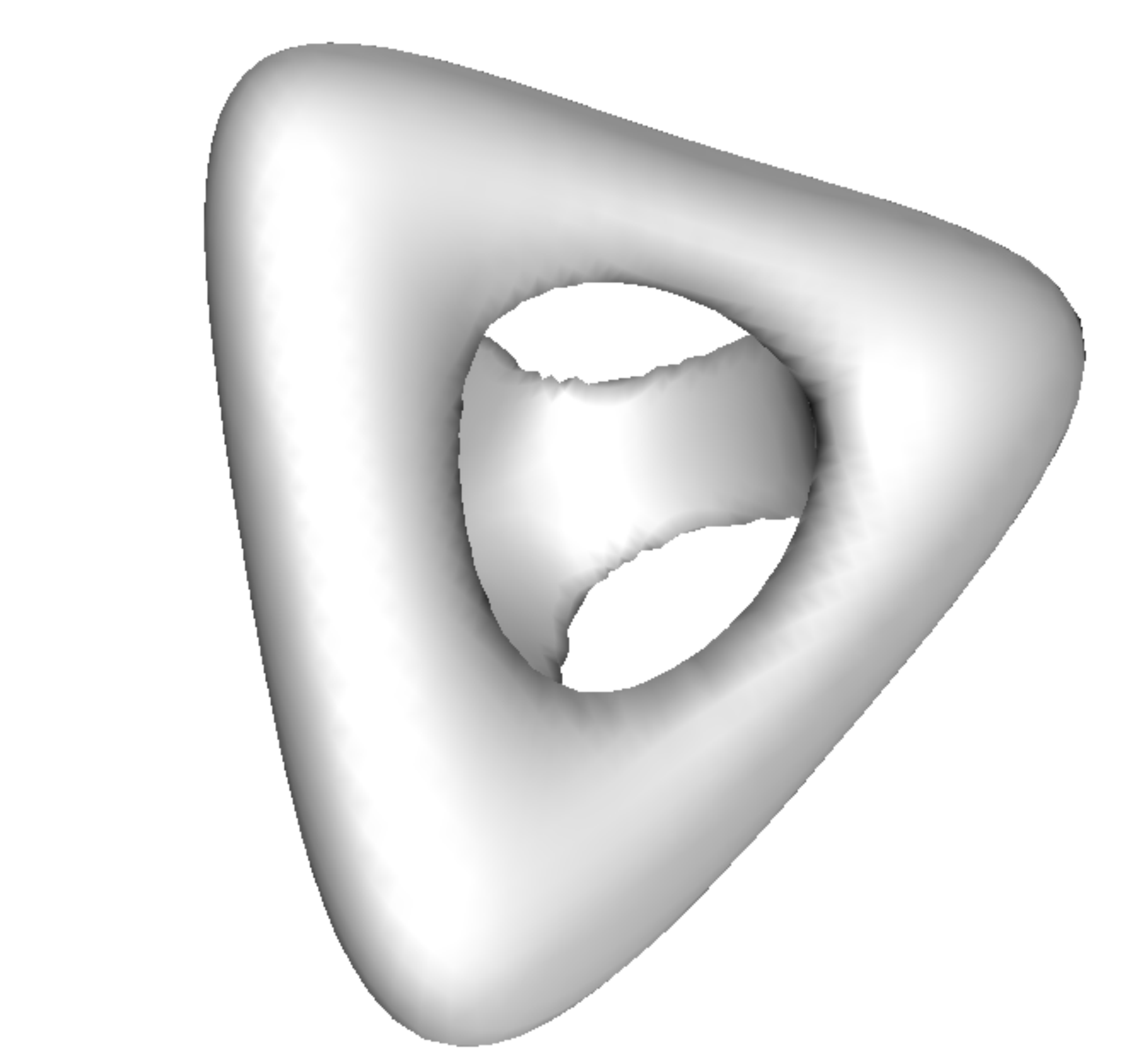} \\

\raisebox{1em}{\begin{sideways} Thoruses \end{sideways}} &
\includegraphicsVerifier{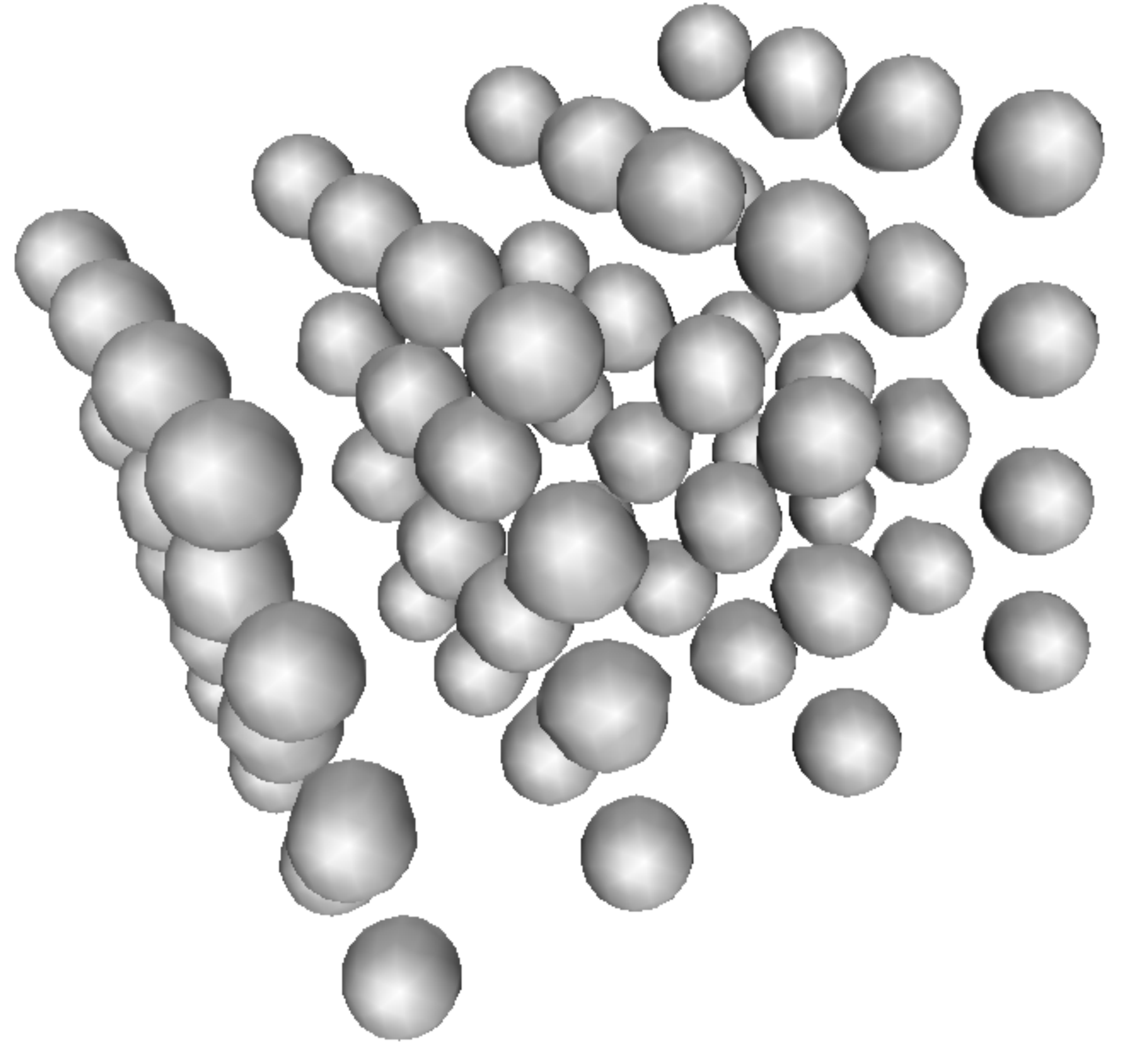} &
\includegraphicsVerifier{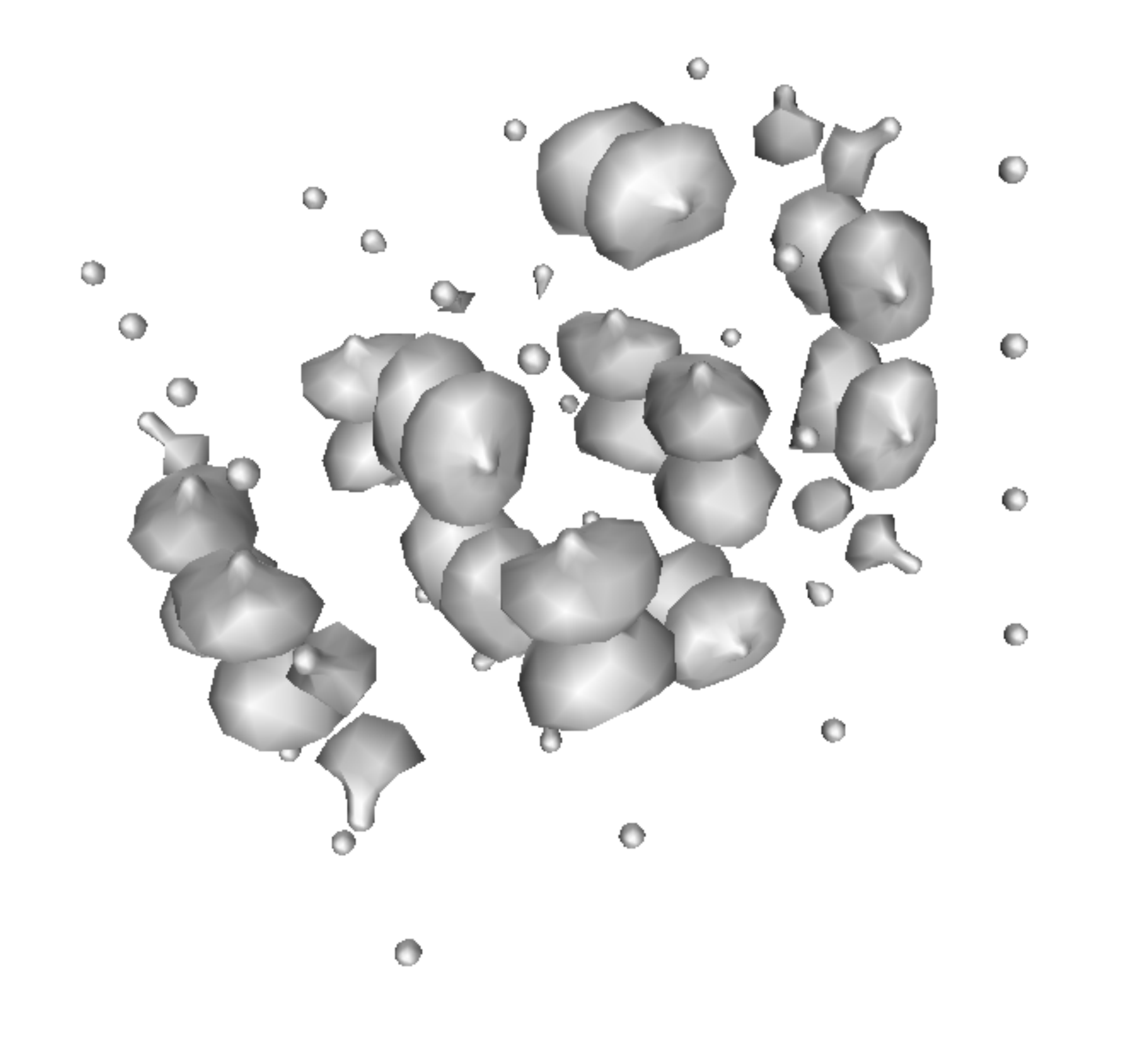} &
\includegraphicsVerifier{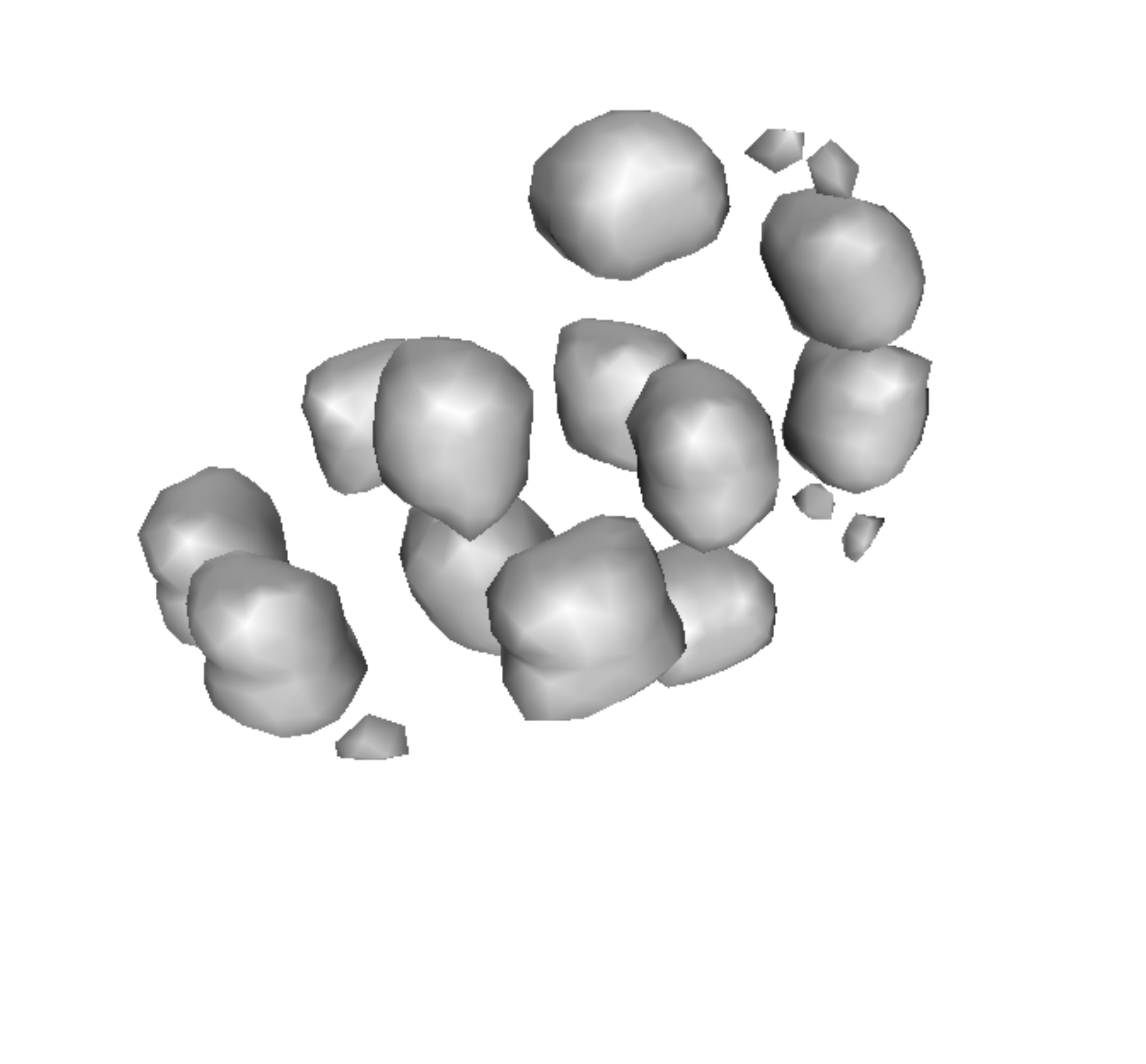} &
\includegraphicsVerifier{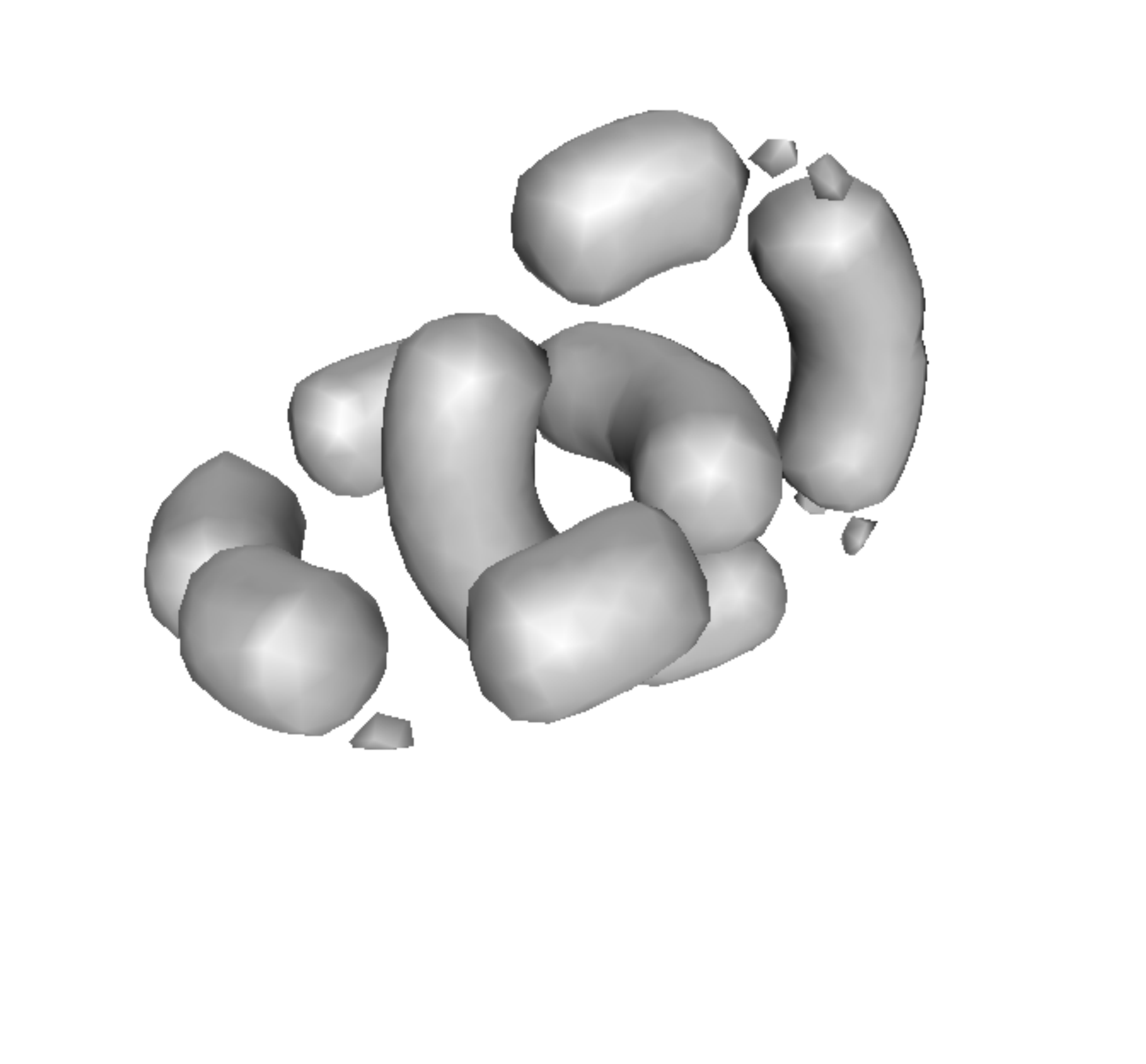} &
\includegraphicsVerifier{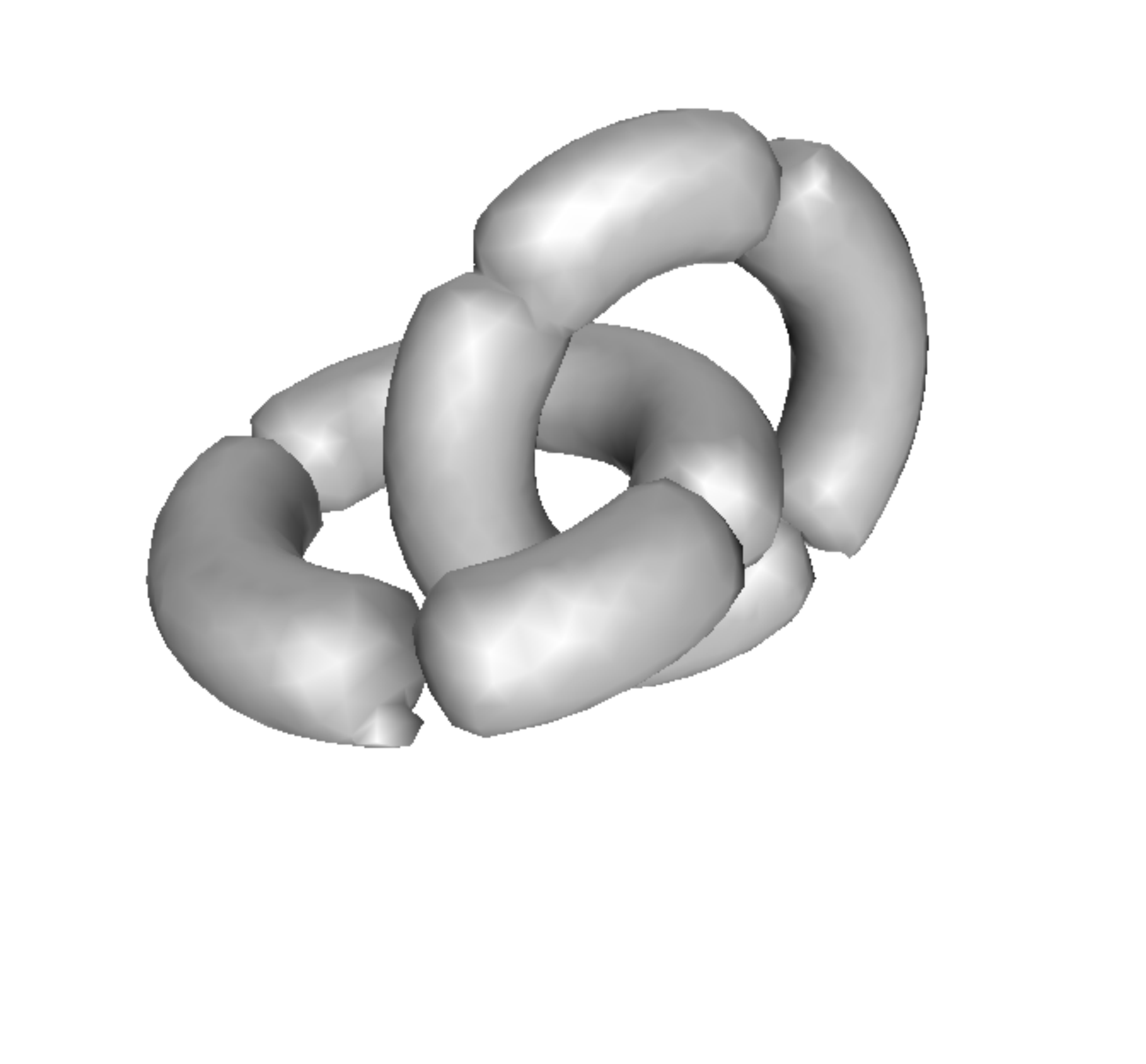} &
\includegraphicsVerifier{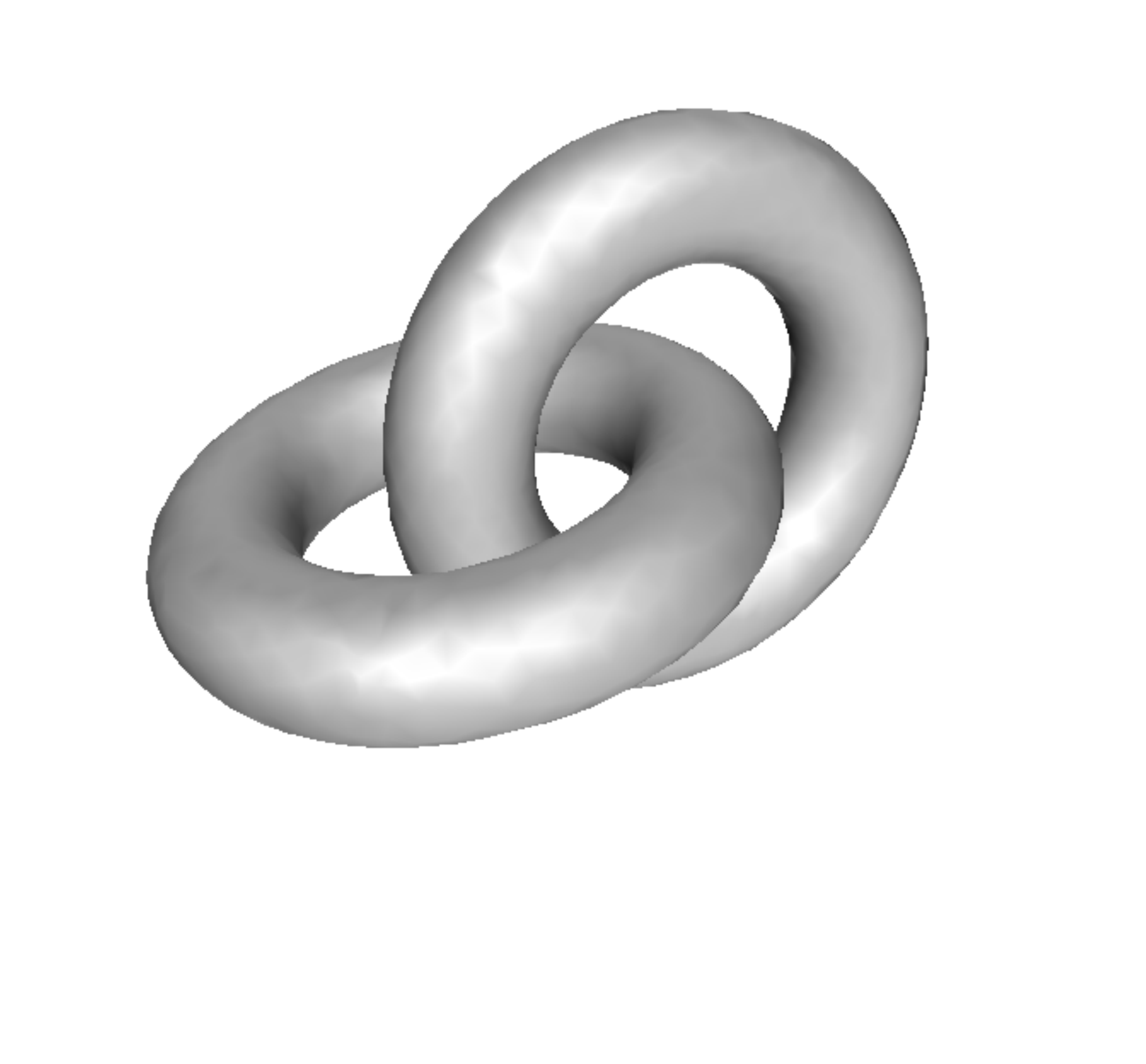} \\

\raisebox{1em}{\begin{sideways} Knots In \end{sideways}} &
\includegraphicsVerifier{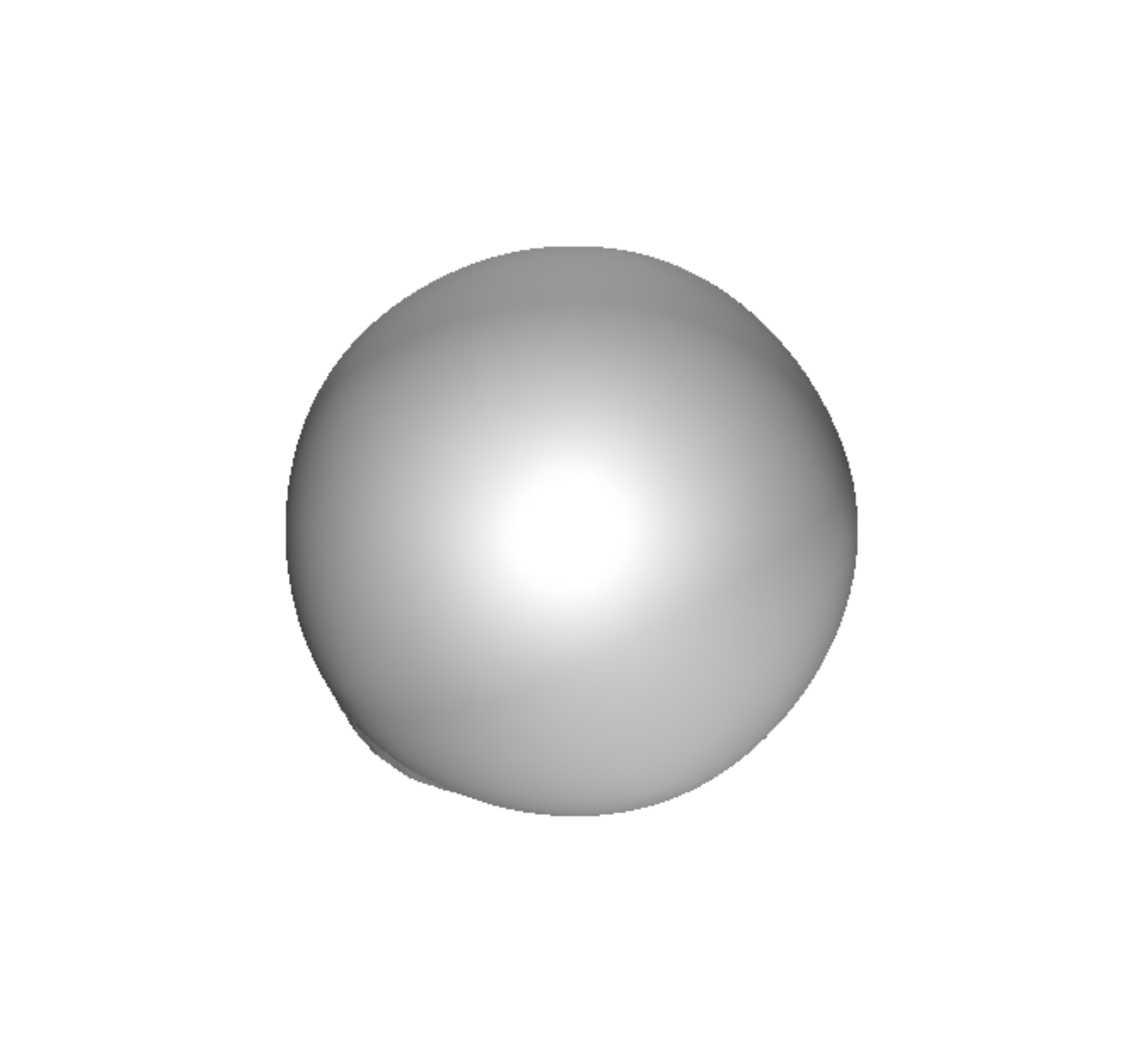} &
\includegraphicsVerifier{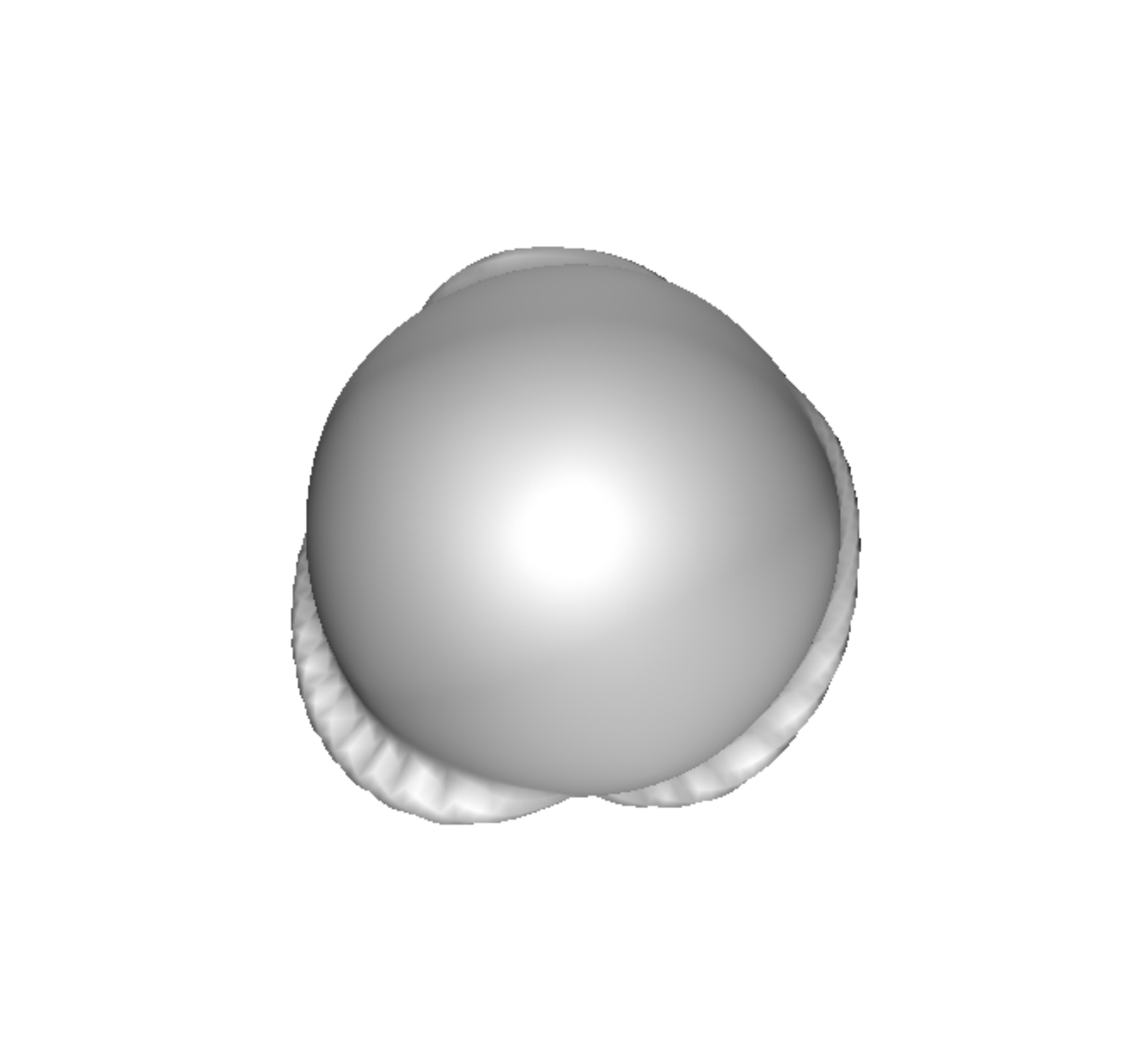} &
\includegraphicsVerifier{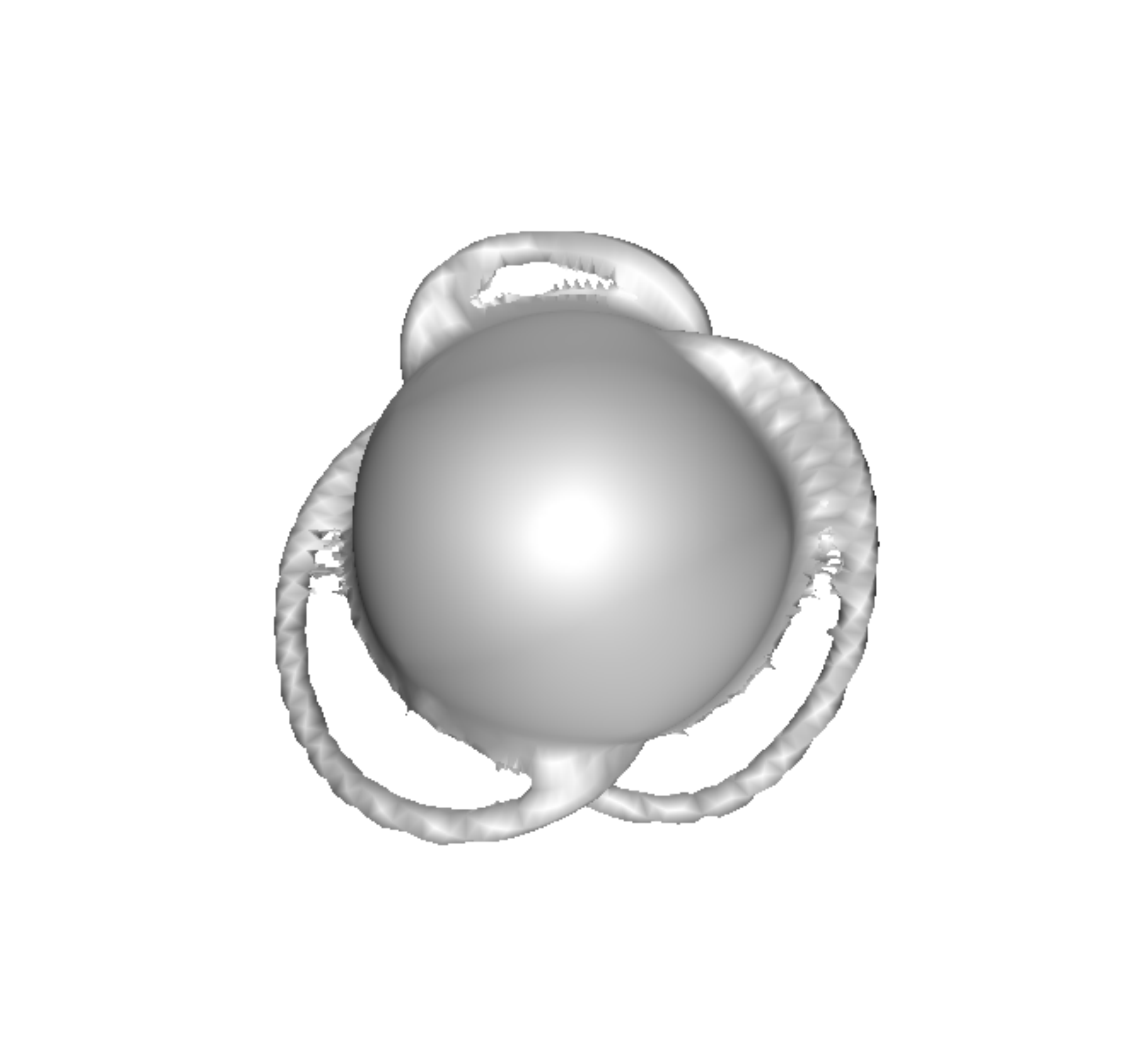} &
\includegraphicsVerifier{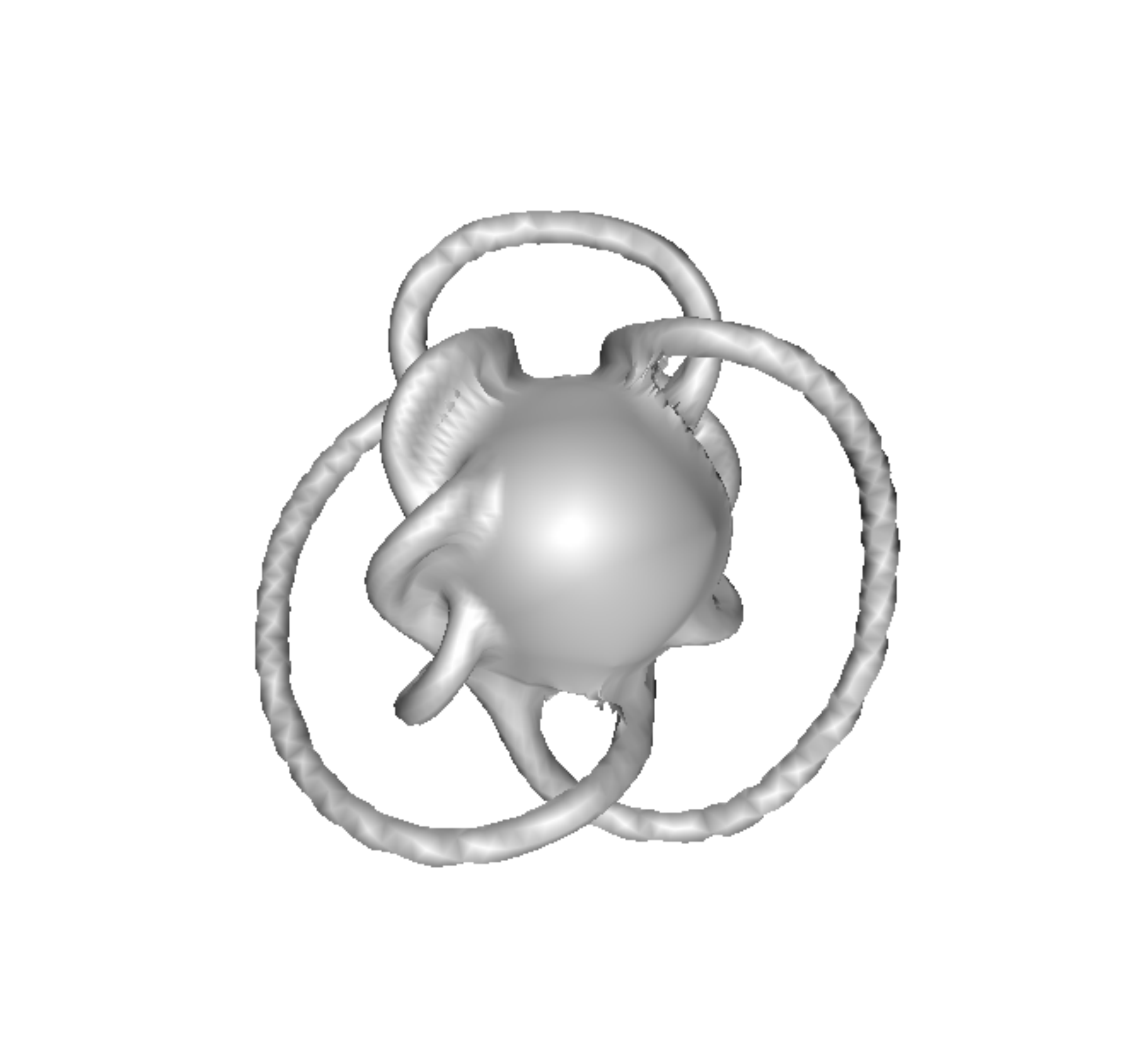} &
\includegraphicsVerifier{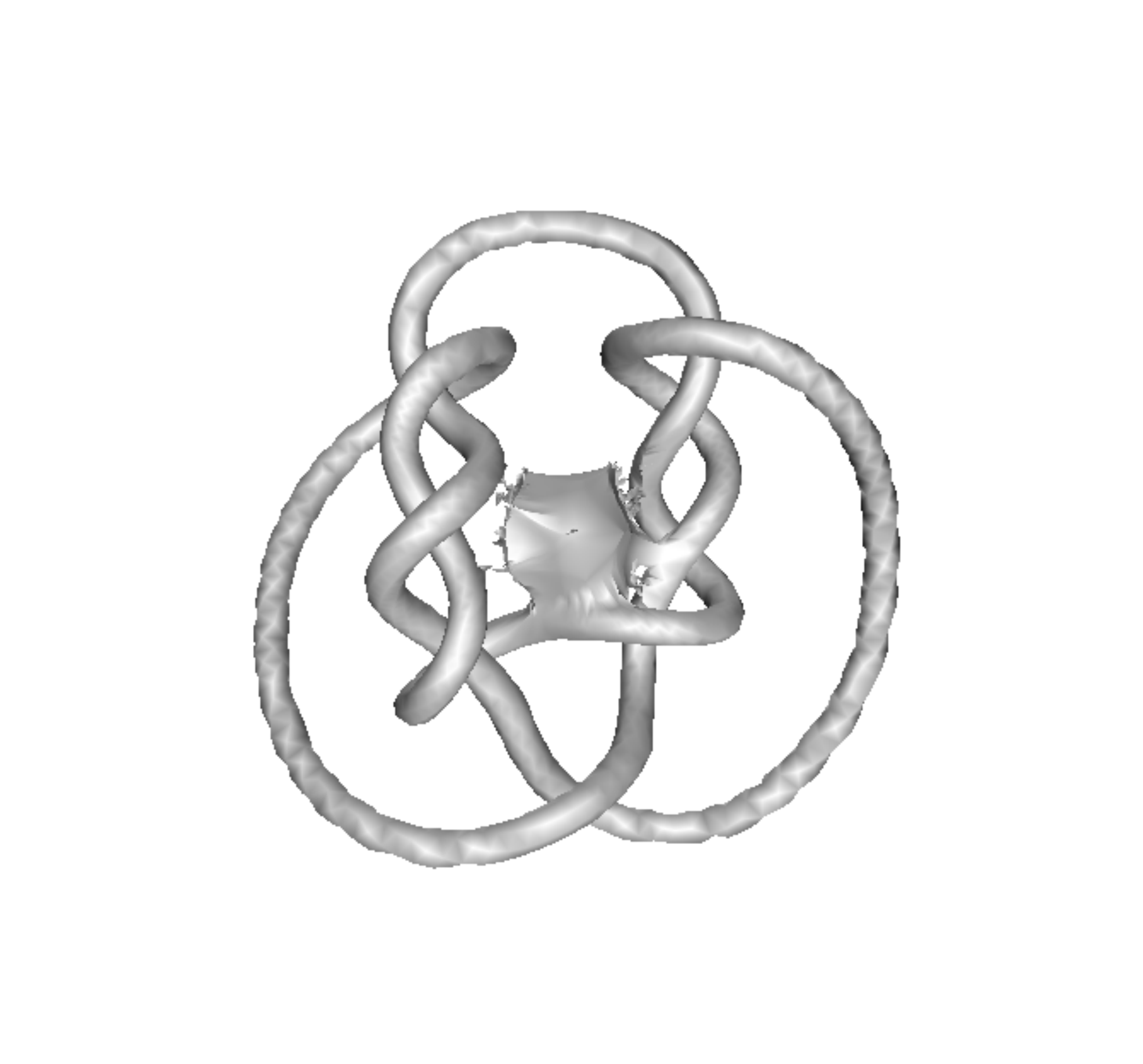} &
\includegraphicsVerifier{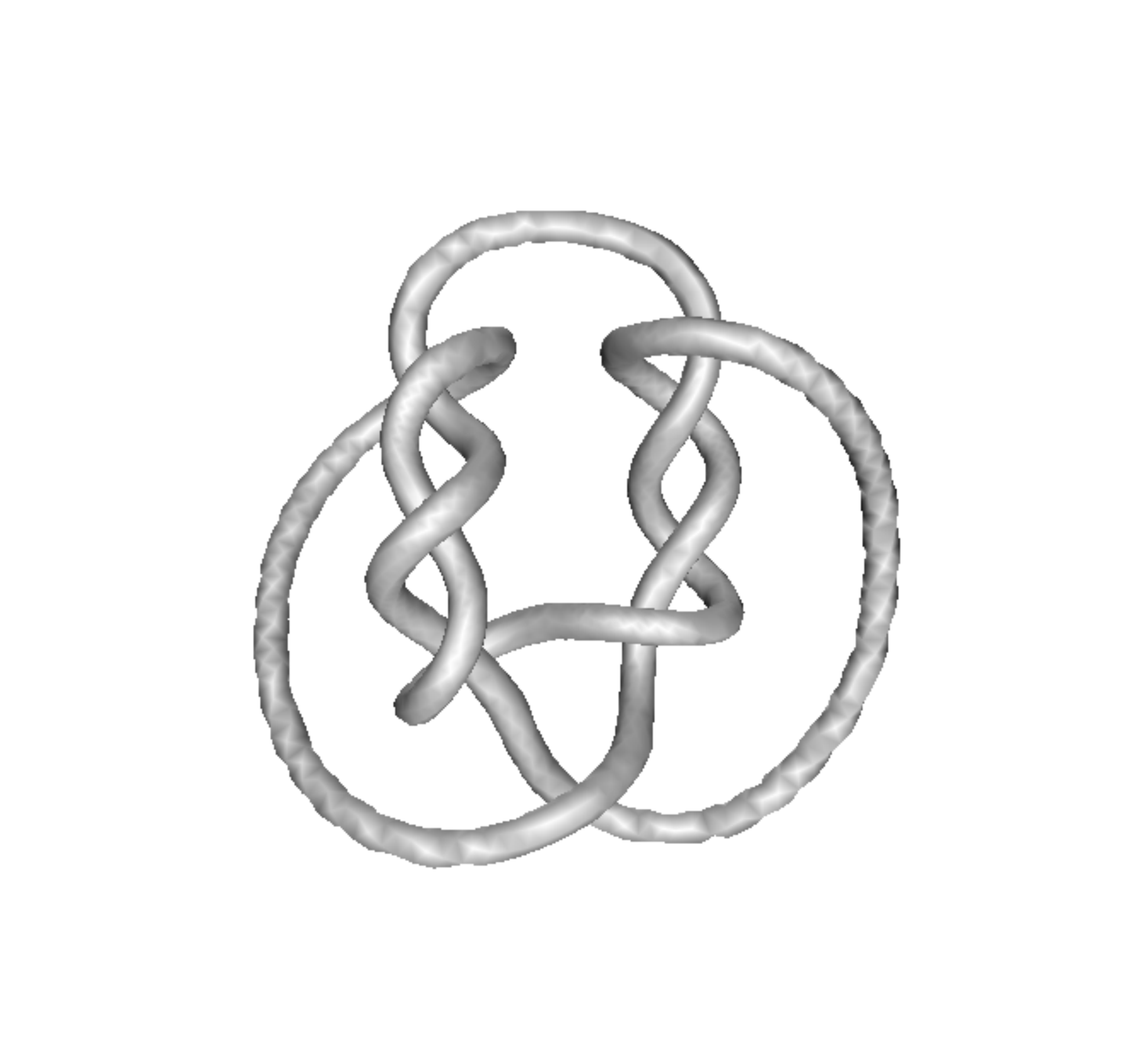} \\

\raisebox{1em}{\begin{sideways} Knots Out \end{sideways}} &
\includegraphicsVerifier{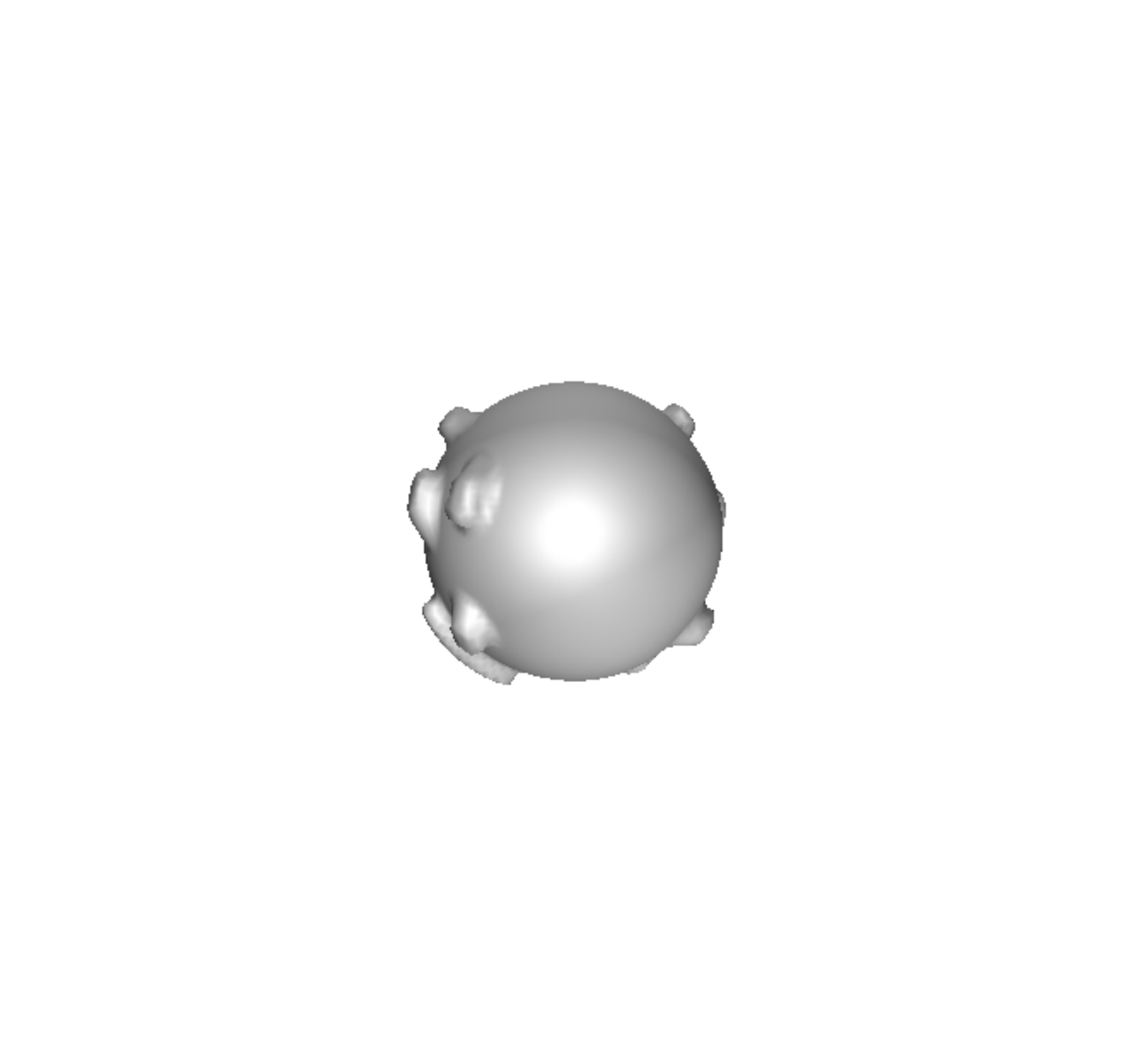} &
\includegraphicsVerifier{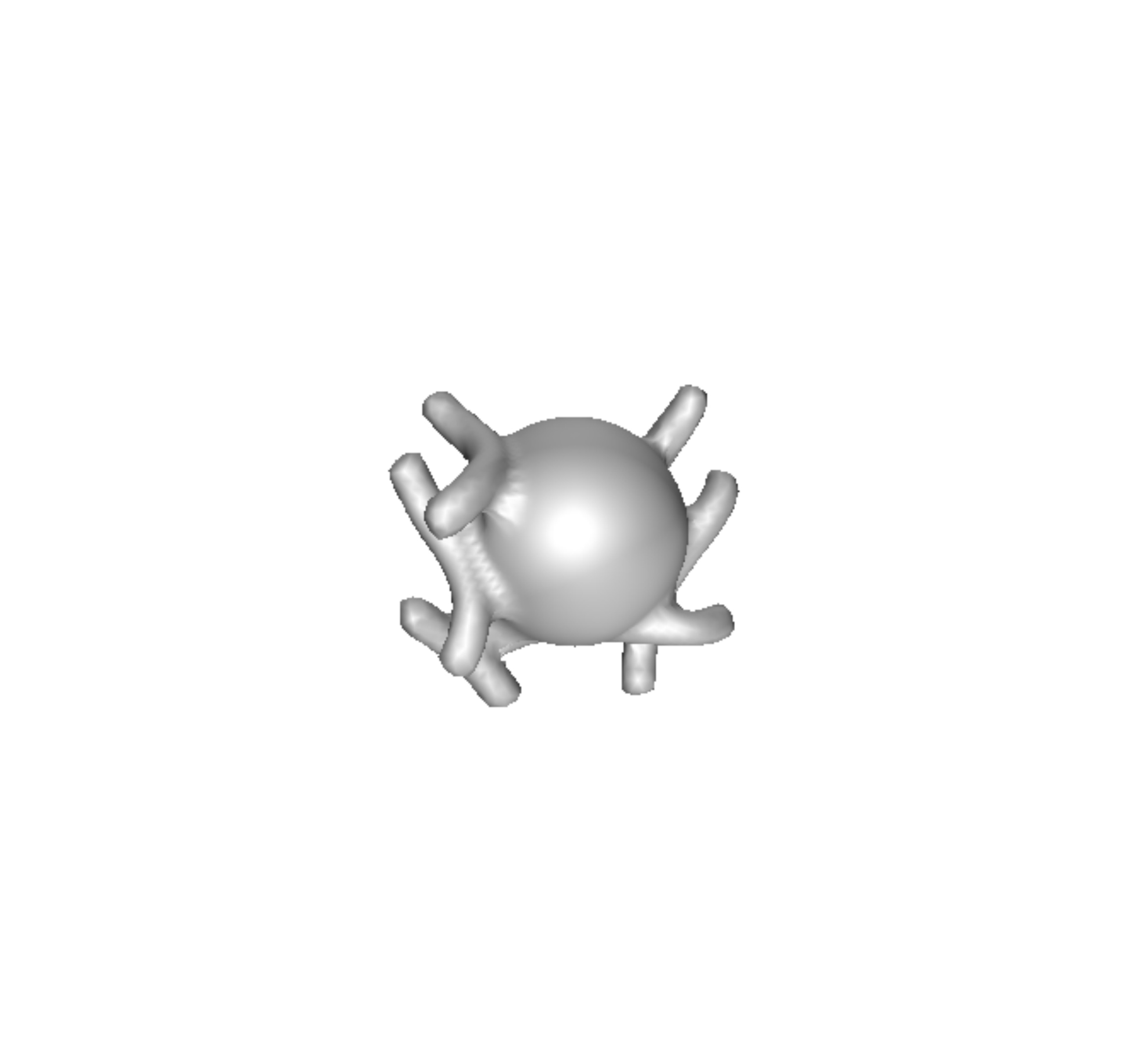} &
\includegraphicsVerifier{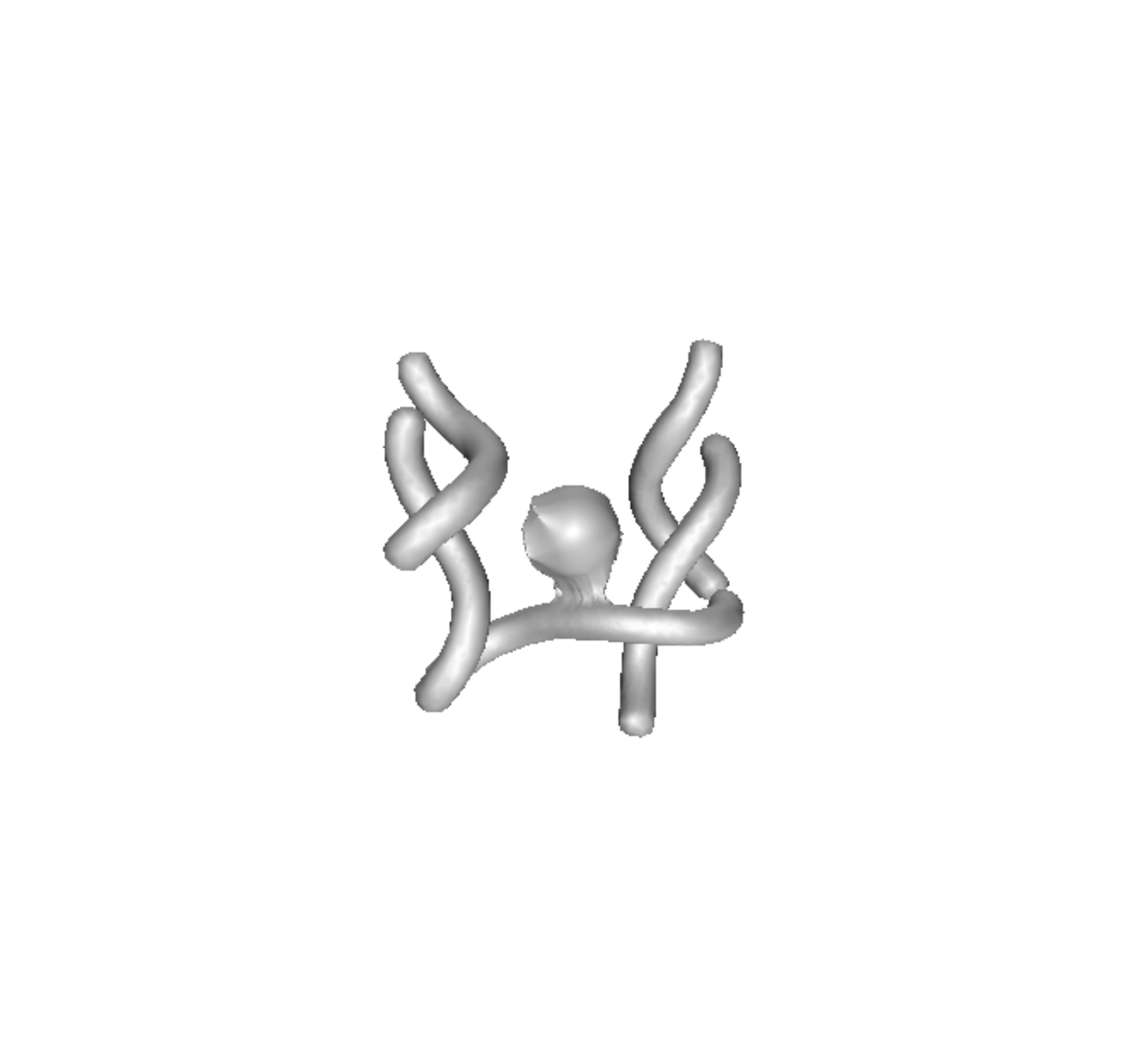} &
\includegraphicsVerifier{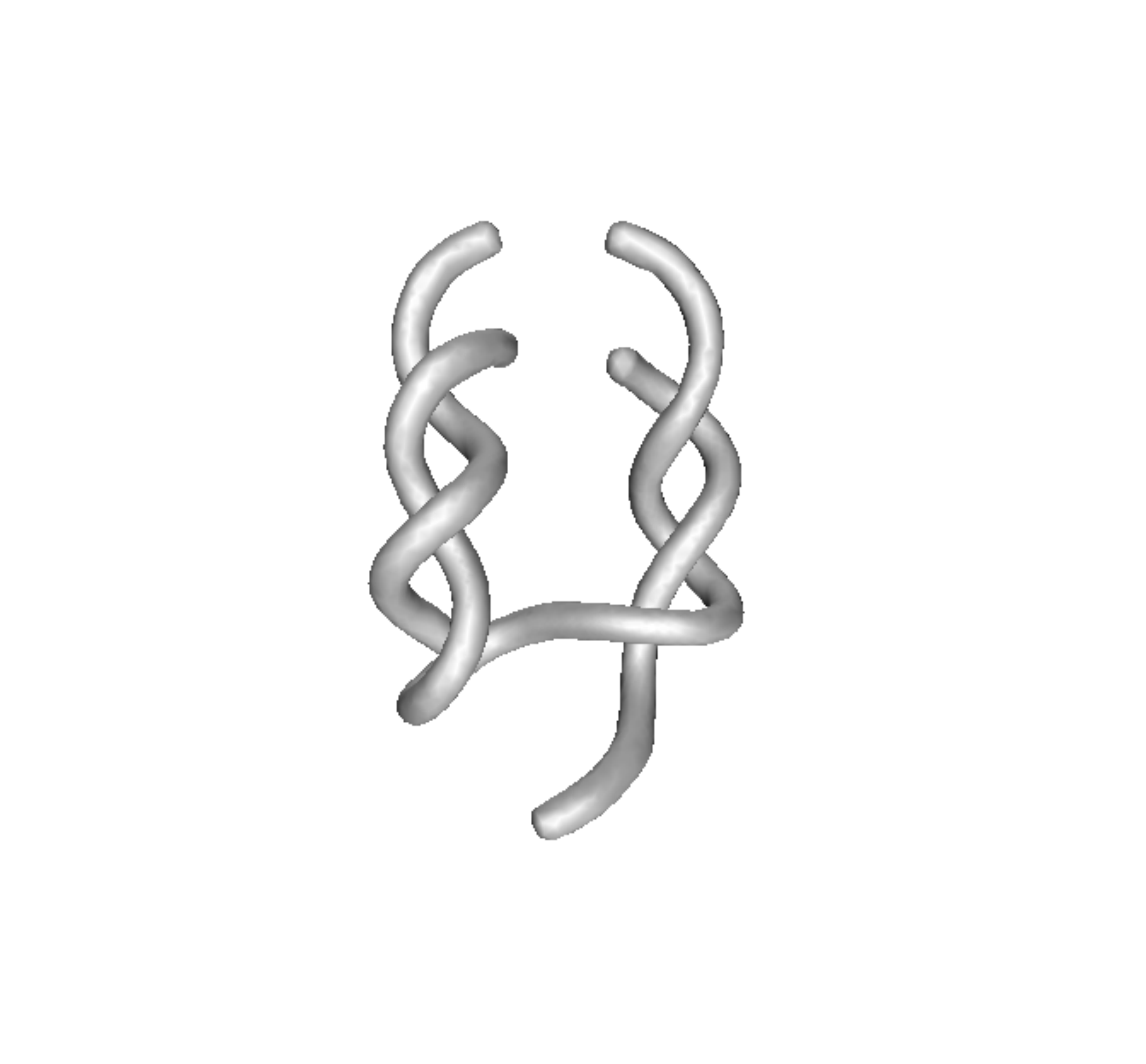} &
\includegraphicsVerifier{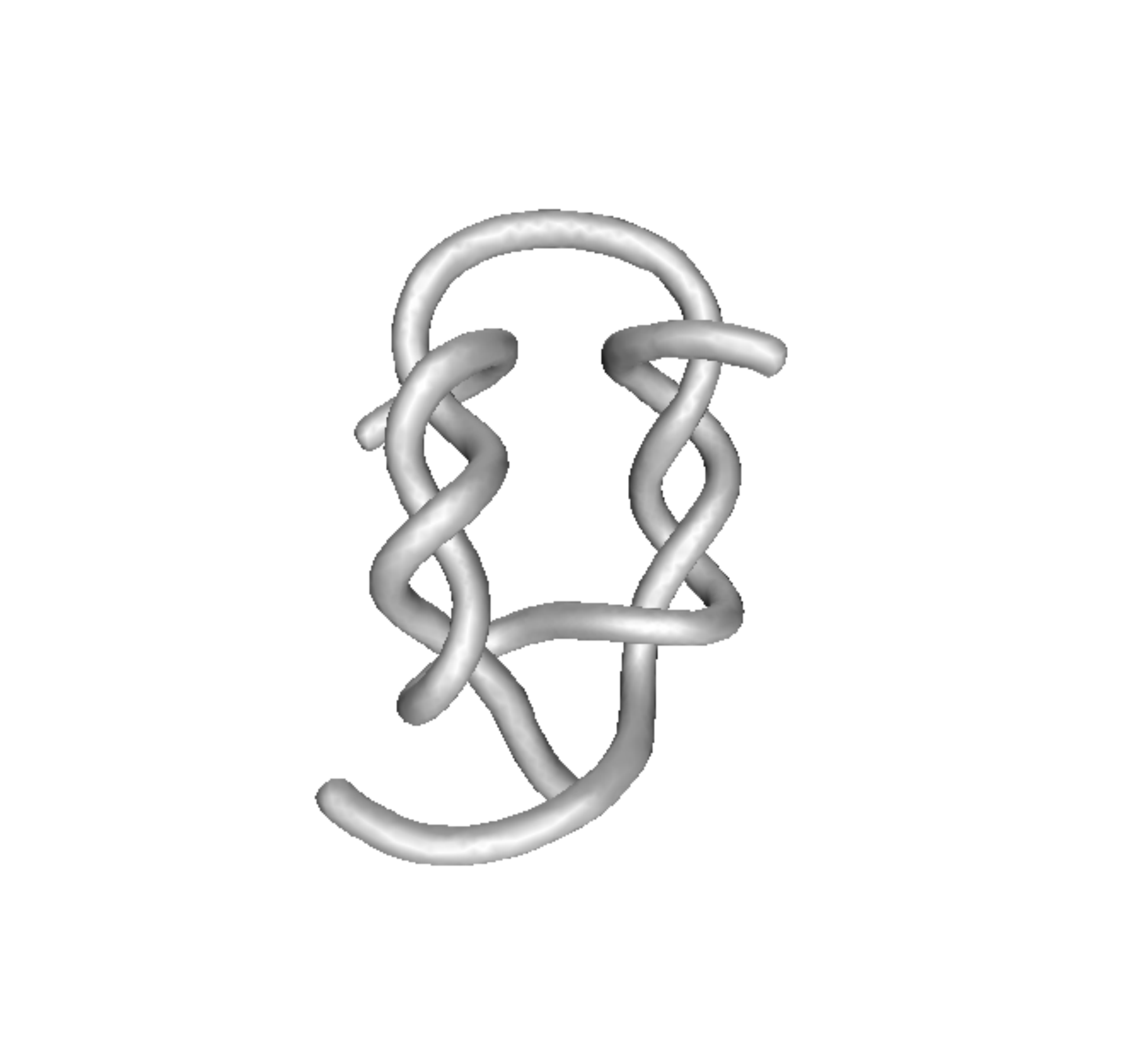} &
\includegraphicsVerifier{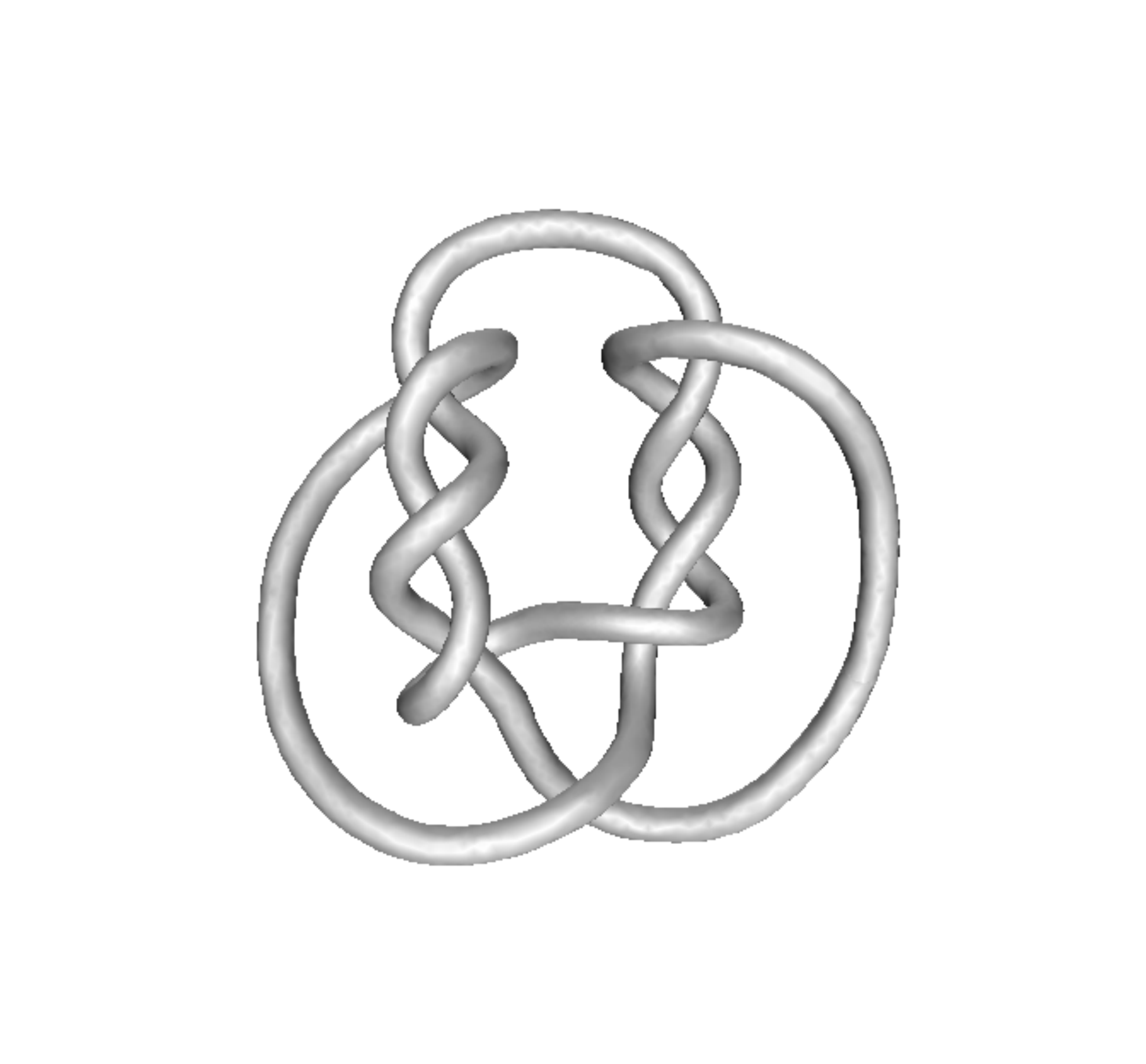} \\

\raisebox{1em}{\begin{sideways} Open Plane\end{sideways}} &
\includegraphicsVerifier{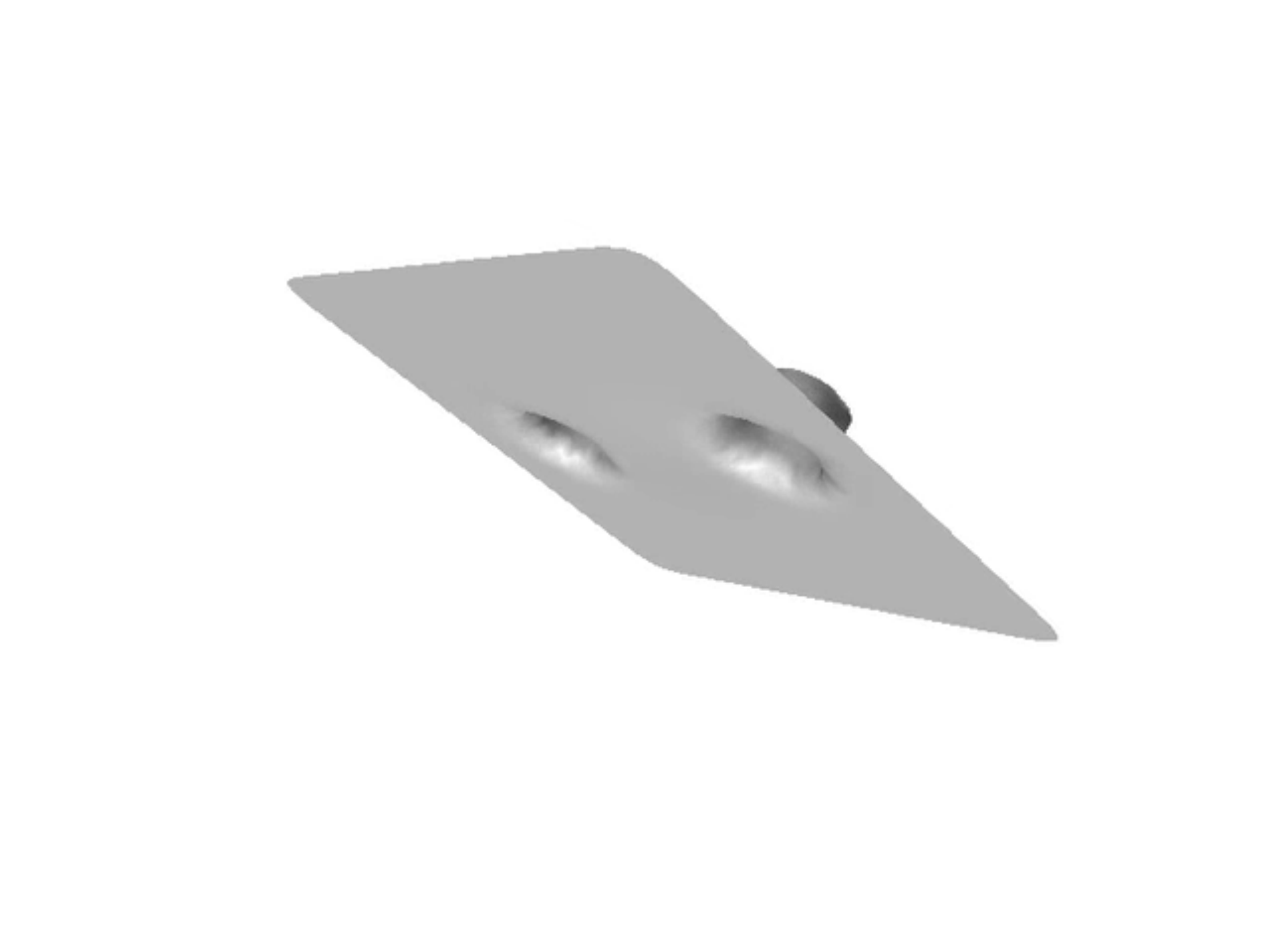} &
\includegraphicsVerifier{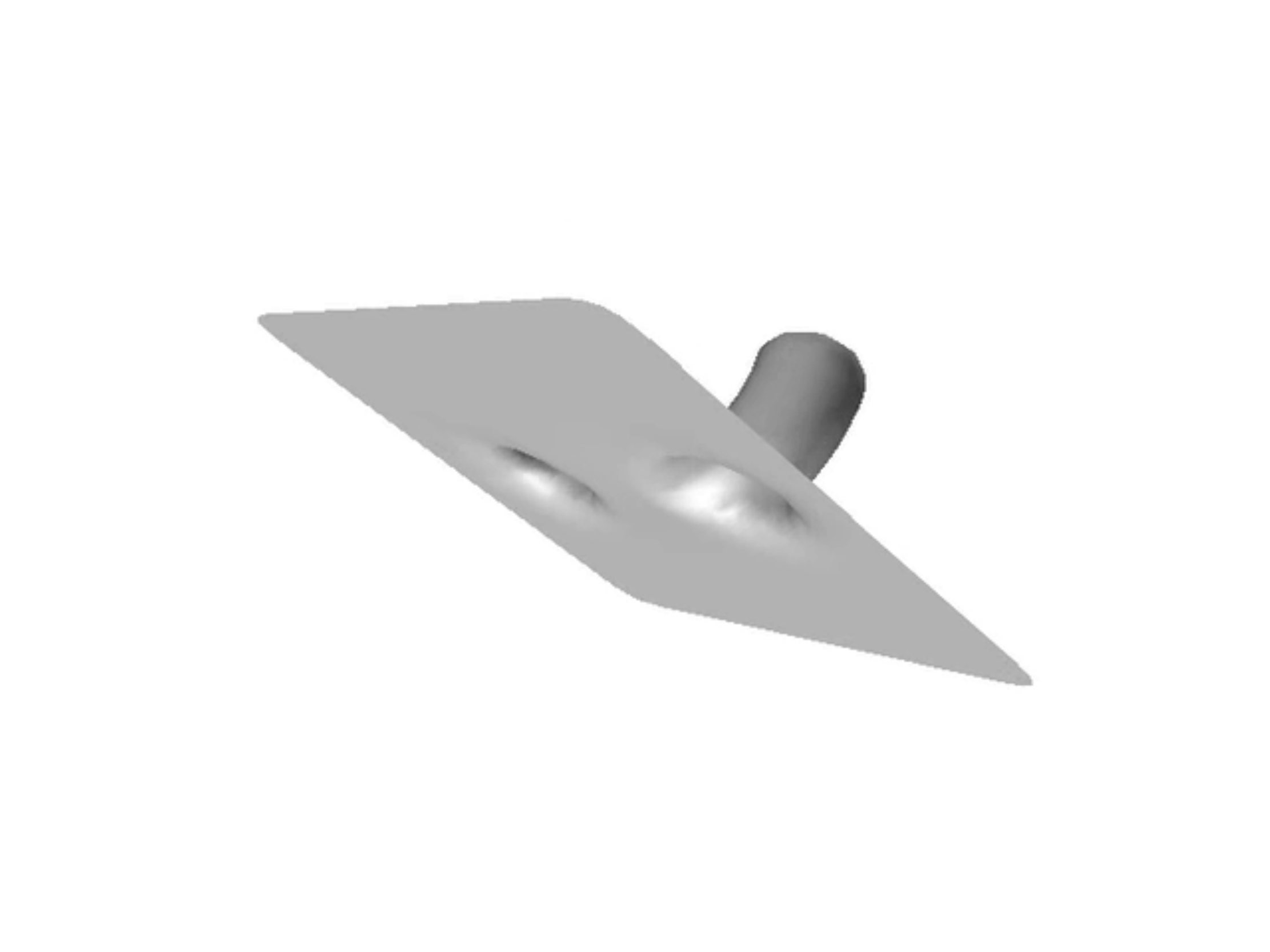} &
\includegraphicsVerifier{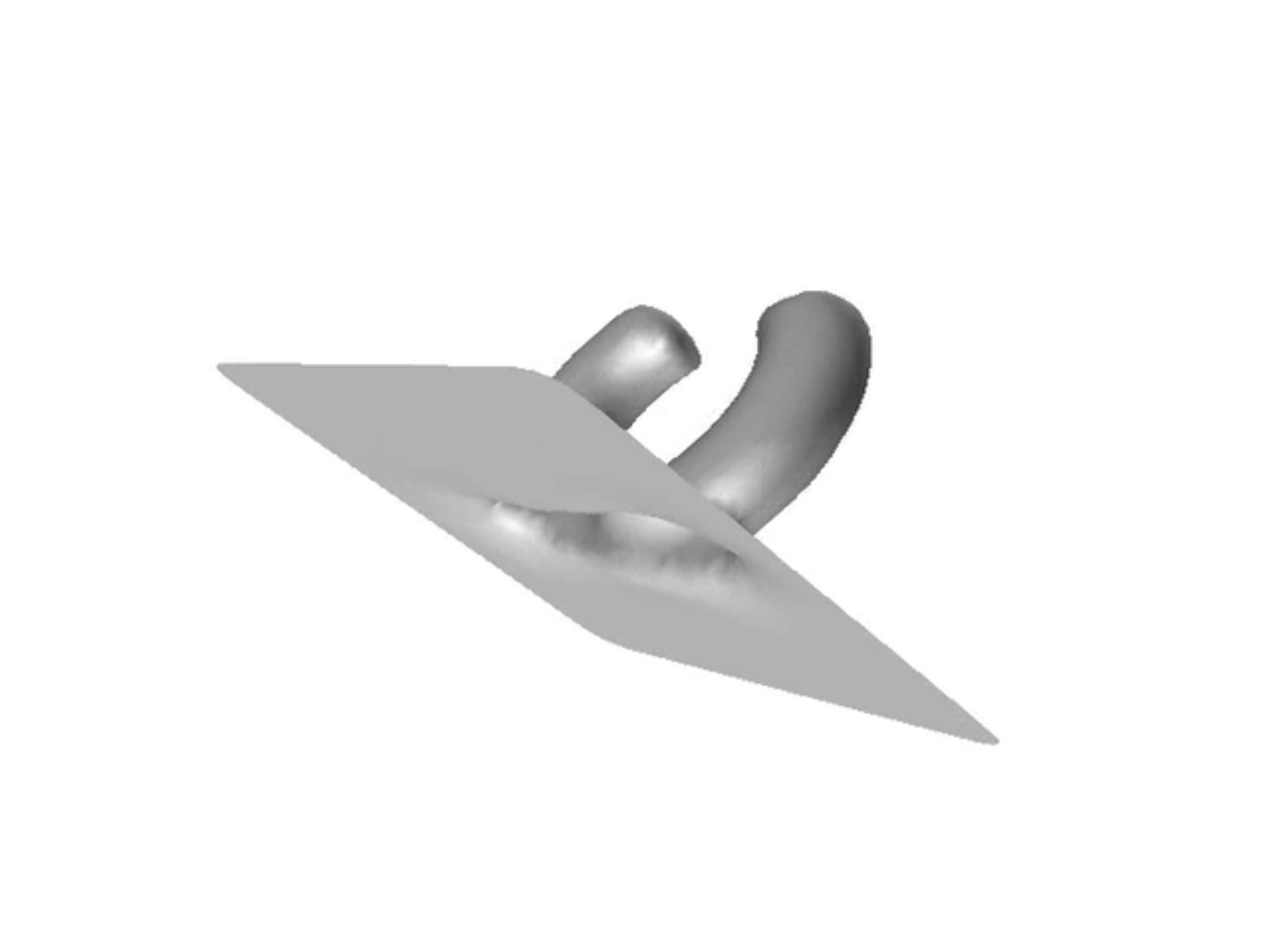} &
\includegraphicsVerifier{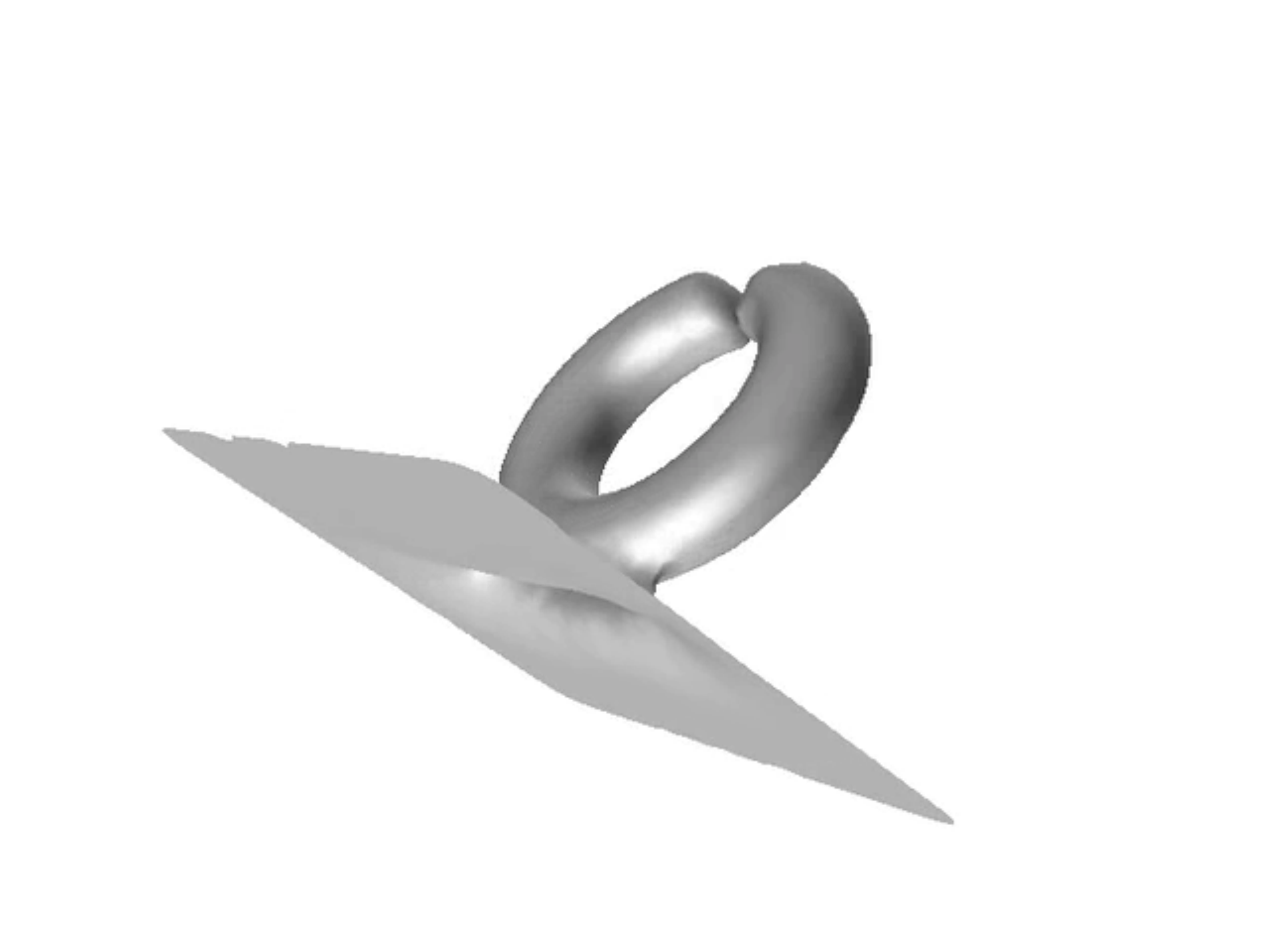} &
\includegraphicsVerifier{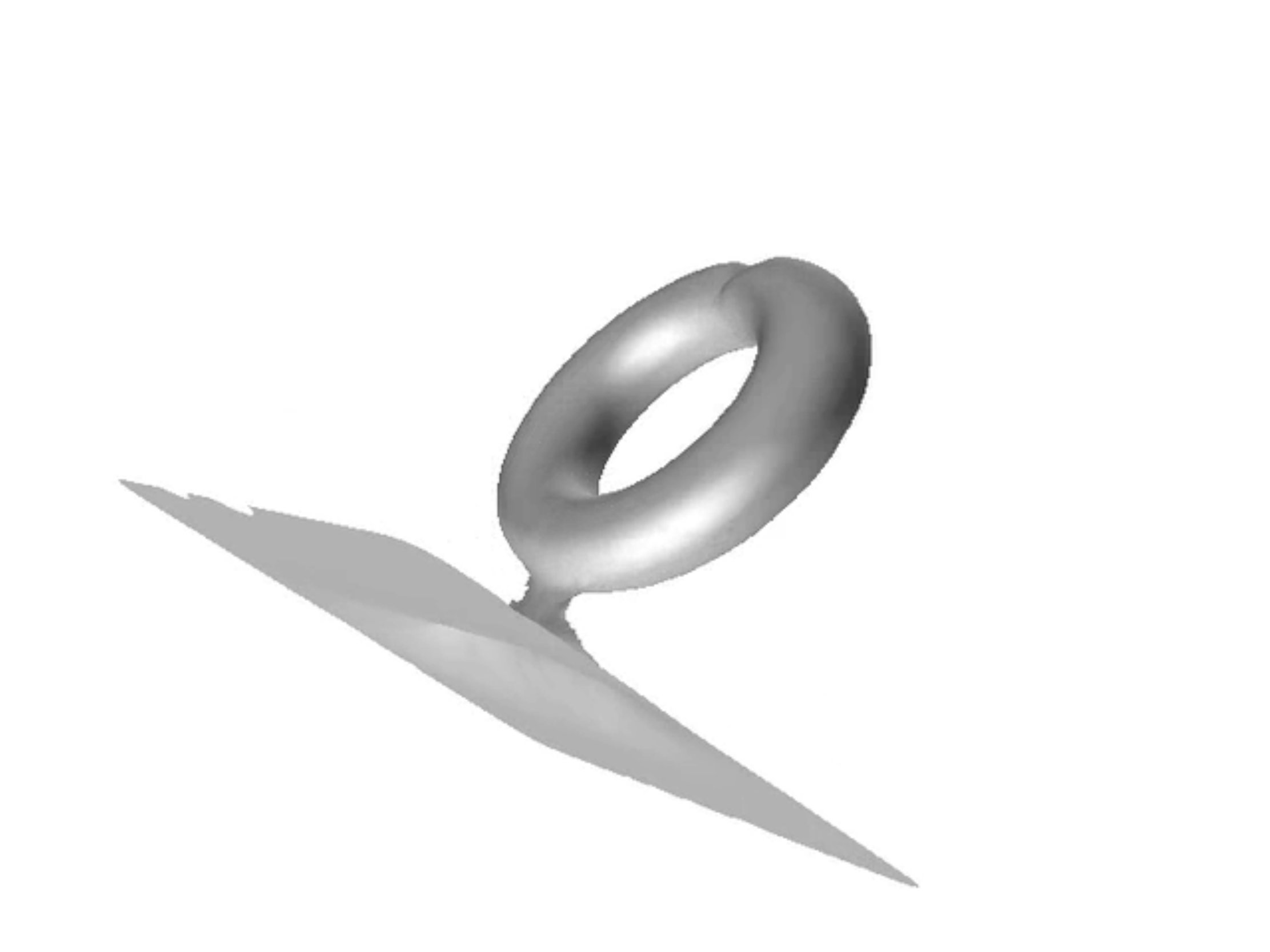} &
\includegraphicsVerifier{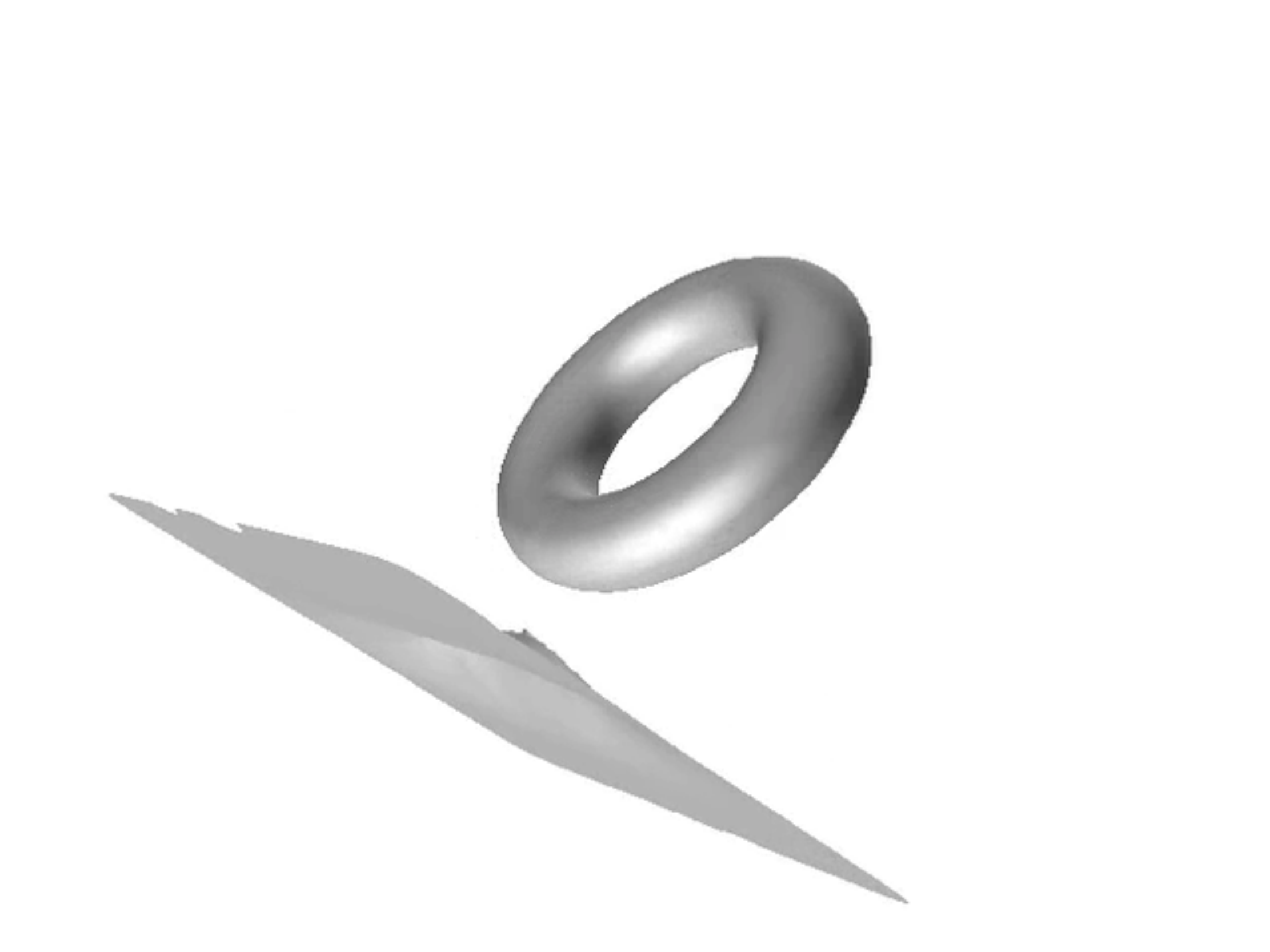} \\

\end{tabular}

\caption{Mesh morphing examples. Different steps for various test
  cases. Each row corresponds to a test case. The first column
  represents the first iteration, whereas the last column represents the
  last iteration.} 
\label{table:morph_evolutions}

\end{figure*}
\end{small}

\begin{tiny}
\begin{table*}[!htbp]
\centering
\begin{tabular}{rrrrrr}
\hline
Dataset & Genus 3 & Thoruses  & Knots In & Knots Out & Open Plane\\
\toprule
Iterations & 54 & 37 &  119 &  430 & 458\\
\# Facets & 4764.14 & 6296.33 & 13244.25 & 3873.11 & 6391.74\\
\# Intersections & 33.88 & 22.67 & 101.52 & 4.86 & 0.98\\
Time (TransforMesh) & 0.65 sec & 0.81 sec & 1.63 sec & 0.18 sec &  0.07 sec\\
Time (Total) & 1.42 sec & 1.78 sec & 3.58 sec & 0.89 sec & 0.17 sec\\
\hline
\end{tabular}

\caption{Mesh morphing statistics for different datasets. The reported values presented in the bottom four rows represent average values accumulated across the iterations. The running time is recorded on a 2.6 GHz Intel Core2Duo processor.}
\label{table:morph_stats}

\end{table*}
\end{tiny}

\begin{figure}[t]
\centering
\includegraphics[height=0.30\columnwidth]{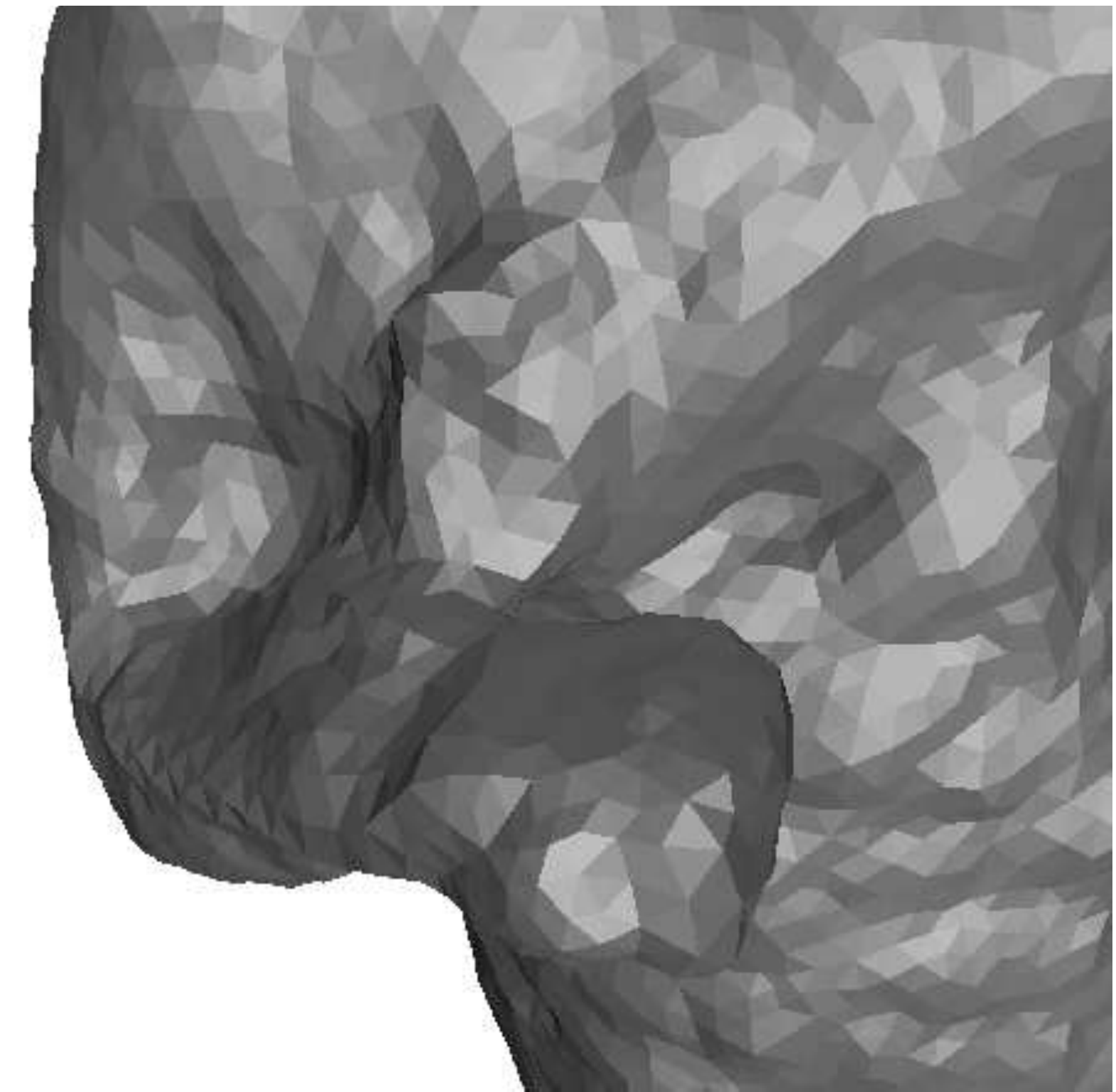} 
\includegraphics[height=0.30\columnwidth]{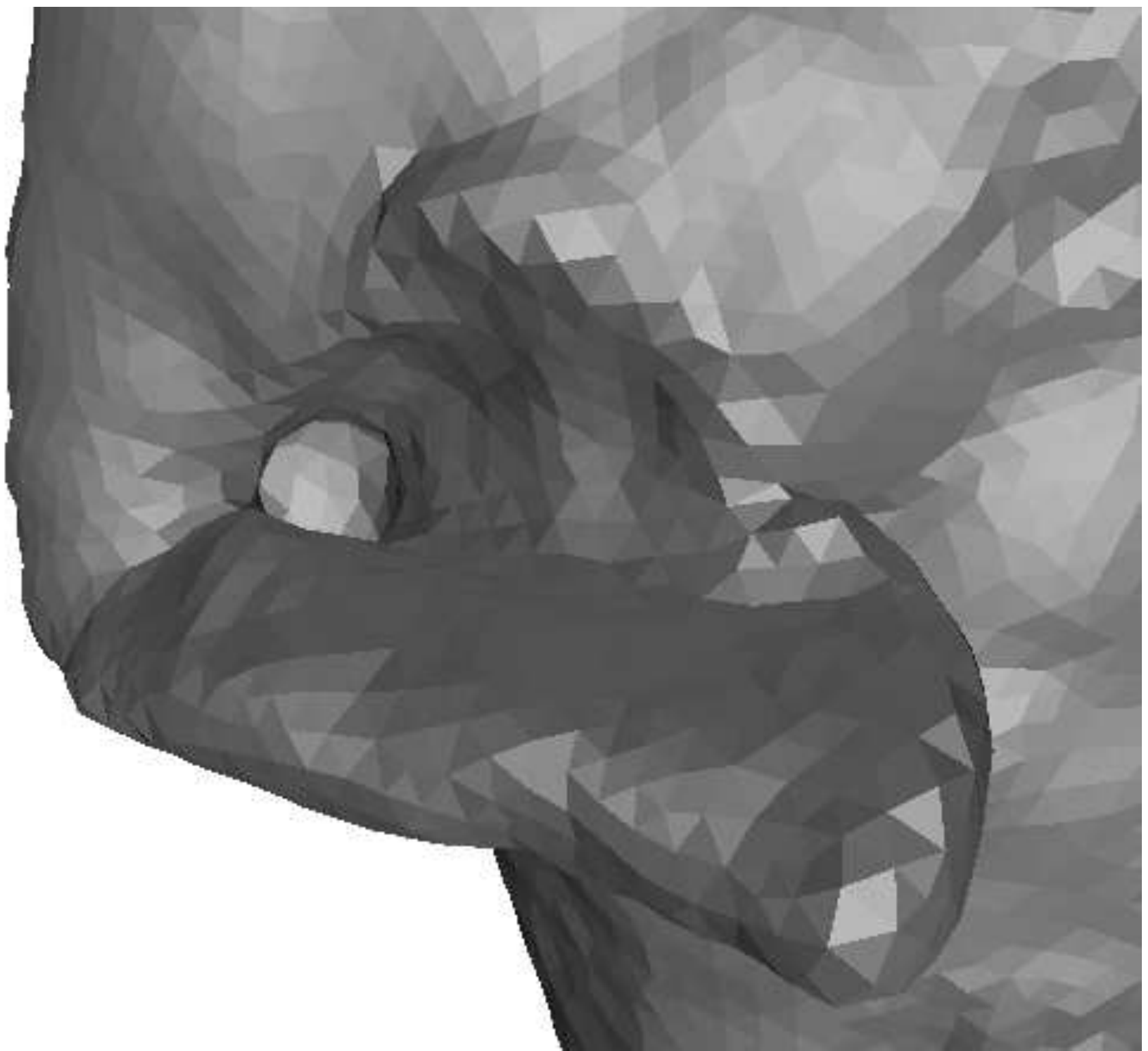} 
\includegraphics[height=0.30\columnwidth]{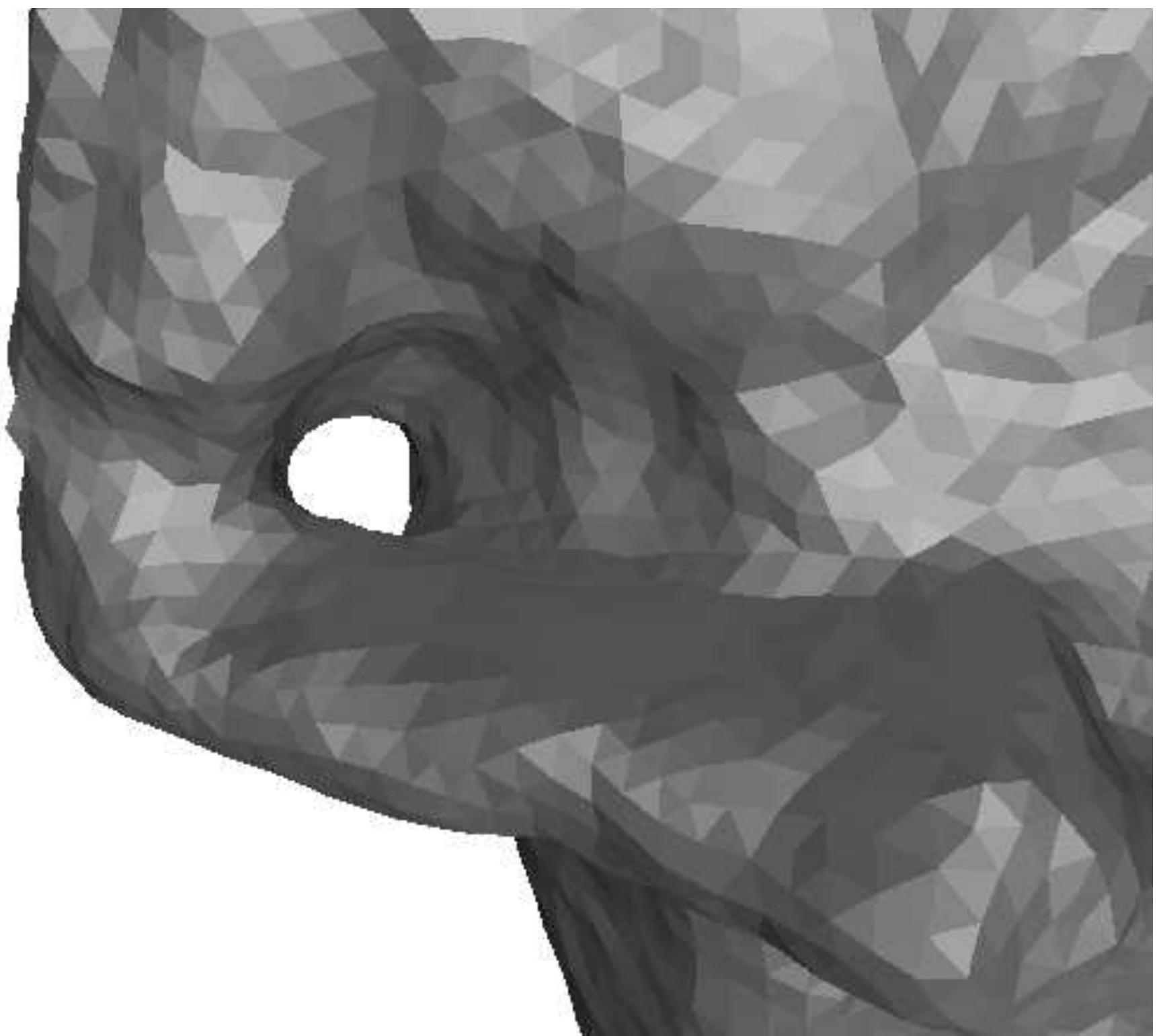} 
\caption{Example of topological changes during mesh morphing in surface tracking. The source surface $S'_t$, shown to the left, is the result of a deformation of the original mesh $S_t$ at time $t$ such that it matches closely the mesh $S_{t+1}$ at $t+1$, shown to the right. The mesh morphing process ensures the proper handling of topological changes (i.e. the whole formation in the arm region). }  
\label{fig:mesh-tracking-1}  
\end{figure}

Additional results of mesh morphing are presented in Figure \ref{fig:mesh-tracking-1}, with meshes obtained from 3-D reconstructions from multiple cameras, in the context of non-rigid surface tracking \cite{Varanasi:2008fr}.

\subsection{Multi-View 3-D Reconstruction}
\label{sec:app-reconstruction}

In this section we explain how our method fits into the
multi-view/image-based 3D reconstruction pipeline. The problem of
reconstructing  an object from images gathered with a large number of cameras has
received a lot of attention in the recent past
\cite{Hernandez:2004,Seitz:2006uq,Pons:2007,Vu:2009a}. It is
interesting to notice that, until recently, there were only a handful
of mesh-based solutions to the surface reconstruction problem. This is
mainly due to the topological problems raised by existing
mesh-evolution methods. In particular, topological-preserving
approaches are ill adapted to the problem of surface
reconstruction. Topological-adaptive algorithms, such as \MeshAlg{}
provide a more flexible solution that allows to better resolve for local 
details using topological changes. 

In \cite{Pons:2007} the multi-view reconstruction problem is cast into
an energy-minimization problem using photometric constraints.
It is well known that topological changes may take place during the
minimization process, e.g., Figure
\ref{fig:rec_from_points_closeup}. Surface evolution based on a level-set formulation is proposed
in  \cite{Pons:2007}. Our contribution to this class of reconstruction
methods is to extend such surface
evolution approaches to meshes that allow to focus on the shape's surface instead of
a bounding volume.

The method described below was applied both to visual hulls, e.g.,
\cite{Franco:2003} and to sparse point-based 3-D data, e.g.,
\cite{Zaharescu:2009tg}. The former representation constitutes the initial
mesh that needs be improved using photometric information from the
available images. The latter representation can be easily turned into
a rough mesh using \cite{Amenta:2001} for example. 

\subsubsection{Methodology}

The initial meshed surface corresponds to an extended bounding box
obtained using image silhouettes and a geometric approach that
involves cone intersections in 3-D, i.e.,
\cite{FrancoBoyer09}. Such a mesh is only a
coarse approximation of the observed surface. One main limitation of
visual hull approaches is that they do not recover concave
regions. The initial surface can be improved by considering
photometric information in the images. The underlying principle is
that, with a correct geometry, and under the Lambertian surface
assumption, the mesh should be photo-consistent, i.e., its projections
in the images should have similar photometric information
\cite{seitz1999}. 

The photometric constraints are casted into an energy minimization
framework, using a similarity measure between pairs of cameras that are
close to each other, as proposed by Pons {\it et
  al.}~\cite{Pons:2007}. The problem is solved in practice via
gradient 
descent.  $E_{img}$ is the derivative of the local photoconsistency
term, in the normal direction, that can be computed using several
methods. To compute such a derivative, we use one of  the most
efficient approaches \cite{Pons:2007},
based on the normalized cross-correlation. The evolution equation is
in this case: 

\begin{equation}
\mathcal{\vec F}_{reconstruction} = \frac{\partial S}{\partial t}  =  E_{img}(x) \mathbf{N}(x).
\label{eq:surface_evolution}
\end{equation}

In \cite{Pons:2007} the surface evolution is
implemented within the level-set framework. We extended it to 
meshes using the \MeshAlg{}  algorithm. The level-set solution
performs surface evolution using a coarse-to-fine approach in order to
escape from local minima.  Traditionally, in level-set approaches, the
implicit function that embeds the surface $S$ is discretized evenly on
a 3-D grid. As a side-effect, all the facets of the recovered surface
are of a maximum size, set by the discretization grid cell. In contrast, mesh based
approaches do not impose such a constraint and allow facets of all
sizes on the evolving surface. This is particularly useful when
starting from rough surface estimates, such as visual hulls, where the initial mesh contains
triangles of all dimensions. In addition, the dimension of visual
facets appears to be a relevant information since regions where the
visual reconstruction is less accurate, i.e. concave regions on the
observed surface, are described by bigger facets on the visual
hull. Thus, we adopt an approach in which  bigger
triangles are processed first, until they are stabilized, then the whole
process is repeated at a finer scale.  

The mesh evolution algorithm depicted in  Figure
\ref{fig:generic_alg_layout} requires a number of parameters to be set
in advance. In the case of 3-D reconstruction we used the following
parameter settings in all our examples:
$t=0.001$ for the time step, $\alpha=0.1$ and
$e_{avg}$ for the maximum movement amplitude and $\beta=0.1$ for the
smoothing term. The meshes have an adaptive mesh resolution. As
mentioned earlier, we ran the algorithm at different scales, starting
from scale $s_{max}$ to $s_{min}=1$ in  $\lambda=\sqrt{2}$
decrements. For each scale $s_i$, the input images and camera matrices
are downscaled accordingly. The appropriate edge size interval is set
to $e_{1}=edgeSize(1,1)$ $e_{2i}=edgeSize(5,i)$, where
$edgeSize(p_1,p_2)$ is a function that computes the desired edge size
such that it has $p_1$ pixels using images at scales $p_2$. The
initial scale $s_{max}$ is computed such that the largest edges of the
initial mesh measure 5 pixels when projected into the images at scale
$s_{max}$. When the finer scale is reached, new iterations are run by
decreasing $e_2$ from $edgeSize(5,1)$ to $edgeSize(2,1)$ in
$\lambda=\sqrt 2$ decrements. 

\begin{tiny}
\begin{table*}[!htbp]
\centering
\begin{tabular}{l r r r r r r r r }
\hline
\multirow{2}{*}{ }  & \multicolumn{2}{c}{Temple Ring} & \multicolumn{2}{c}{Temple Sparse Ring} & \multicolumn{2}{c}{Dino Ring} & \multicolumn{2}{c}{Dino Sparse Ring} \\
 \cline{2-9}
& Acc. & Compl.& Acc. & Compl.& Acc. & Compl.& Acc. & Compl.\\
\toprule
Pons {\it et al.}~\cite{Pons:2007} & 0.60mm  & 99.5\% & 0.90mm  & 95.4\% & 0.55mm &  99.0\% & 0.71mm &  97.7\% \\

Furukawa and Ponce~\cite{Furukawa:2009pami} & 0.47mm  & 99.6\% & $\mathbf{0.63mm}$  & $\mathbf{99.3\%}$ & $\mathbf{0.28mm}$ &  $\mathbf{99.8\%}$ & $\mathbf{0.37mm}$ &  $\mathbf{99.2}$\% \\

Hernandez and Schmitt~\cite{Hernandez:2004} & 0.52mm  & 99.5\% & 0.75mm  & 95.3\% &  0.45mm &  97.9\% & 0.60mm &  98.52\% \\

Vu et al.\cite{Vu:2009a} & $\mathbf{0.45mm}$  & $\mathbf{99.8\%}$ &   &  &  0.53mm &  99.7\% &  &   \\

\MeshAlg{} & 0.55mm & 99.2\% &  0.78mm & 95.8\% &  0.42mm & 98.6\% &  0.45mm & $\mathbf{99.2}$\%\\
\hline
\end{tabular}

\caption{Middleburry 3-D Reconstruction Results. Accuracy: the distance d in mm that brings 90\% of the result R within the ground-truth surface G. Completeness: the percentage of G that lies within 1.25mm of R.} 
\label{table:results}
\end{table*}
\end{tiny}

\subsubsection{Results}
\label{sec:results}

\begin{figure*}[!htbp]
\centering
\begin{tabular}{ c c c  c}
\hline
Dino & Temple & Box & Dinosaur \\
\toprule
\includegraphics[width=0.16\textwidth]{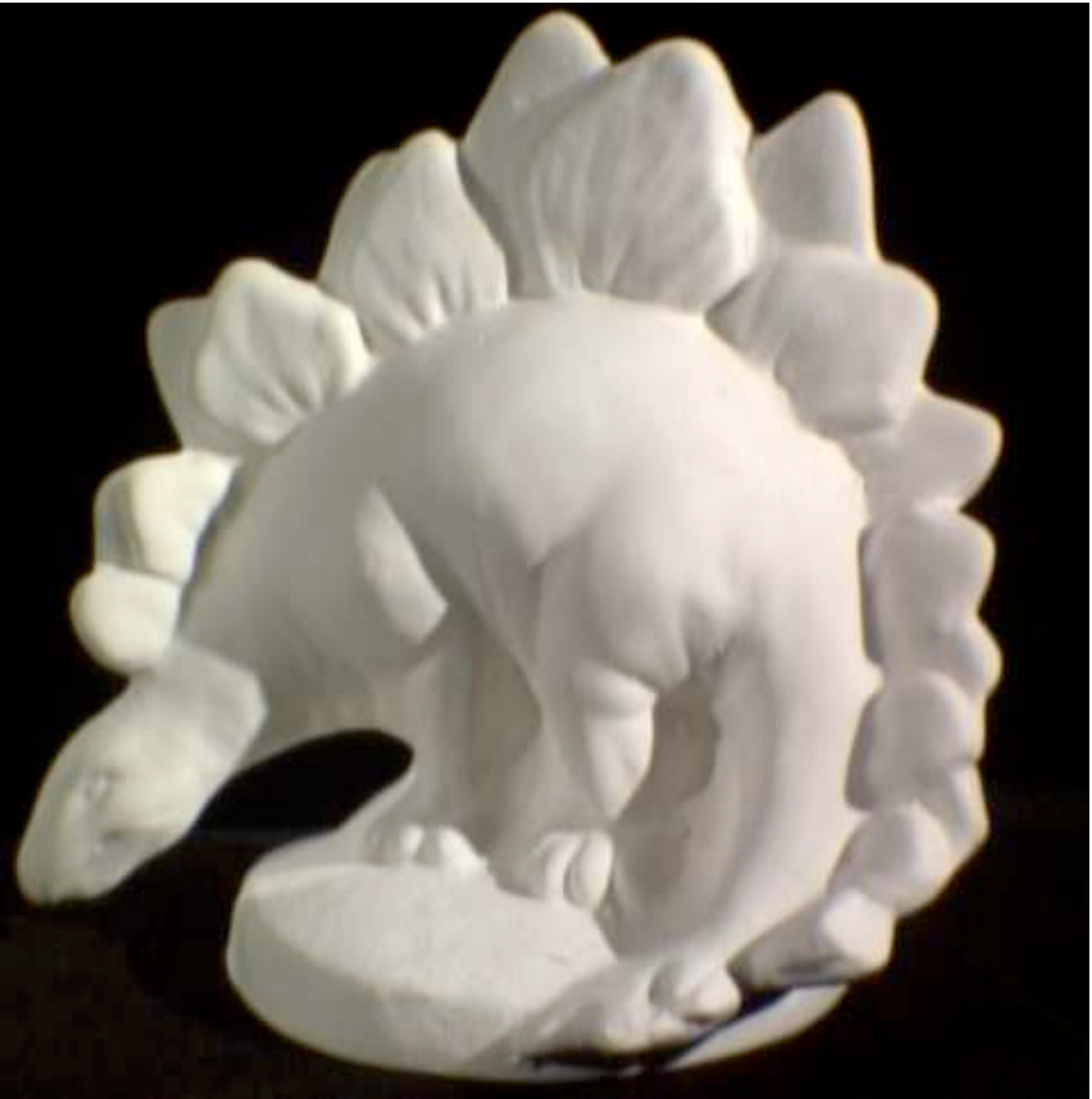} &
\includegraphics[width=0.16\textwidth]{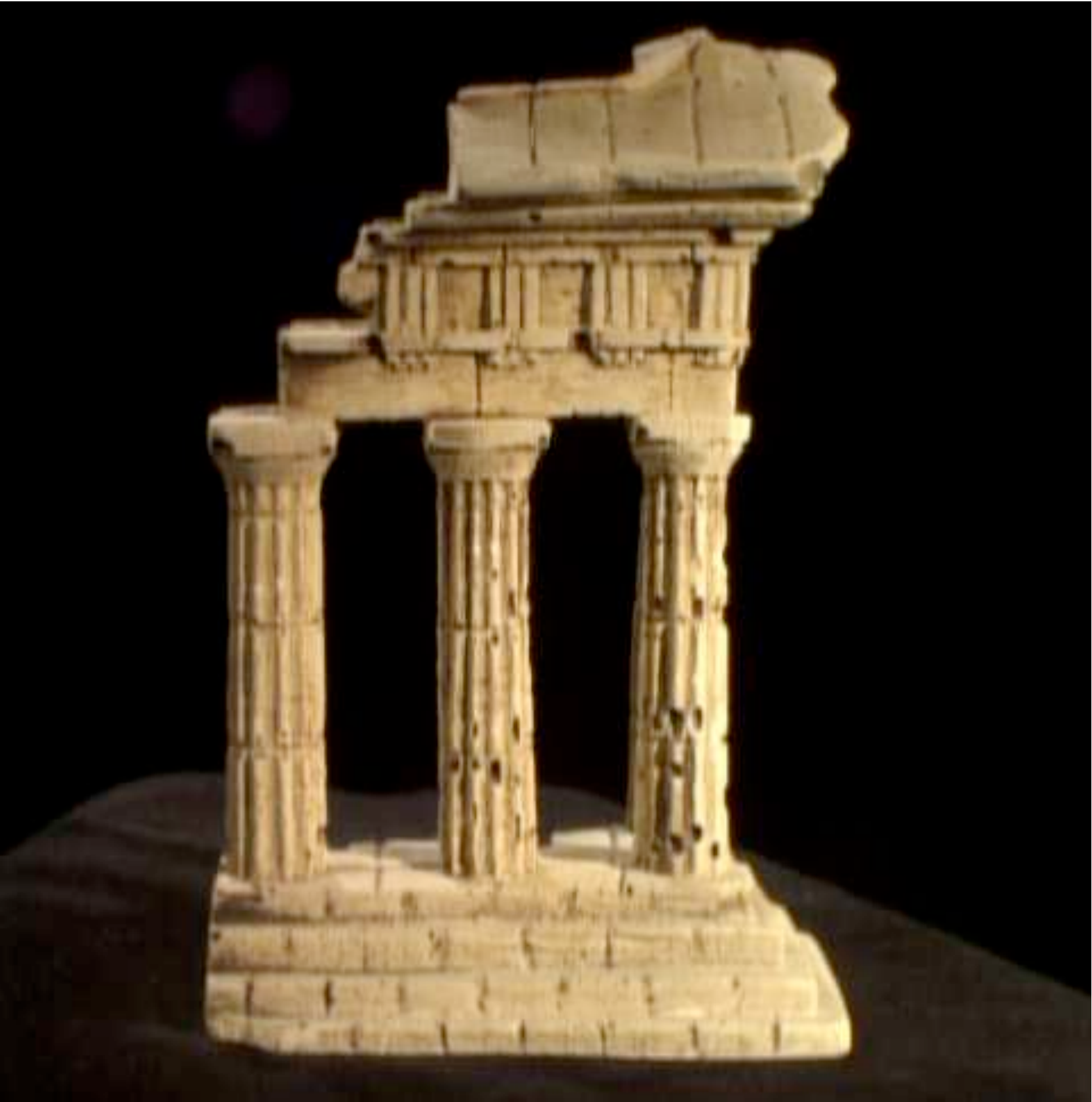}&
\includegraphics[width=0.16\textwidth]{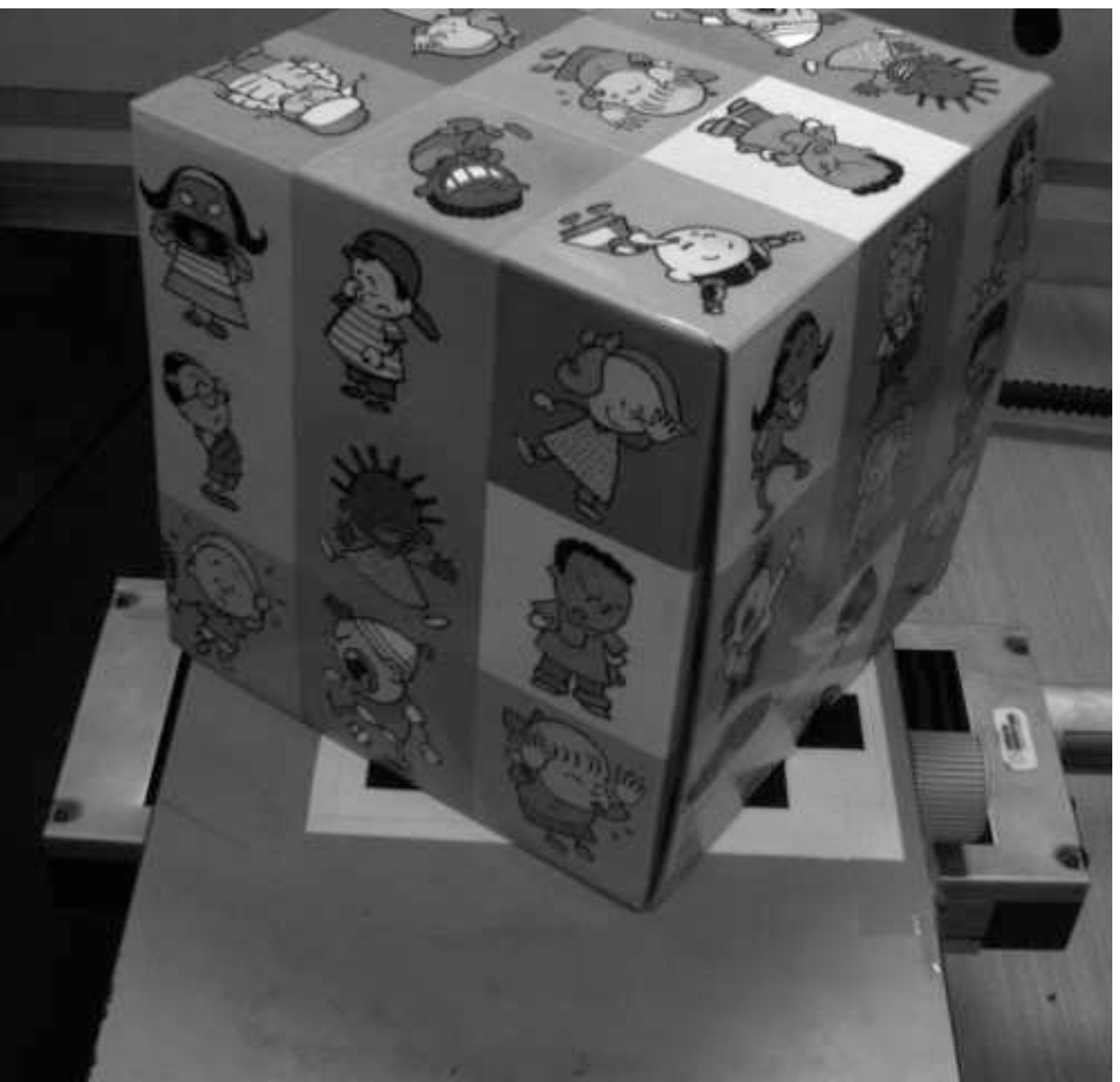}&
\includegraphics[width=0.16\textwidth]{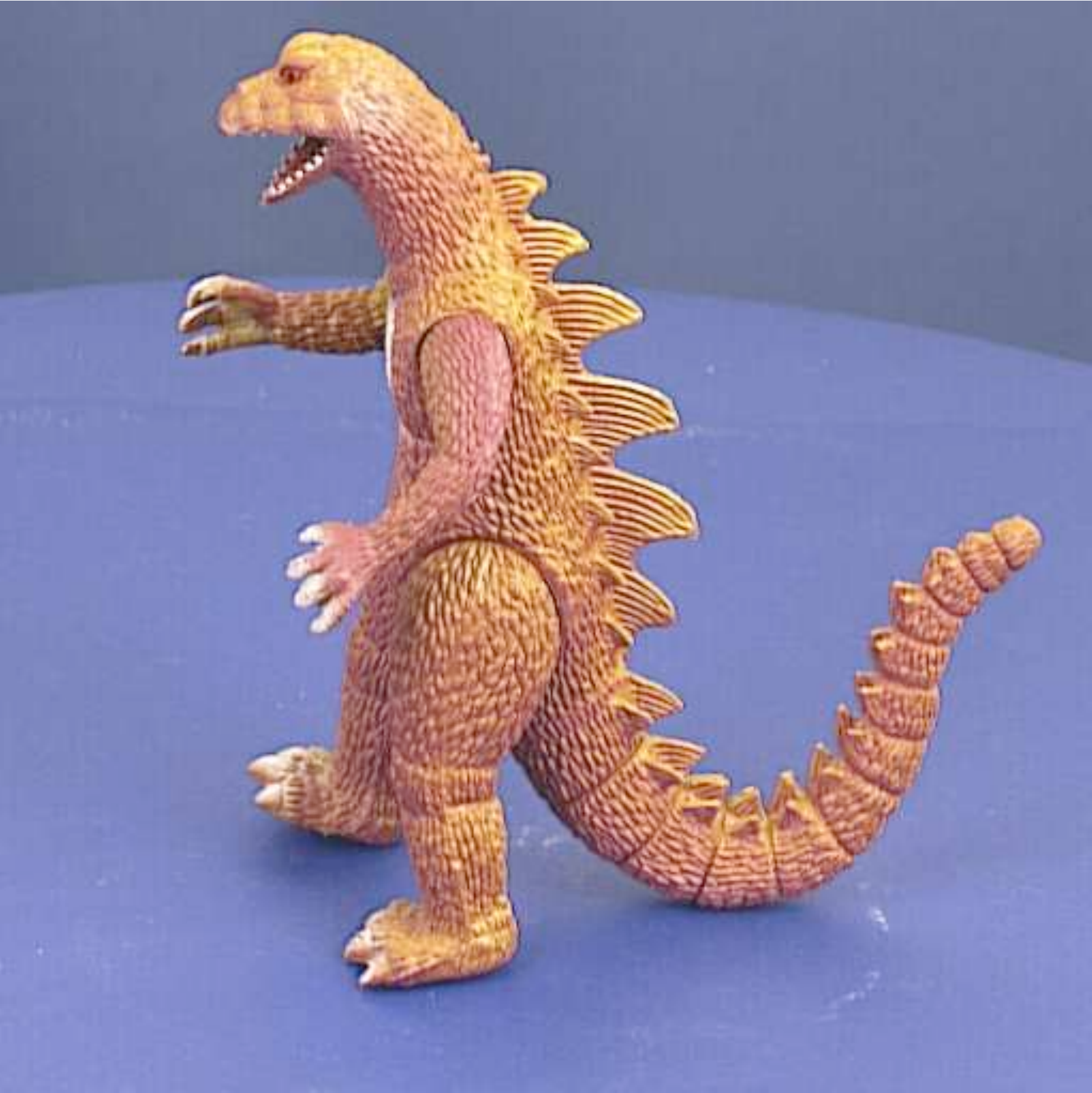}\\

\includegraphics[width=0.16\textwidth]{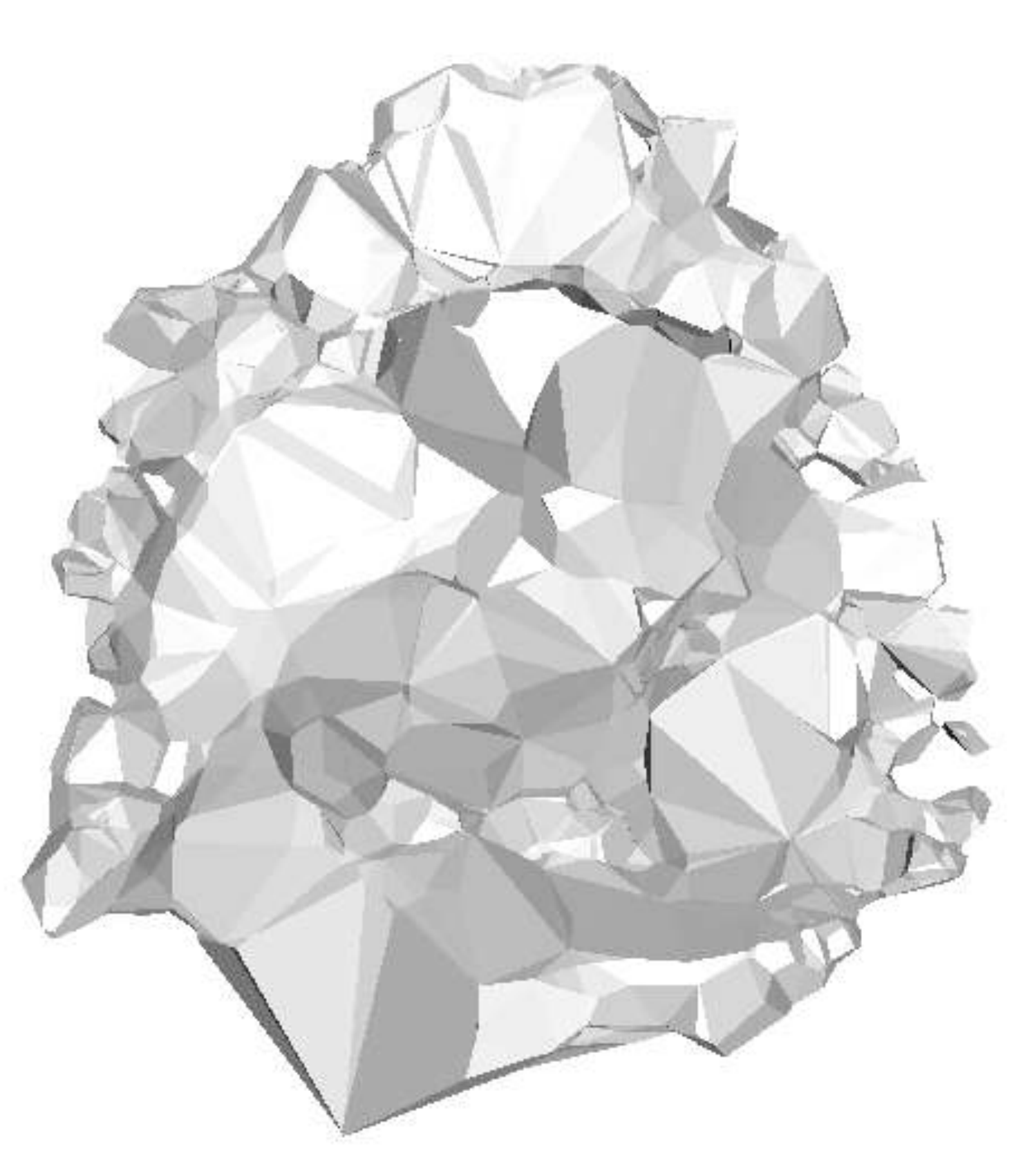} &
\includegraphics[width=0.16\textwidth]{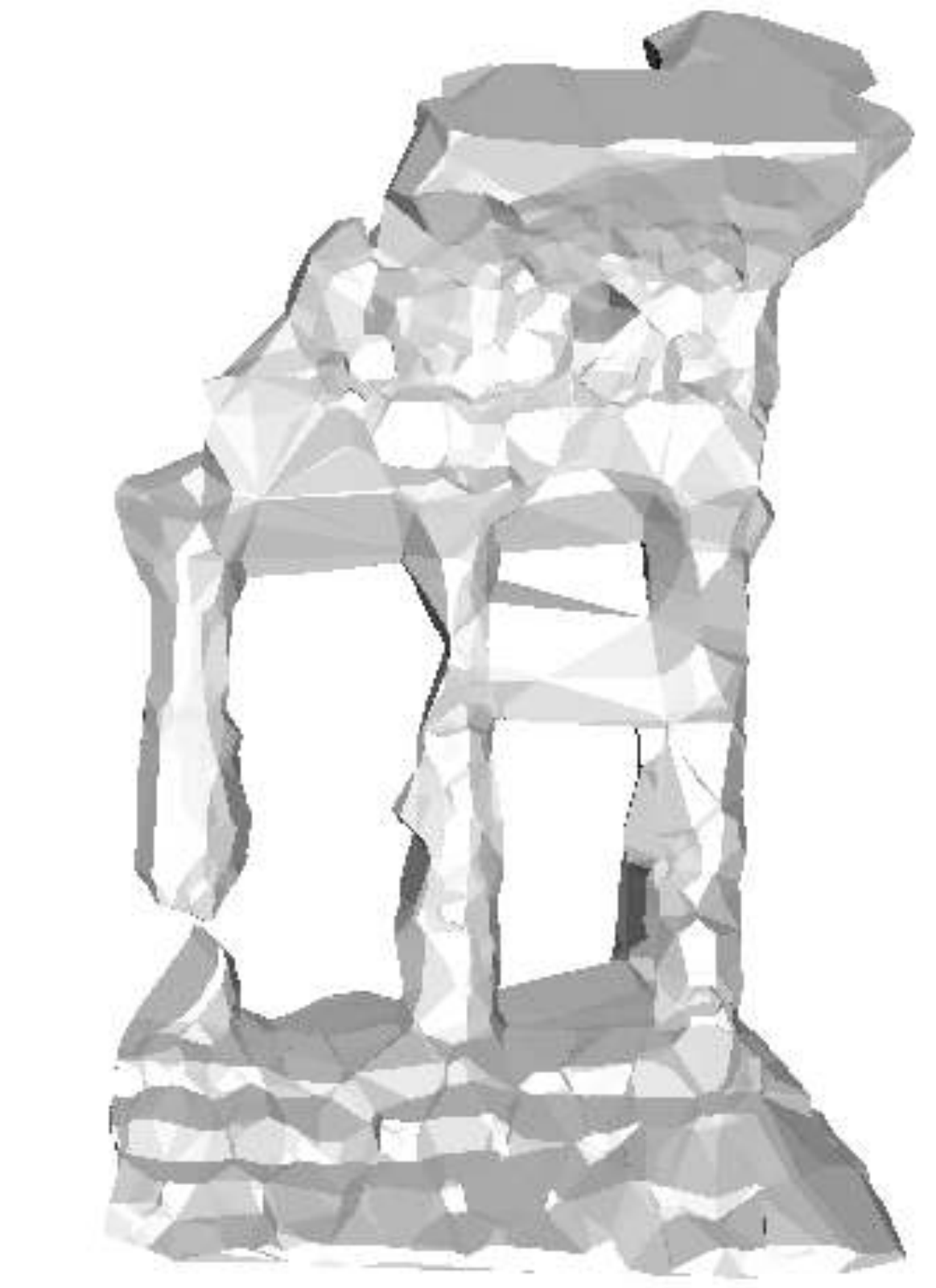}&
\includegraphics[width=0.16\textwidth]{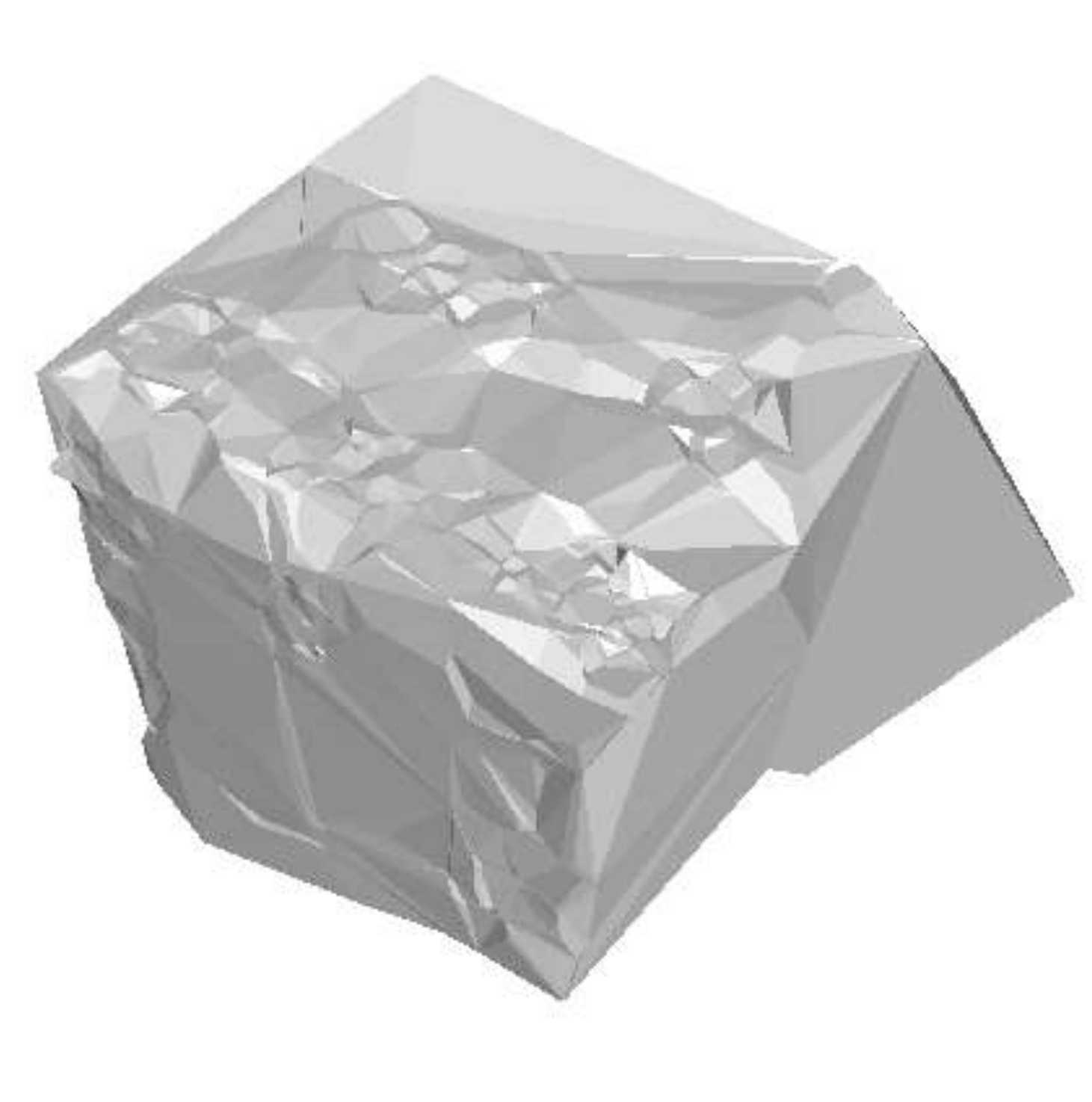}&
\includegraphics[width=0.16\textwidth]{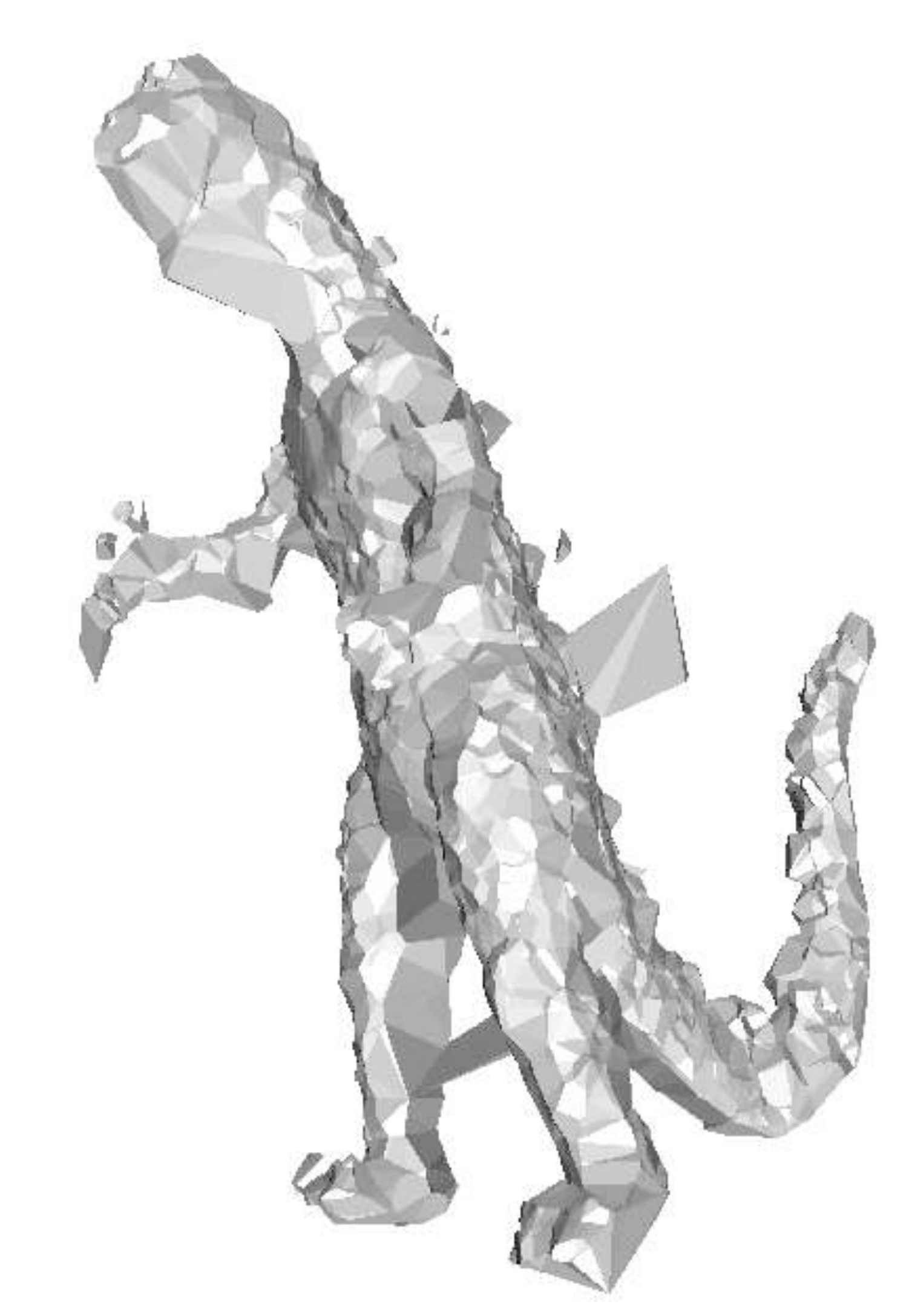}\\

\includegraphics[width=0.16\textwidth]{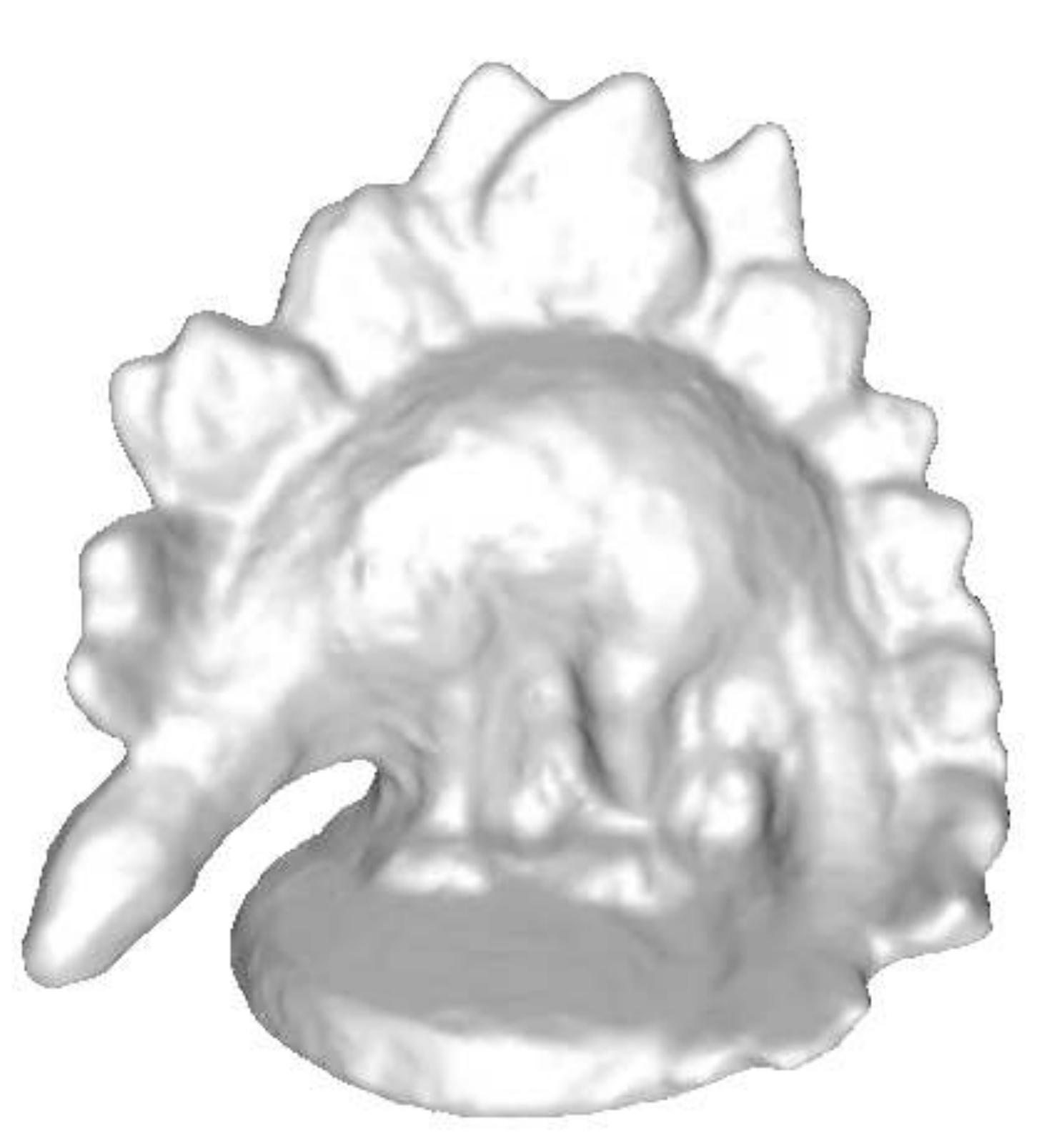} &
\includegraphics[width=0.16\textwidth]{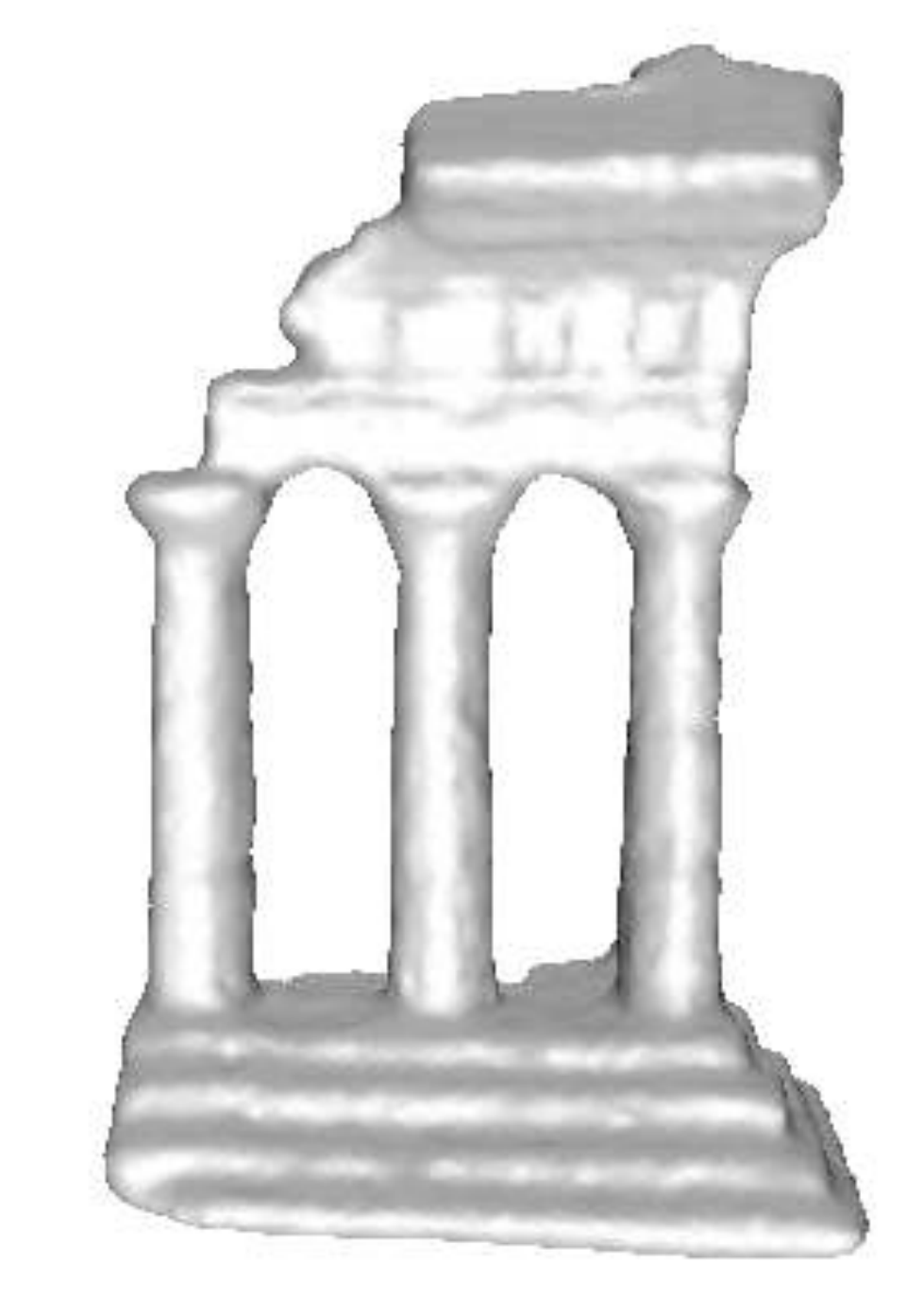}&
\includegraphics[width=0.16\textwidth]{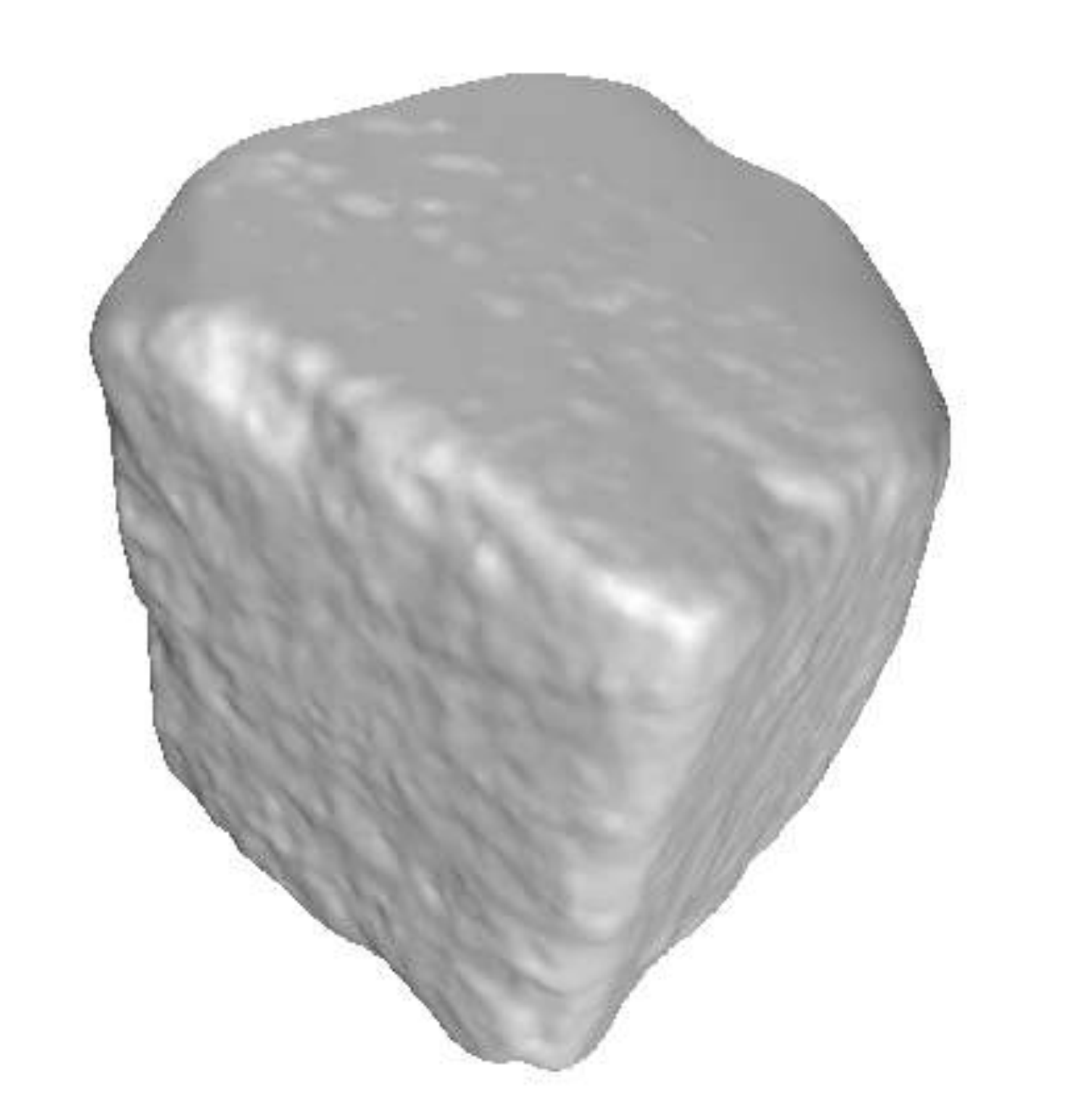}&
\includegraphics[width=0.16\textwidth]{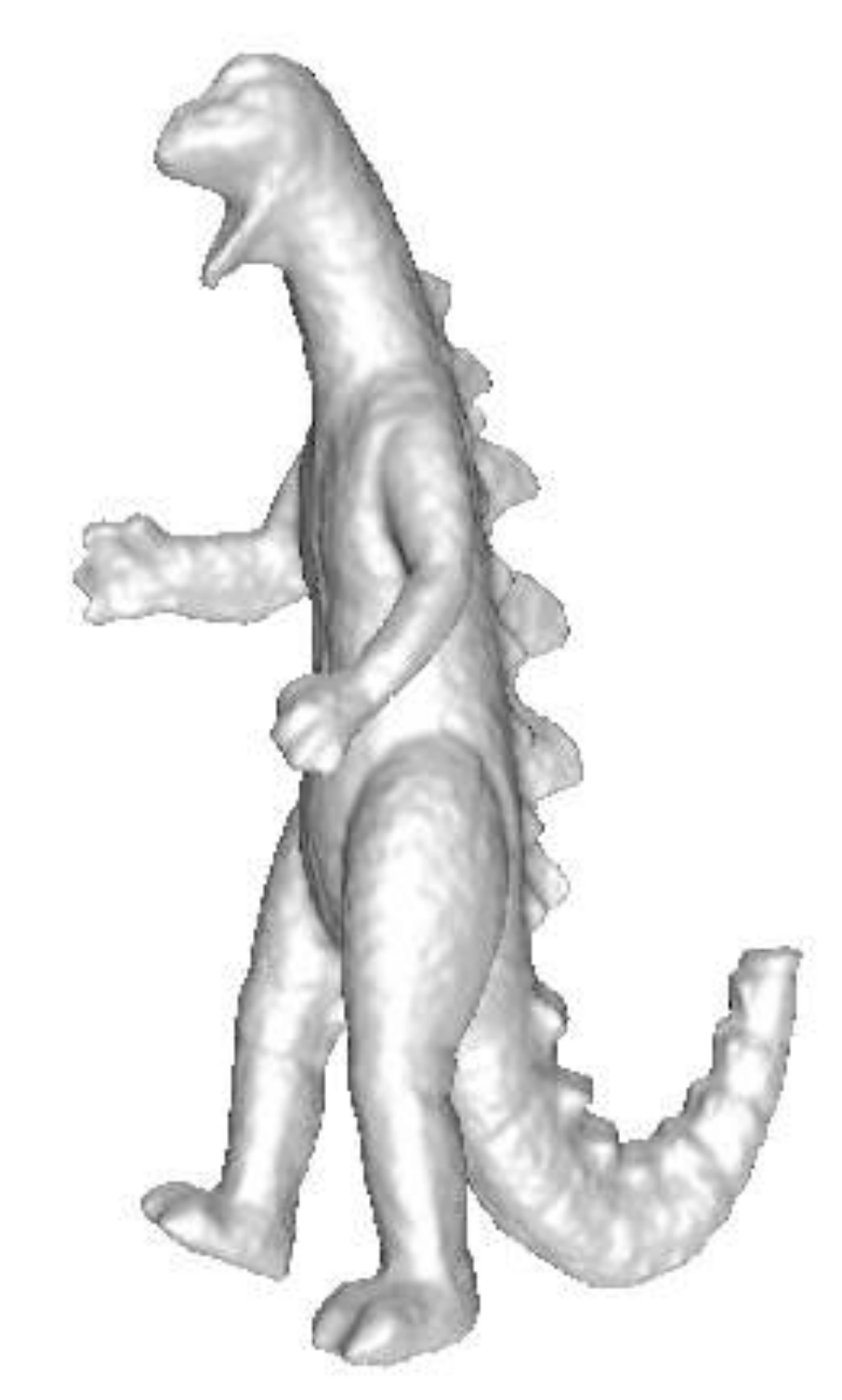}\\
\hline
\end{tabular}
\caption[rec_mesh]{Additional dense reconstruction results. First Row: Sample Input Image; Second Row: A rough mesh obtained using PowerCrust \cite{Amenta:2001} from the sparse 3-D points, reconstructed using \cite{Zaharescu:2009tg};  Third Row: the final dense reconstruction after surface evolution. } 
\label{fig:rec_from_points} 
\end{figure*}

We have tested the mesh evolution algorithm with the datasets provided
by the Multi-View Stereo evaluation site \cite{Seitz:2006uq}\footnote{\url{http://vision.middlebury.edu/mview/}}. The ground-truth
is obtained from laser-scans. 
Comparative and detailed results are extracted from the Middlebury
website and are presented in Table~\ref{table:results}. The table includes
results from Furukawa and Ponce~\cite{Furukawa:2009pami}, Pons
{\it et al.}~\cite{Pons:2007}, Vu {\it et al.}~\cite{Vu:2009a} and
Hernandez and Schmitt~\cite{Hernandez:2004}; all these methods yield
state-of-the-art results. The differences between all these methods are very small,
ranging between $0.01mm$ to $0.1mm$. Some of our reconstruction
results are shown in Figure~\ref{fig:results-middleburry} and Figure~
\ref{fig:rec_from_points}. 
While Vu et al. \cite{Vu:2009a} used the same
energy functional as part of their 3-D reconstruction pipeline, their
improved results are mostly due to the fact that the mesh
regularization term takes into account photo-consistency.

\begin{figure}[!htbp]
\centering
\includegraphics[width=0.22\columnwidth]{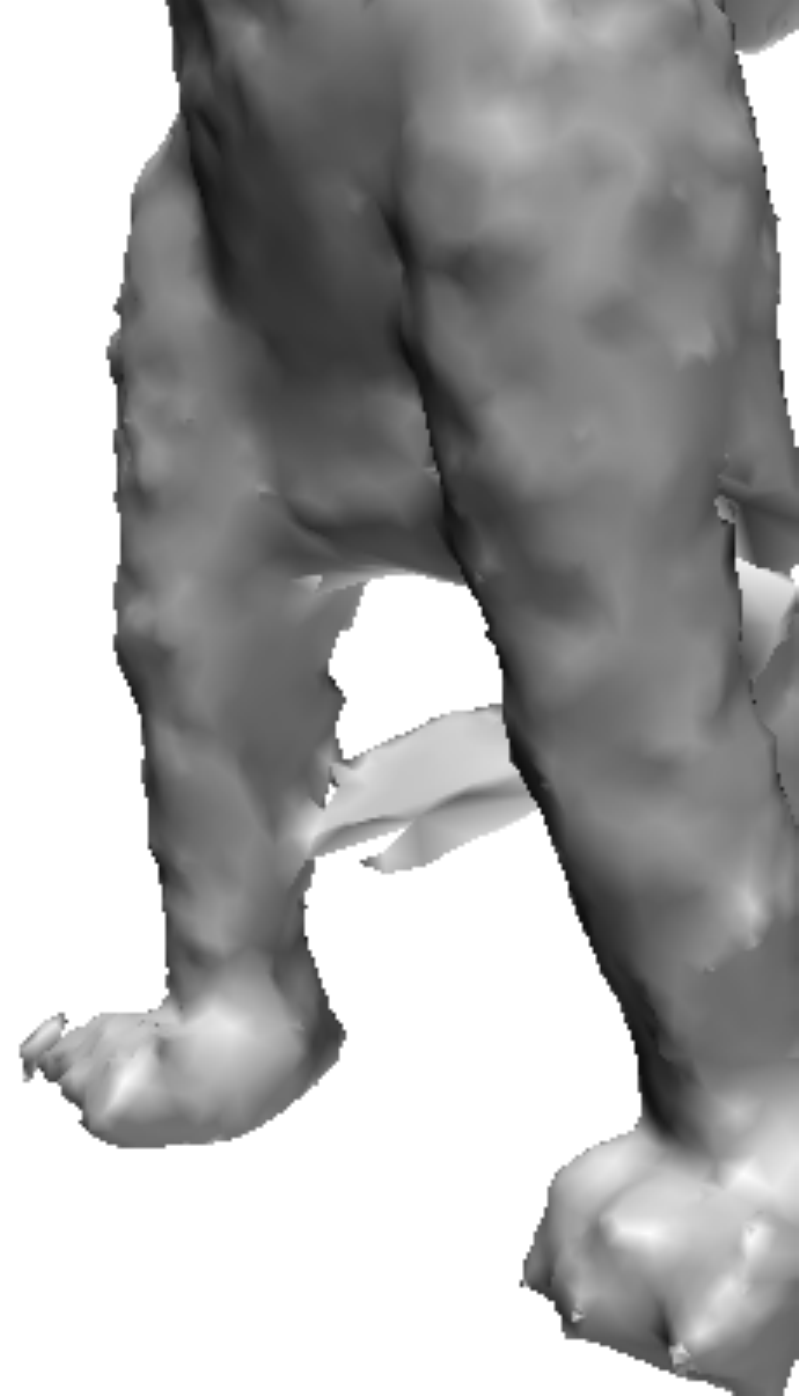} 
\includegraphics[width=0.22\columnwidth]{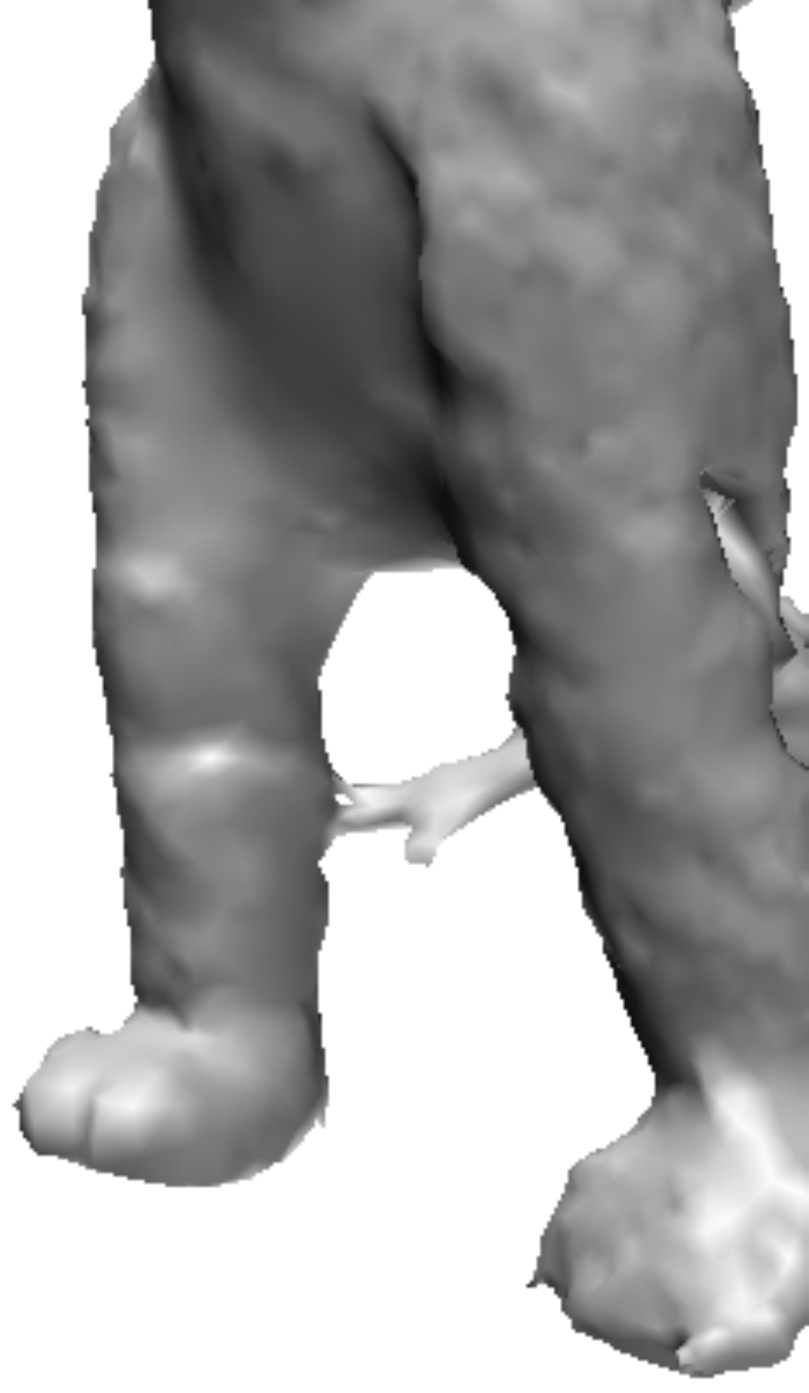} 
\includegraphics[width=0.22\columnwidth]{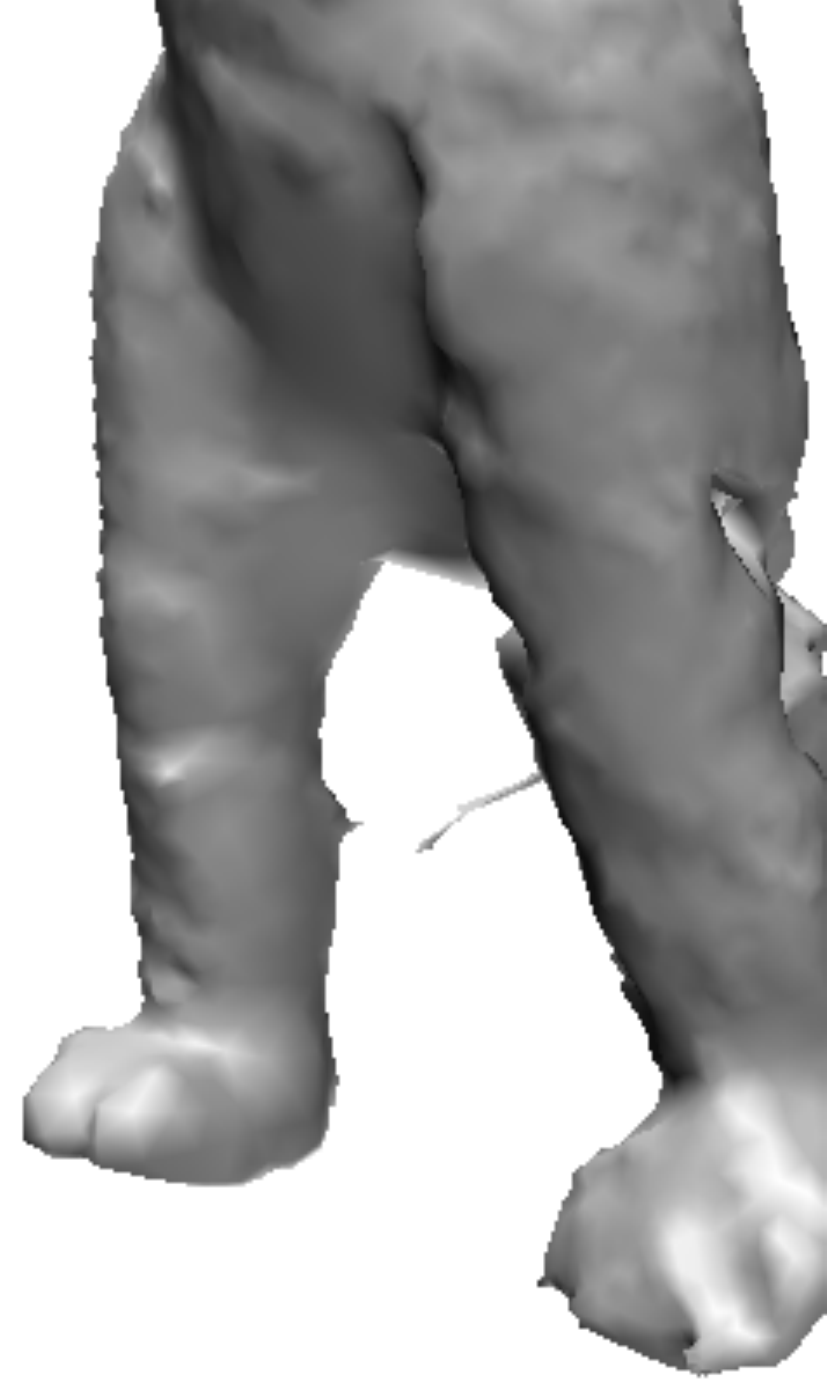} 
\includegraphics[width=0.22\columnwidth]{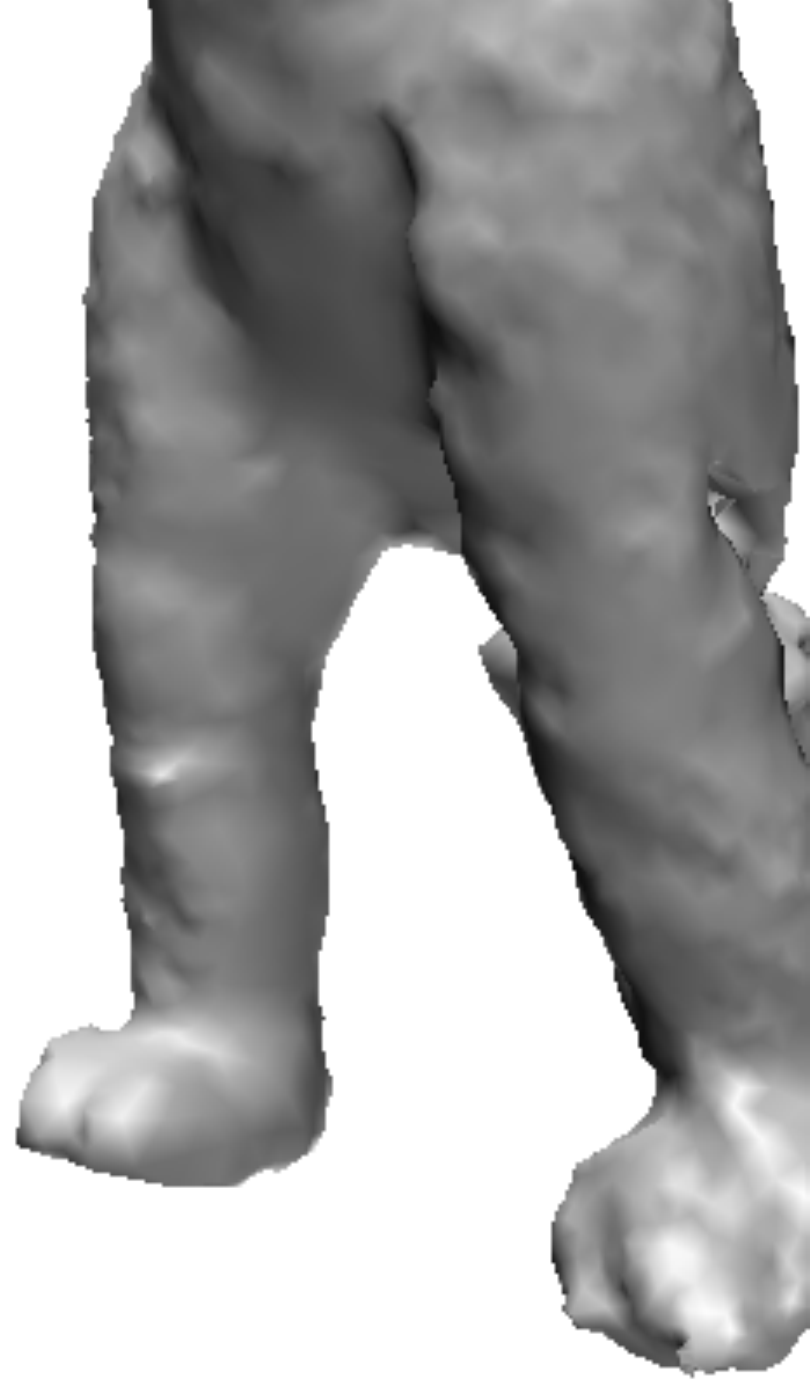} 
\caption[rec_mesh]{Example of topological changes during in 3-D reconstruction for the dinosaur sequence, introduced in Figure \ref{fig:rec_from_points}. The start-up surface, obtained from triangulated 3-D points via PowerCrust\cite{Amenta:2001} contains several topological errors (i.e. the extra branch connecting the dinosaur's limbs). They are corrected during the surface evolution, ash shown in the right most image. } 
\label{fig:rec_from_points_closeup} 
\end{figure}

Figure \ref{fig:rec_from_points} shows the results obtained with our
method when starting with very rough meshes that correspond to coarse
triangulations obtained from a sparse set of 3-D points. 
An example of how \MeshAlg\ handles topological changes is shown in
Figure~\ref{fig:rec_from_points_closeup}. This figure shows a 
typical evolution scenario where there are more ``topological problems'' at
the beginning; As the algorithm converges, self-intersections barely occur. 

Finally, Figure~\ref{fig:results_ben} shows results obtained with the Man-dance sequence
publicly  available  from  the Multiple-video 
database of the PERCEPTION group at INRIA\footnote{\url{http://4drepository.inrialpes.fr/}}.

\begin{small}
\begin{figure*}[!htbp]
\centering
\begin{tabular}{ccccc}

\hline
& Sample Image & Initial Mesh & Final Result & Result Close-up \\
\toprule

\raisebox{5em}{\begin{sideways} Dino \end{sideways}} &
\includegraphics[width=0.16\textwidth, height=0.19\textwidth]{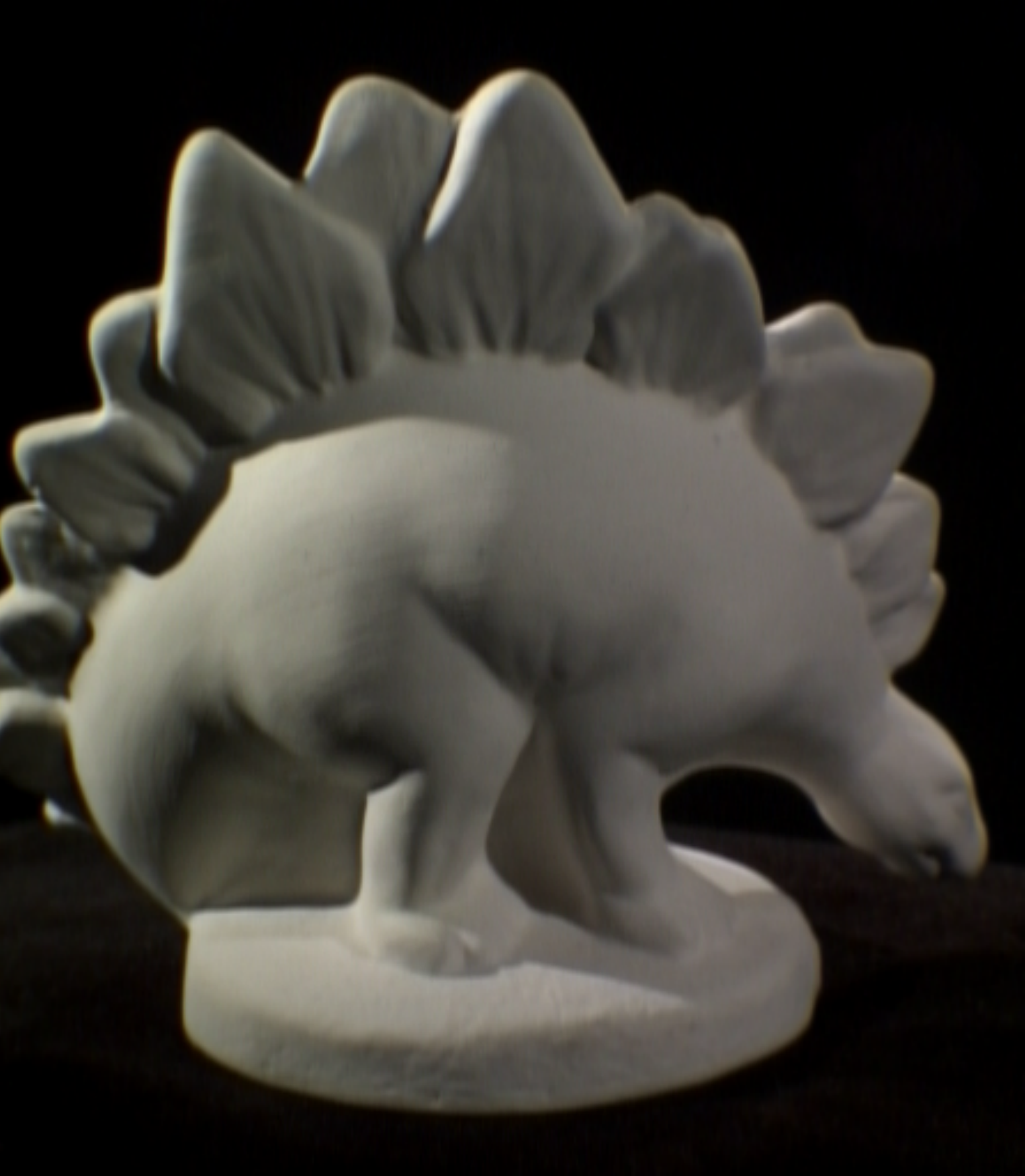} &
\includegraphics[width=0.16\textwidth, height=0.19\textwidth]{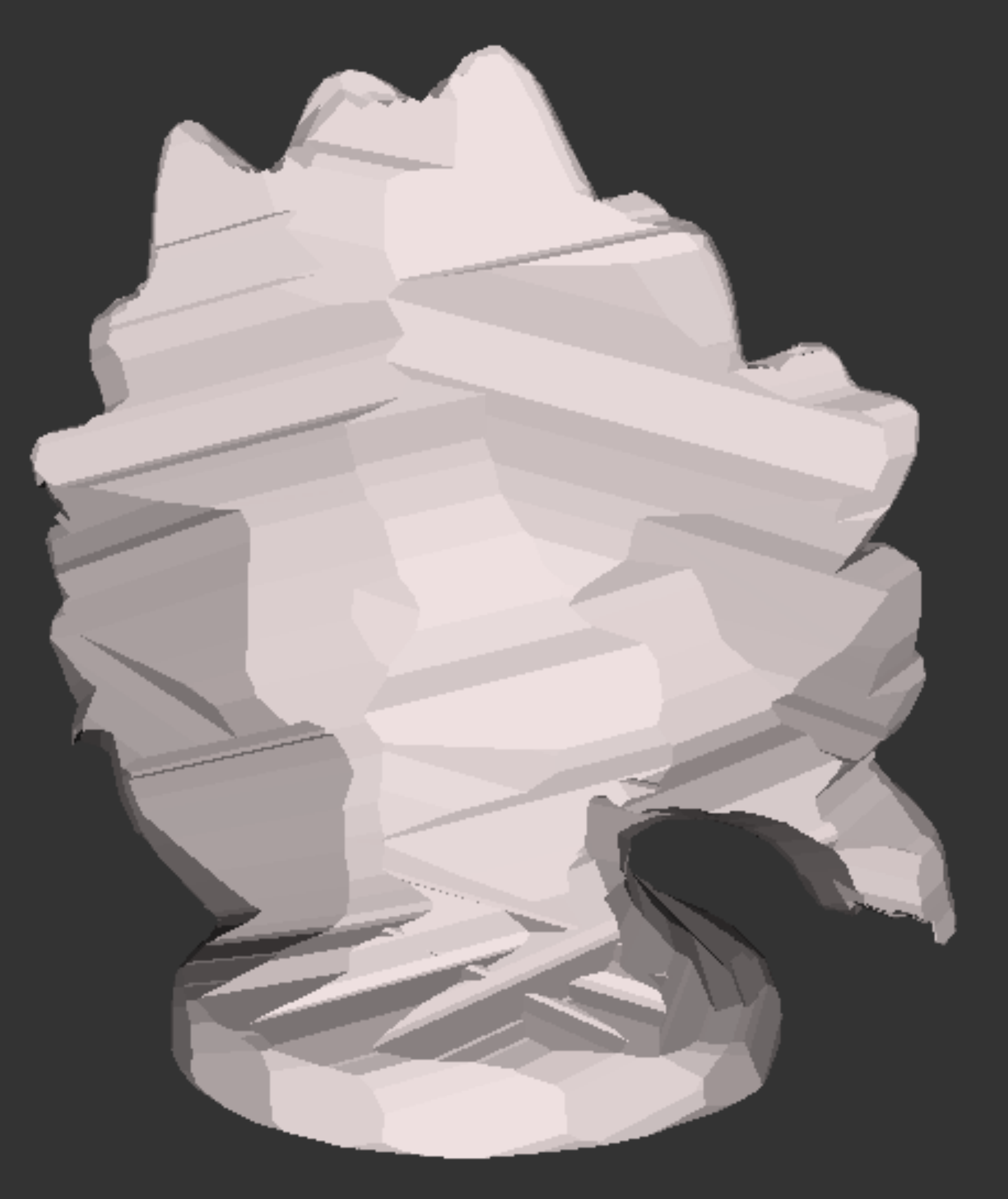} &
\includegraphics[width=0.16\textwidth, height=0.19\textwidth]{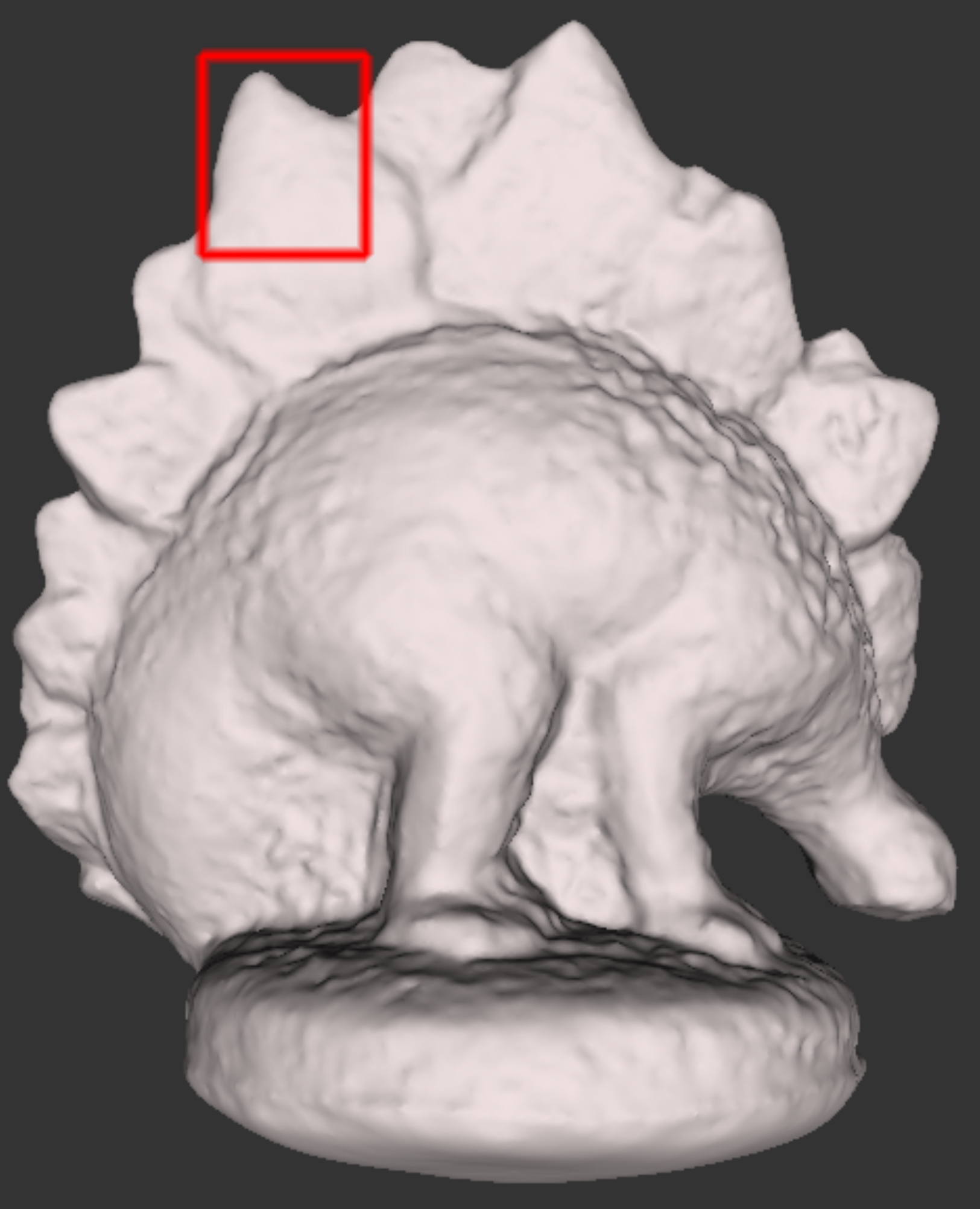} &
\includegraphics[width=0.16\textwidth, height=0.19\textwidth]{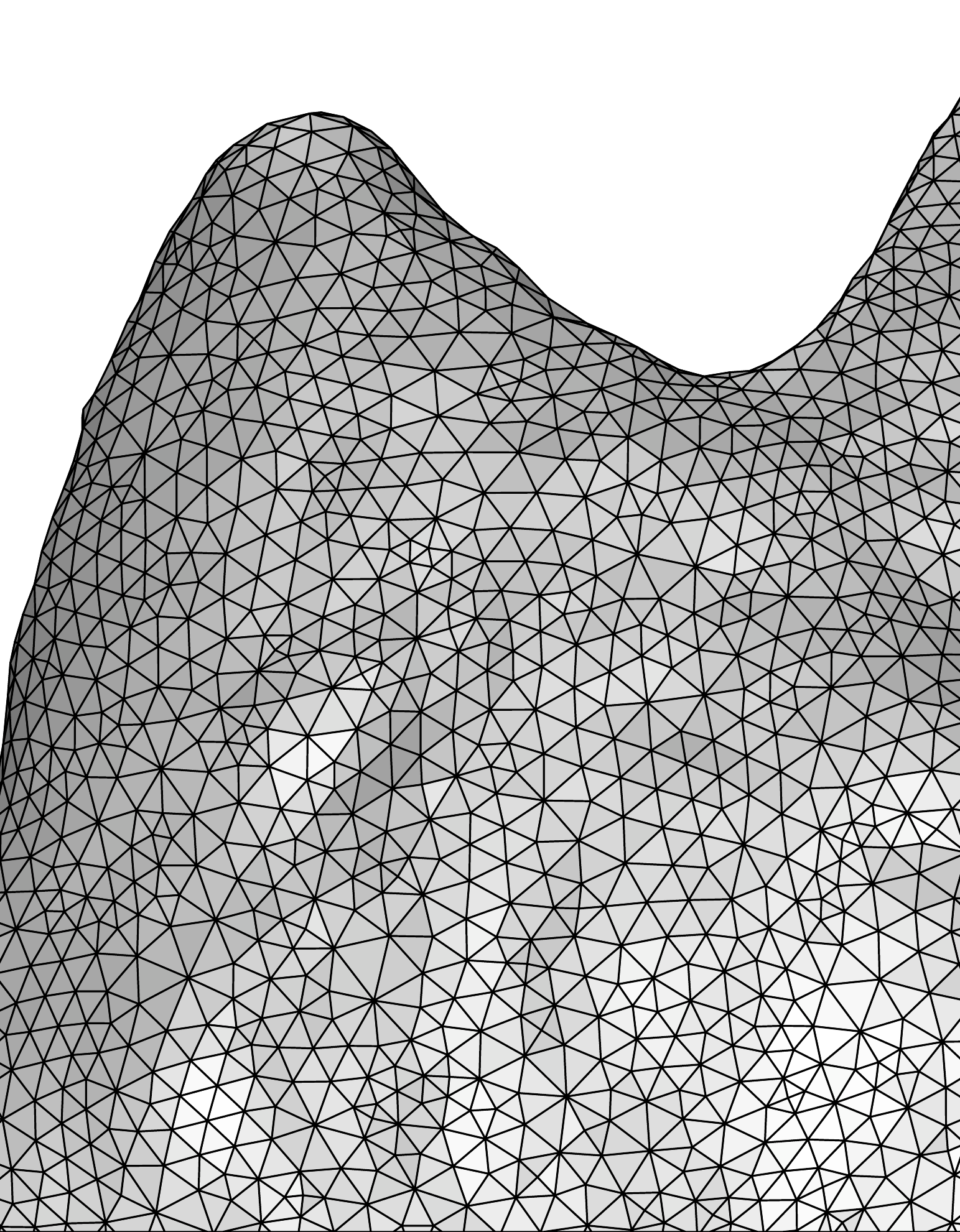} \\

\raisebox{5em}{\begin{sideways} Temple \end{sideways}} &
\includegraphics[width=0.16\textwidth, height=0.19\textwidth]{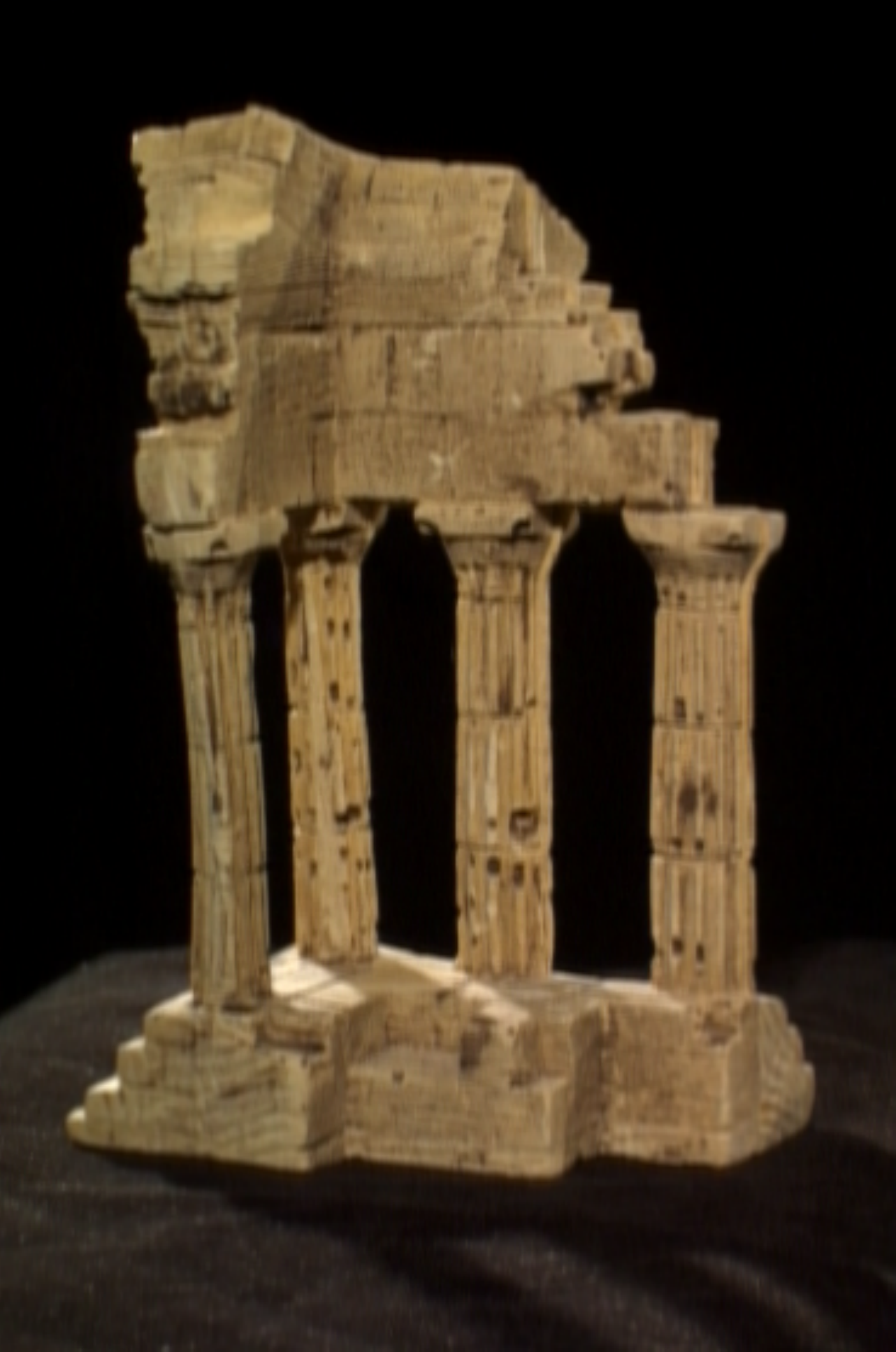} &
\includegraphics[width=0.16\textwidth, height=0.19\textwidth]{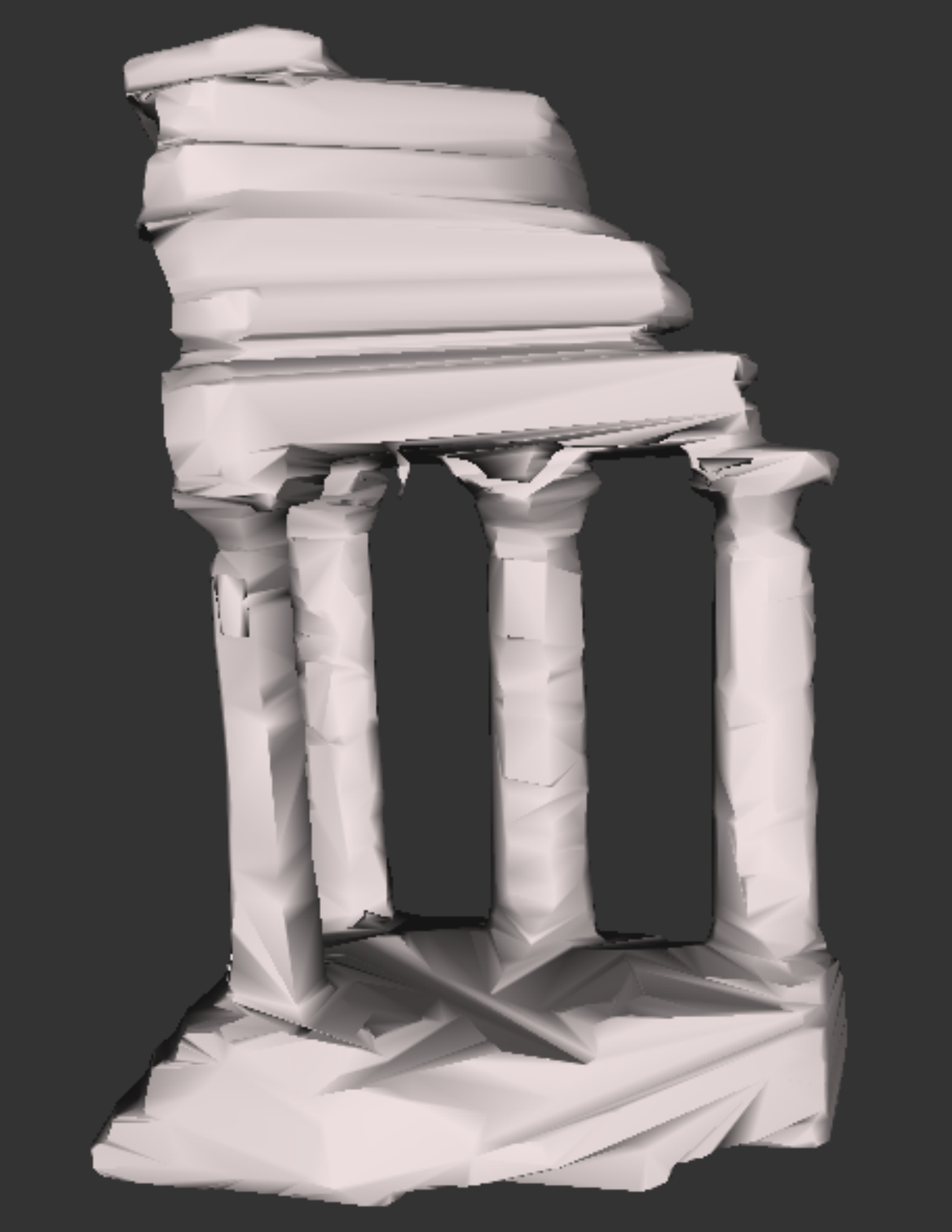} &
\includegraphics[width=0.16\textwidth, height=0.19\textwidth]{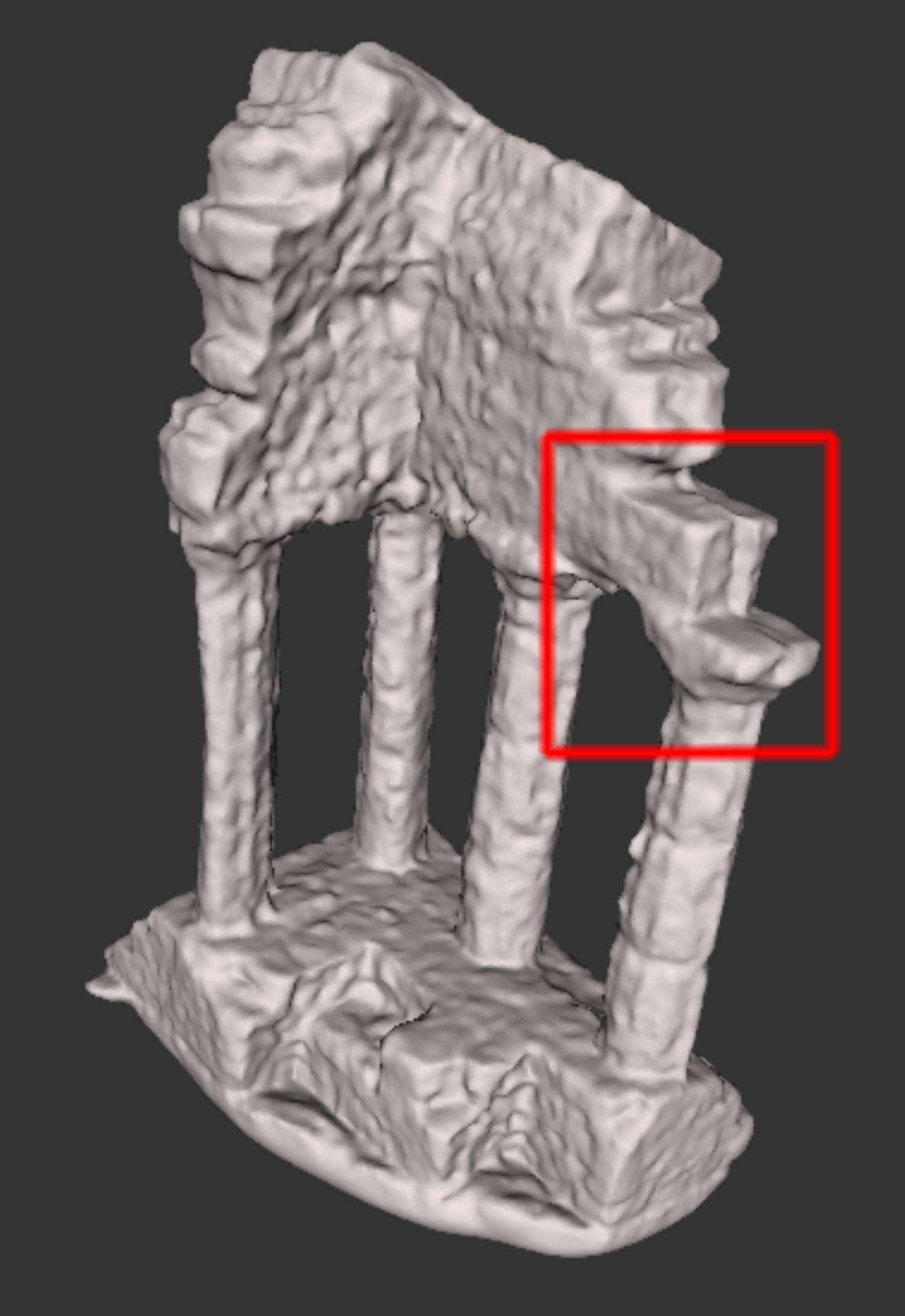} &
\includegraphics[width=0.16\textwidth, height=0.19\textwidth]{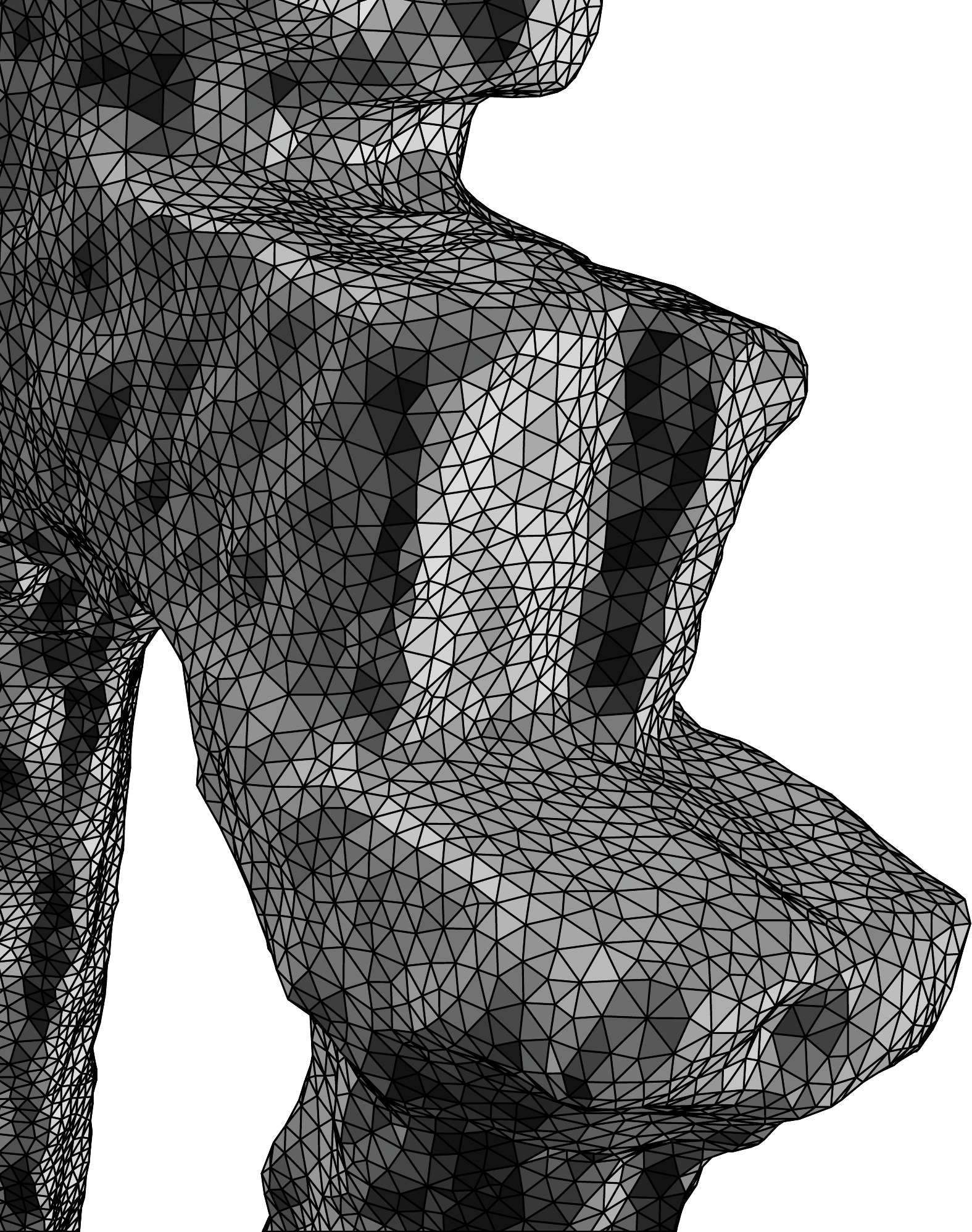} \\
\hline
\end{tabular}
\caption{Reconstruction Results for the Middleburry multiview dataset (dino case and temple case)} 
\label{fig:results-middleburry}
\end{figure*}
\end{small}

\newcommand{\includegraphicsBen}[1]{\includegraphics[height= 4.0cm]{#1}}

\begin{figure*}[htbp]
\begin{center}$
\begin{array}{cccccc}
\hline
&
\textrm{frame 500} &
\textrm{frame 511} &
\textrm{frame 522} &
\textrm{frame 533} &
\textrm{frame 544} \\
\toprule
\raisebox{1em}{\begin{sideways} Sample Input Images \end{sideways}} &
\includegraphicsBen{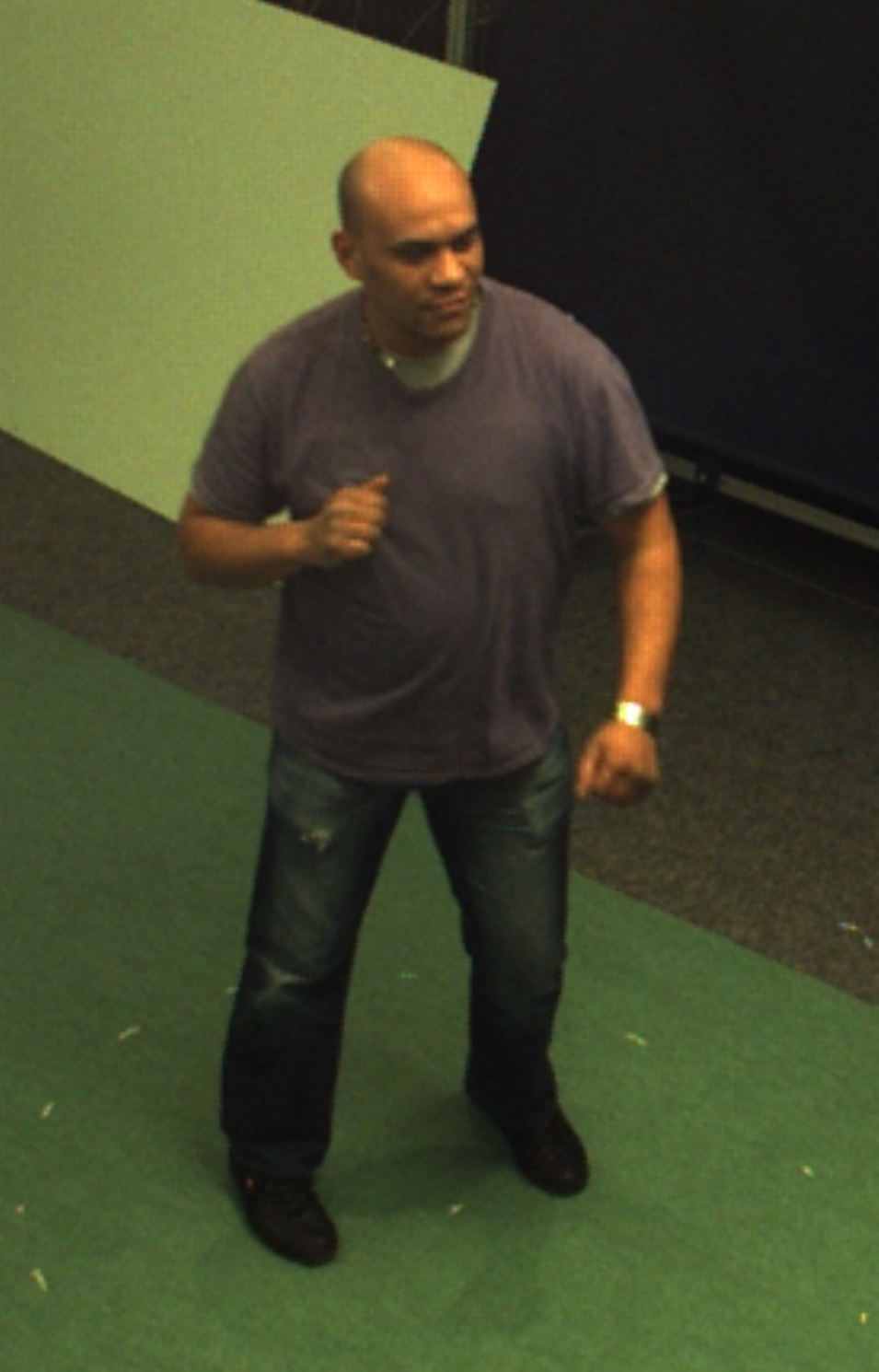} &
\includegraphicsBen{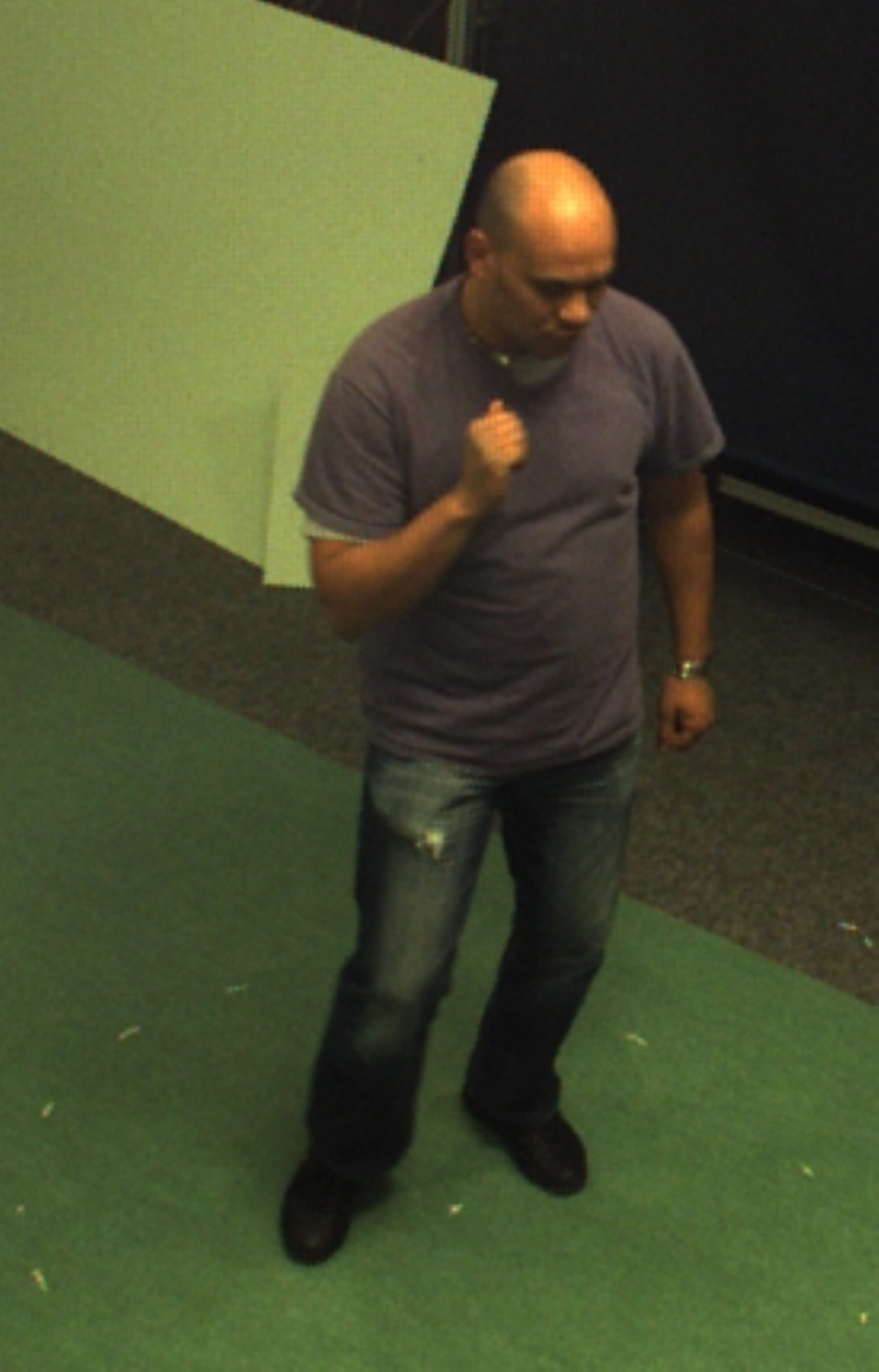} &
\includegraphicsBen{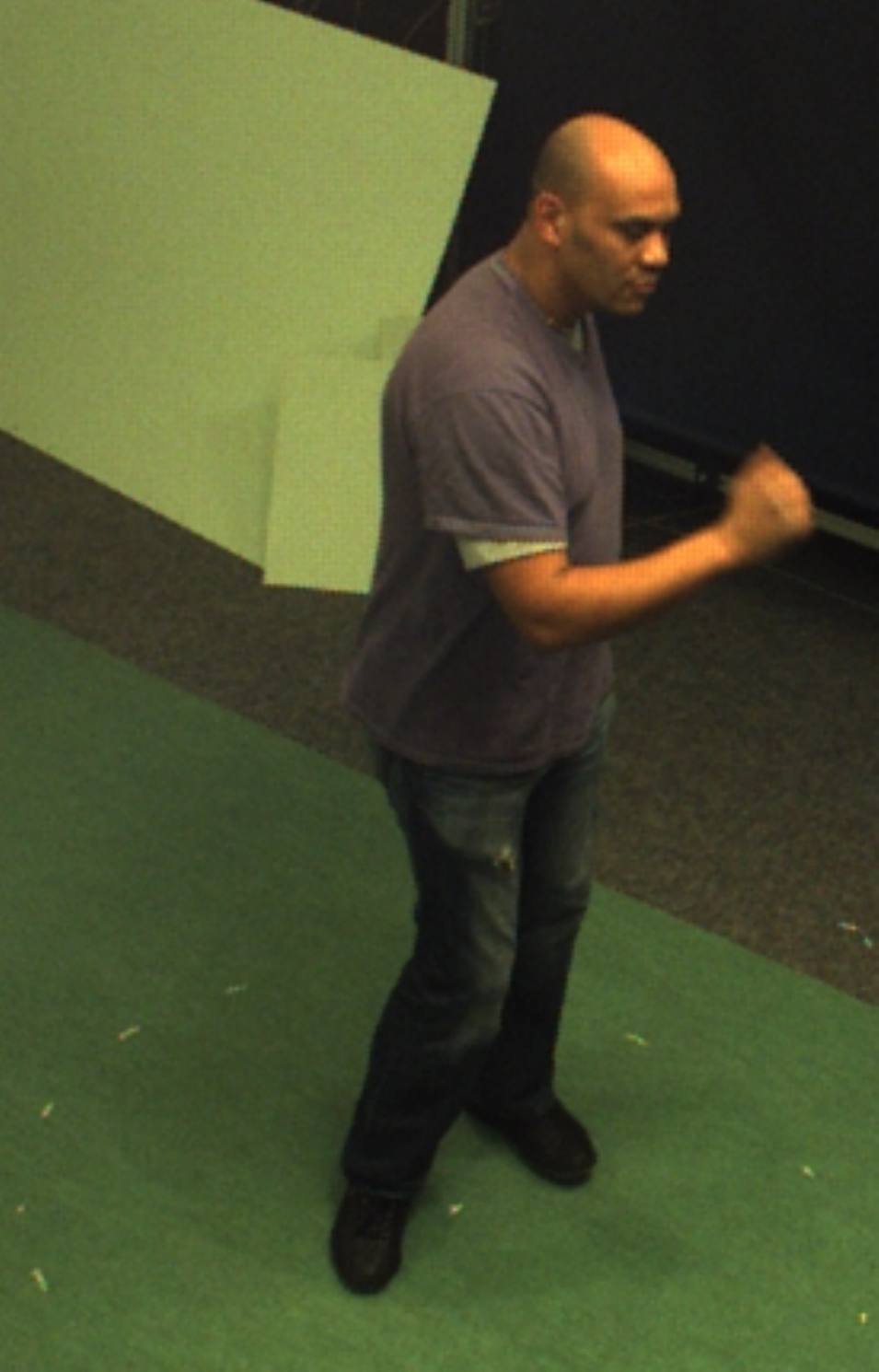} &
\includegraphicsBen{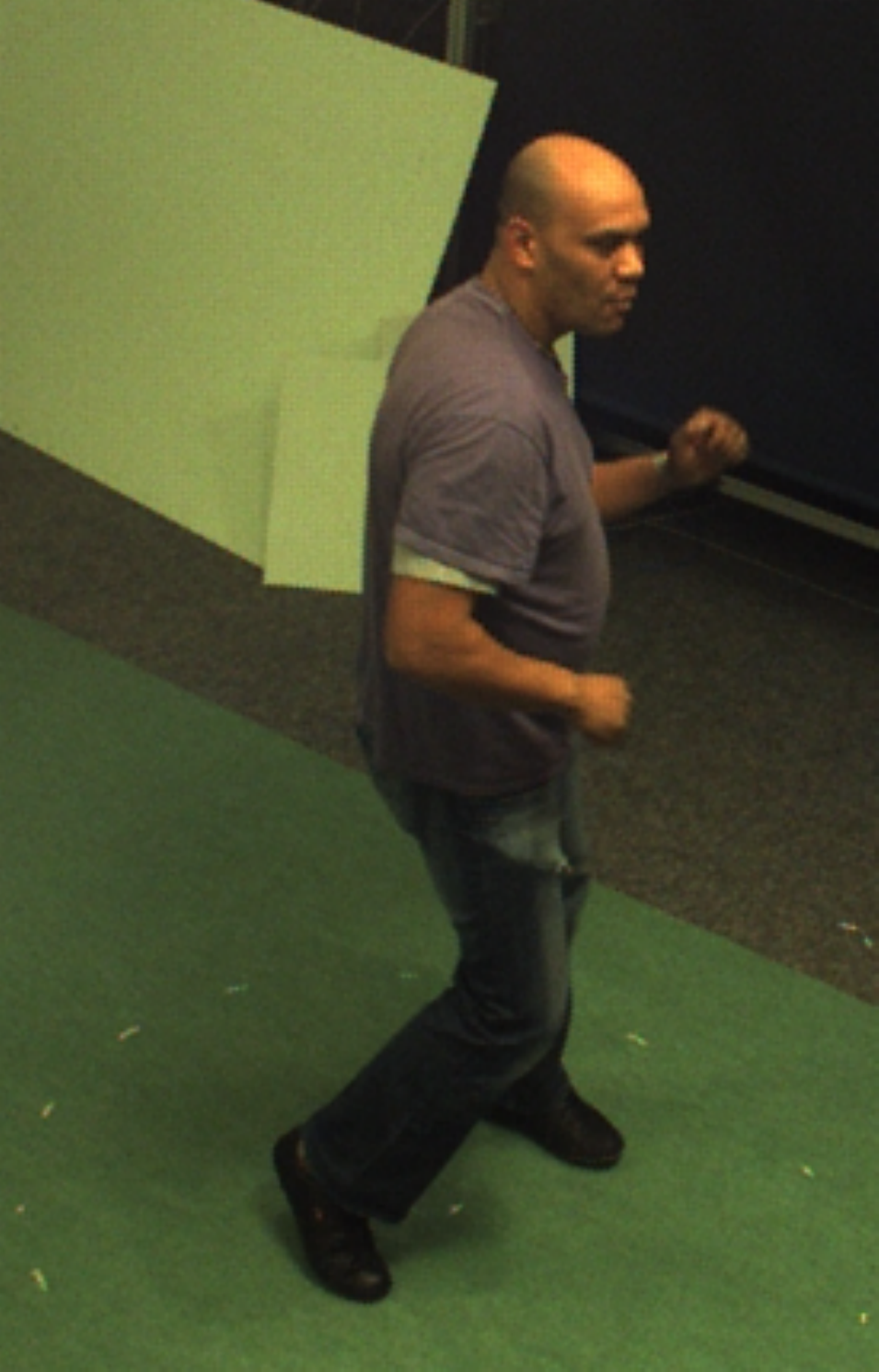} &
\includegraphicsBen{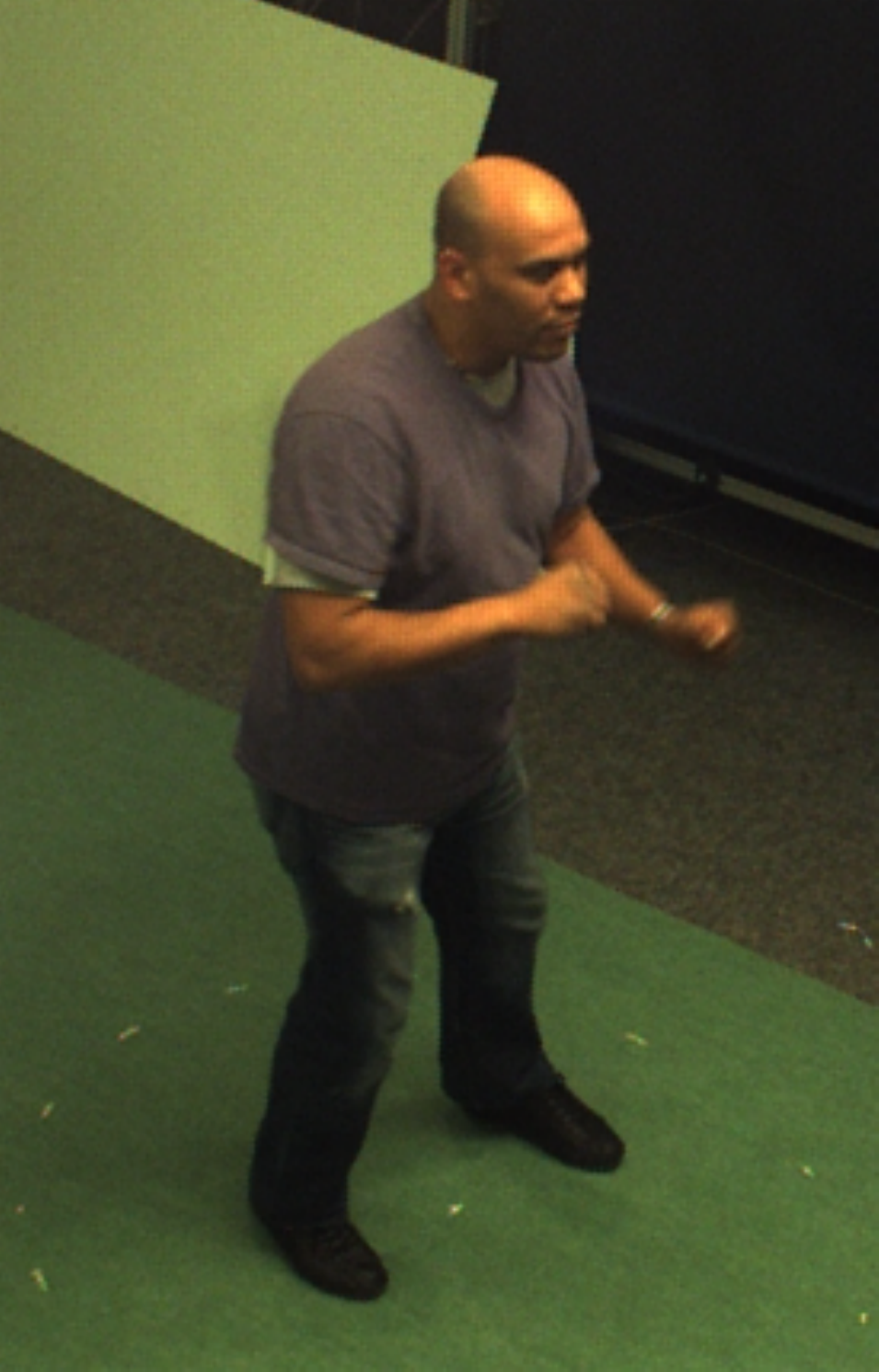} \\

\raisebox{2em}{\begin{sideways} \bigskip Initial Meshes  \end{sideways}} &
\includegraphicsBen{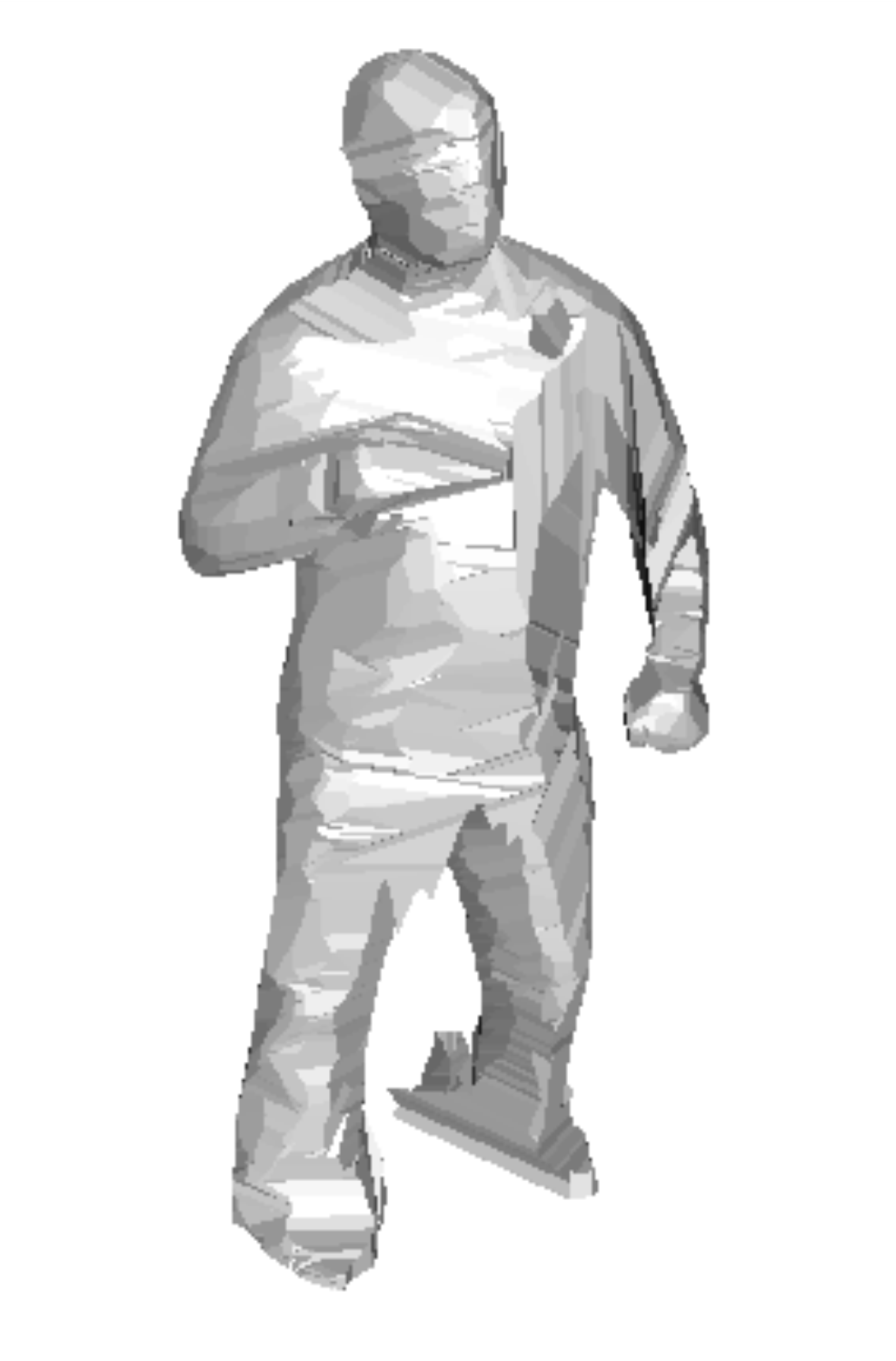} &
\includegraphicsBen{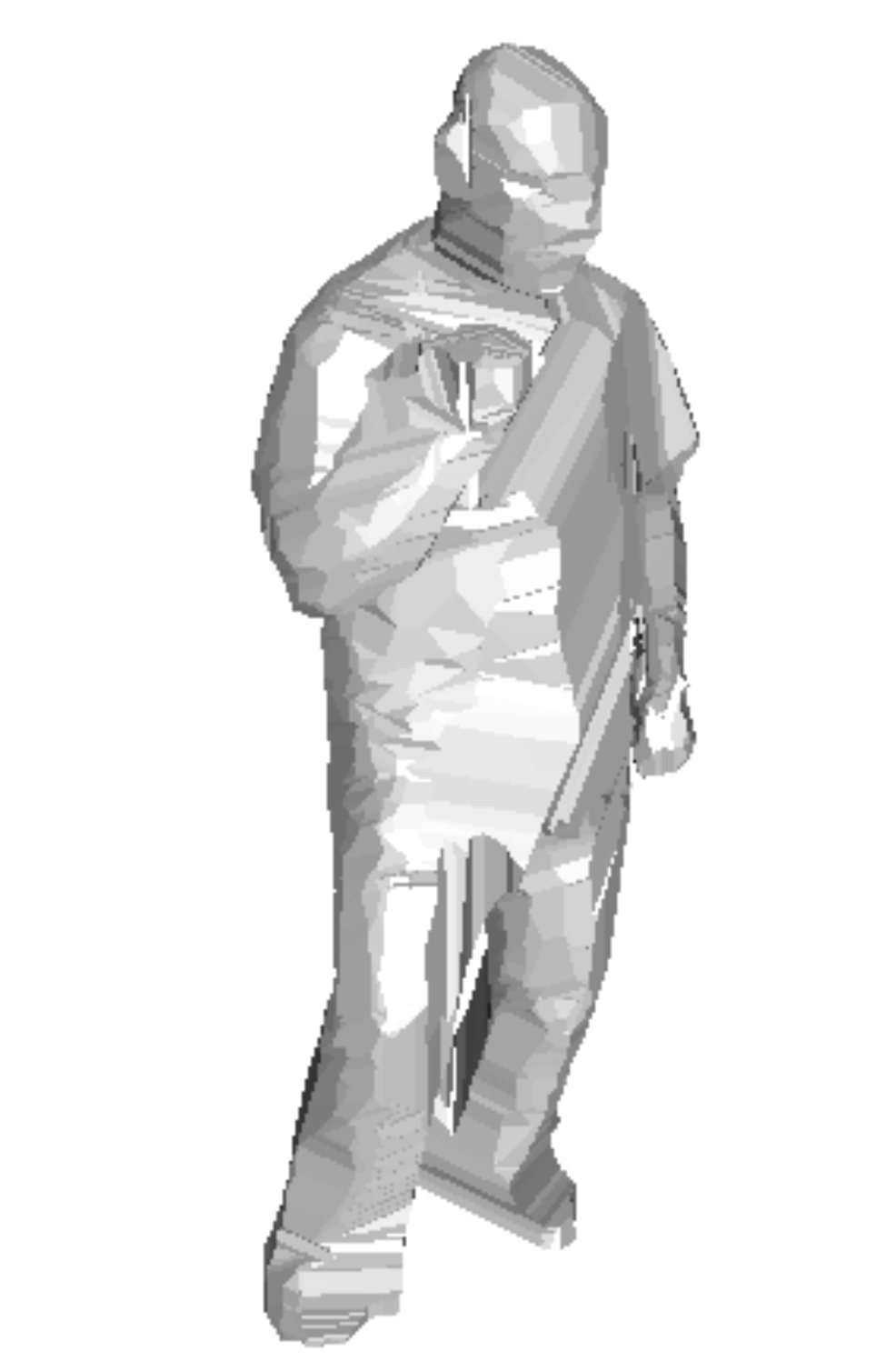} &
\includegraphicsBen{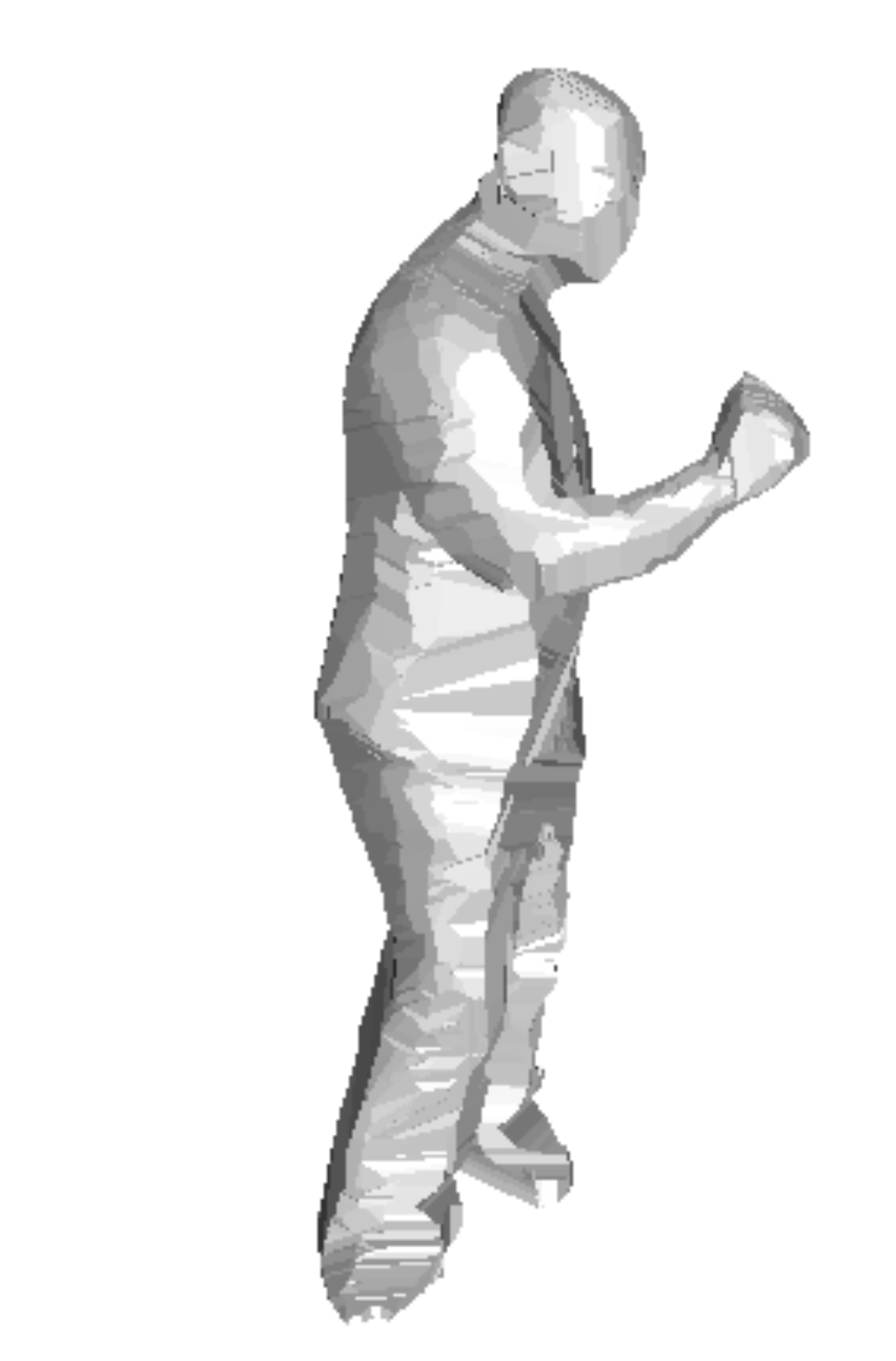} &
\includegraphicsBen{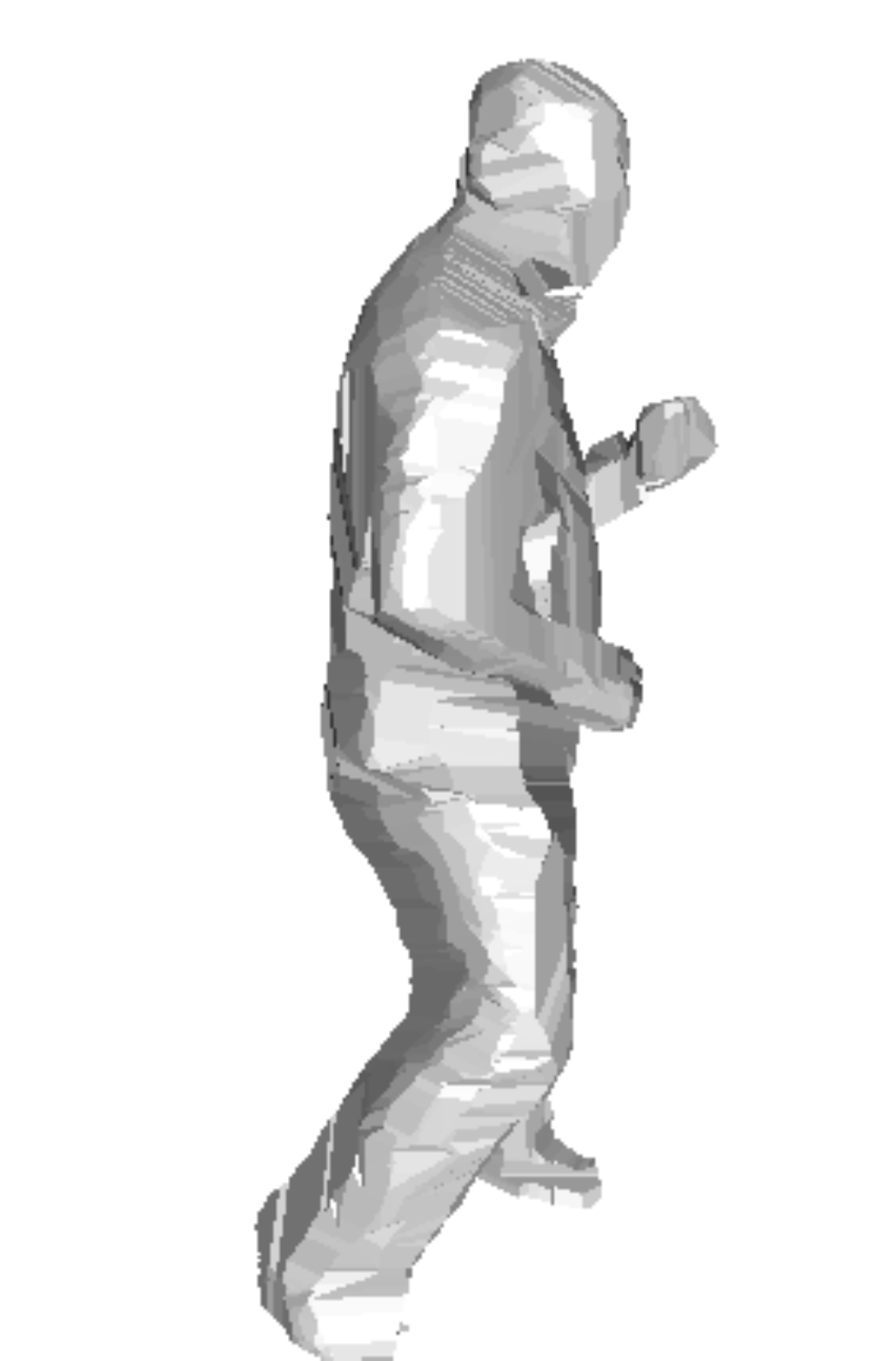} &
\includegraphicsBen{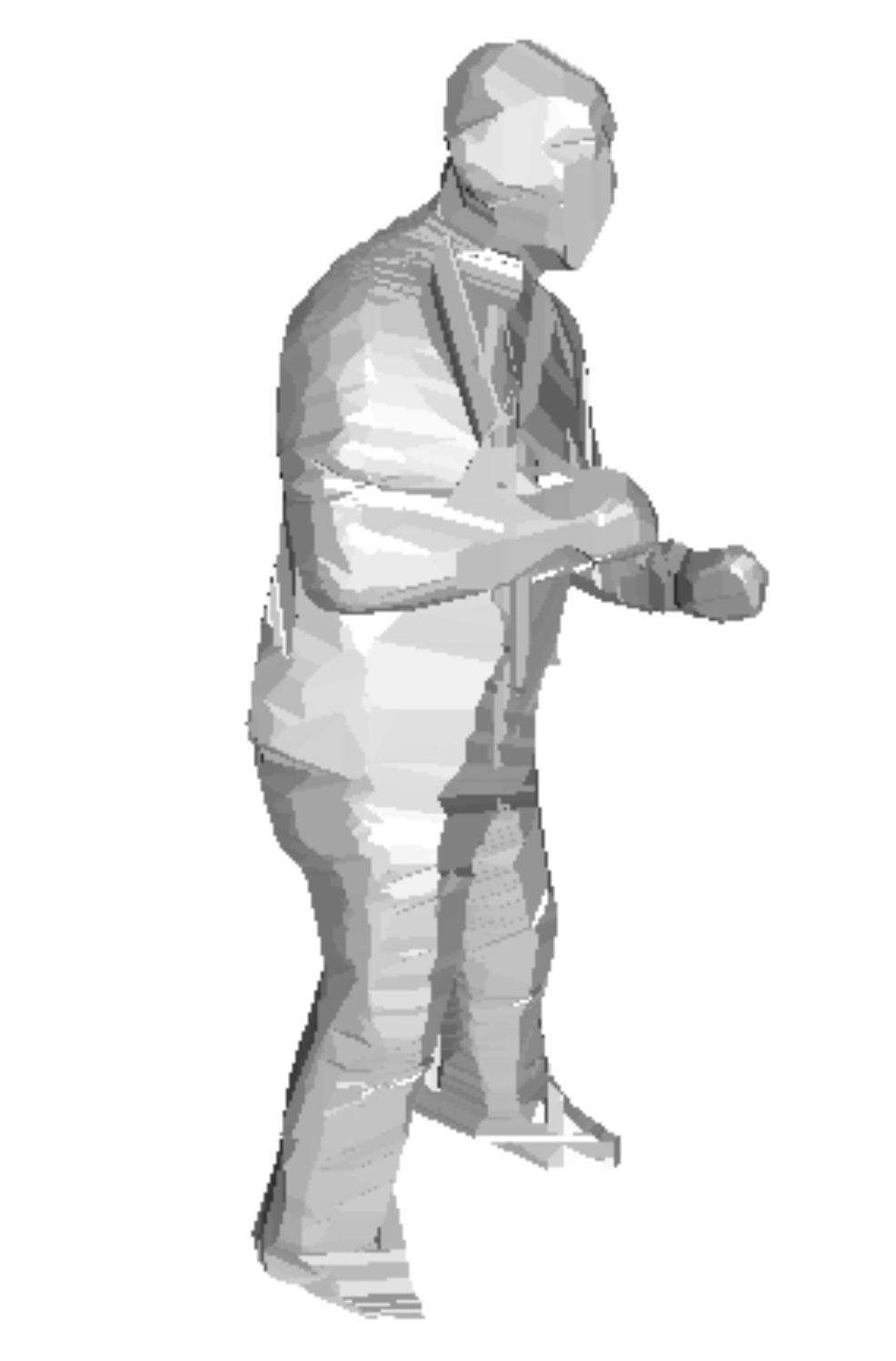} \\

\raisebox{3em}{\begin{sideways}  Final Meshes \end{sideways}} &
\includegraphicsBen{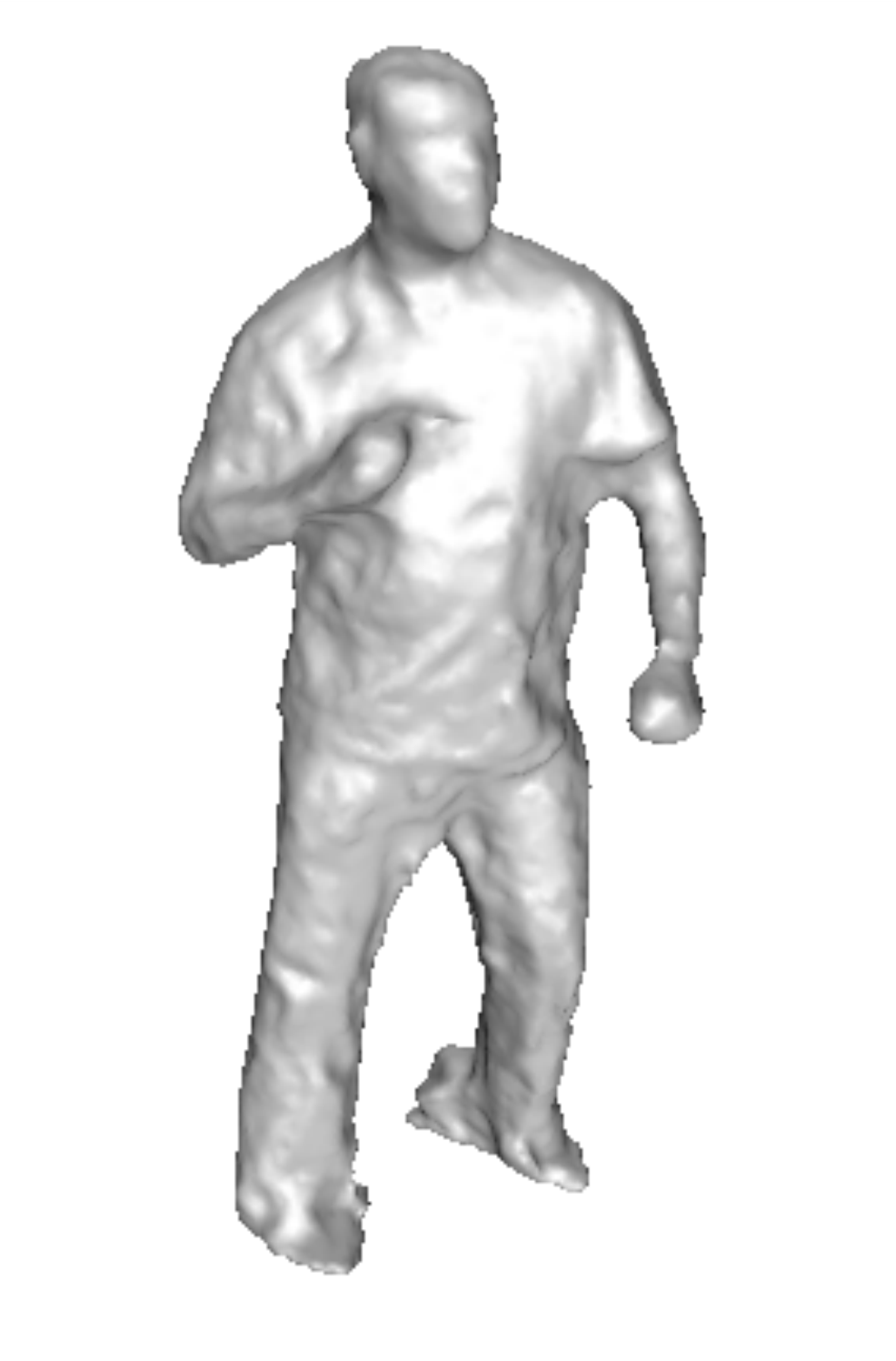} &
\includegraphicsBen{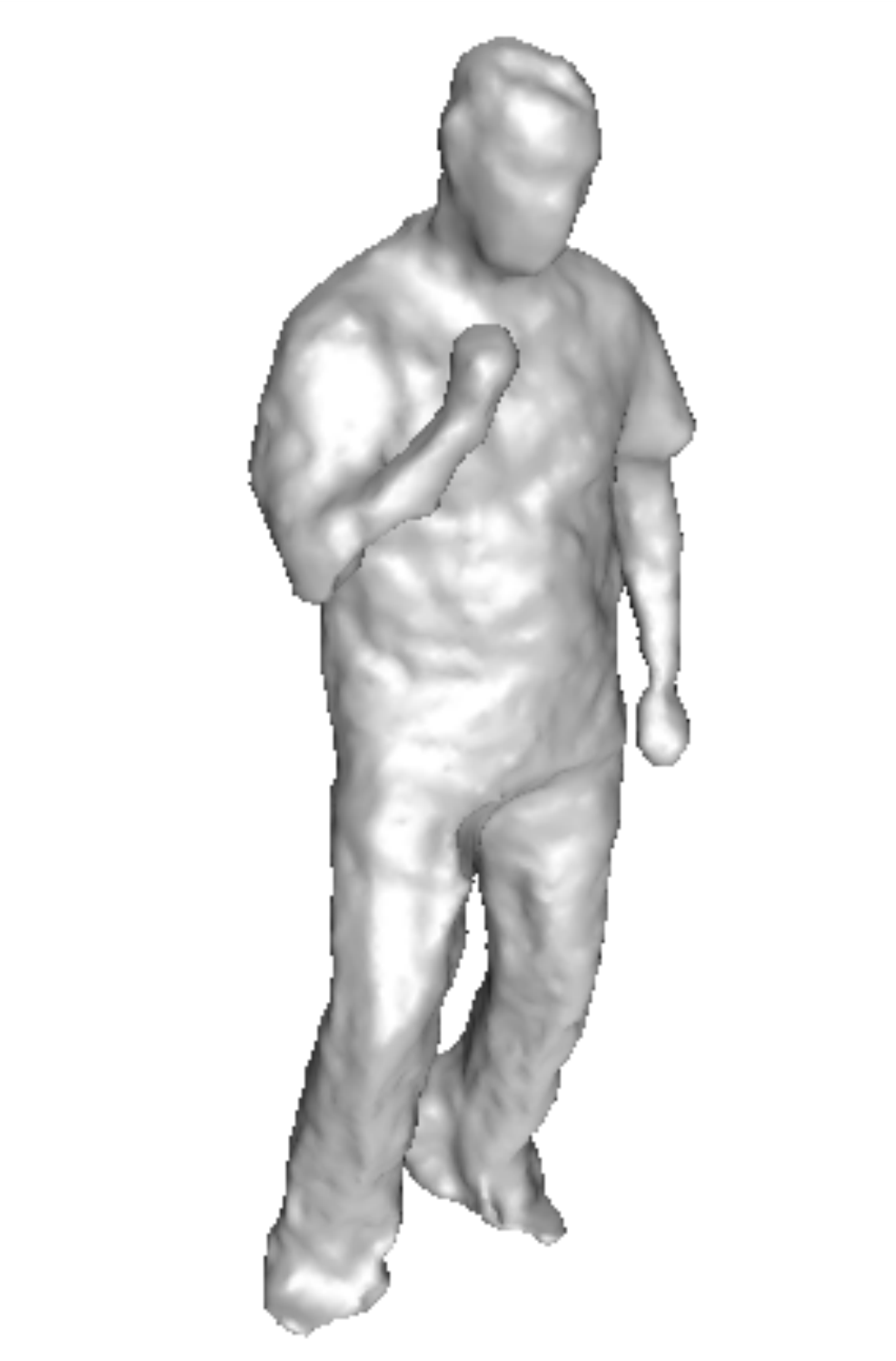} &
\includegraphicsBen{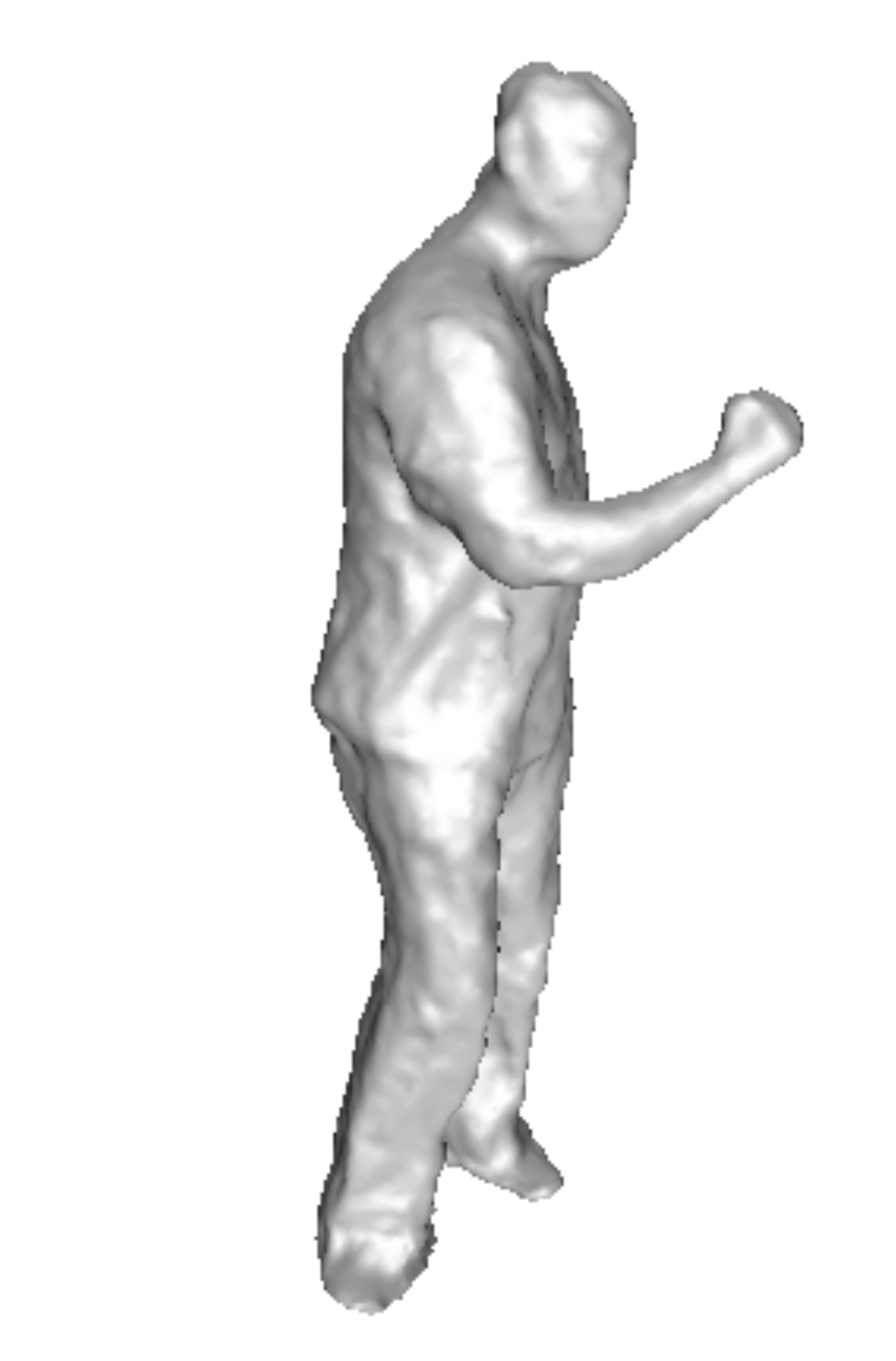} &
\includegraphicsBen{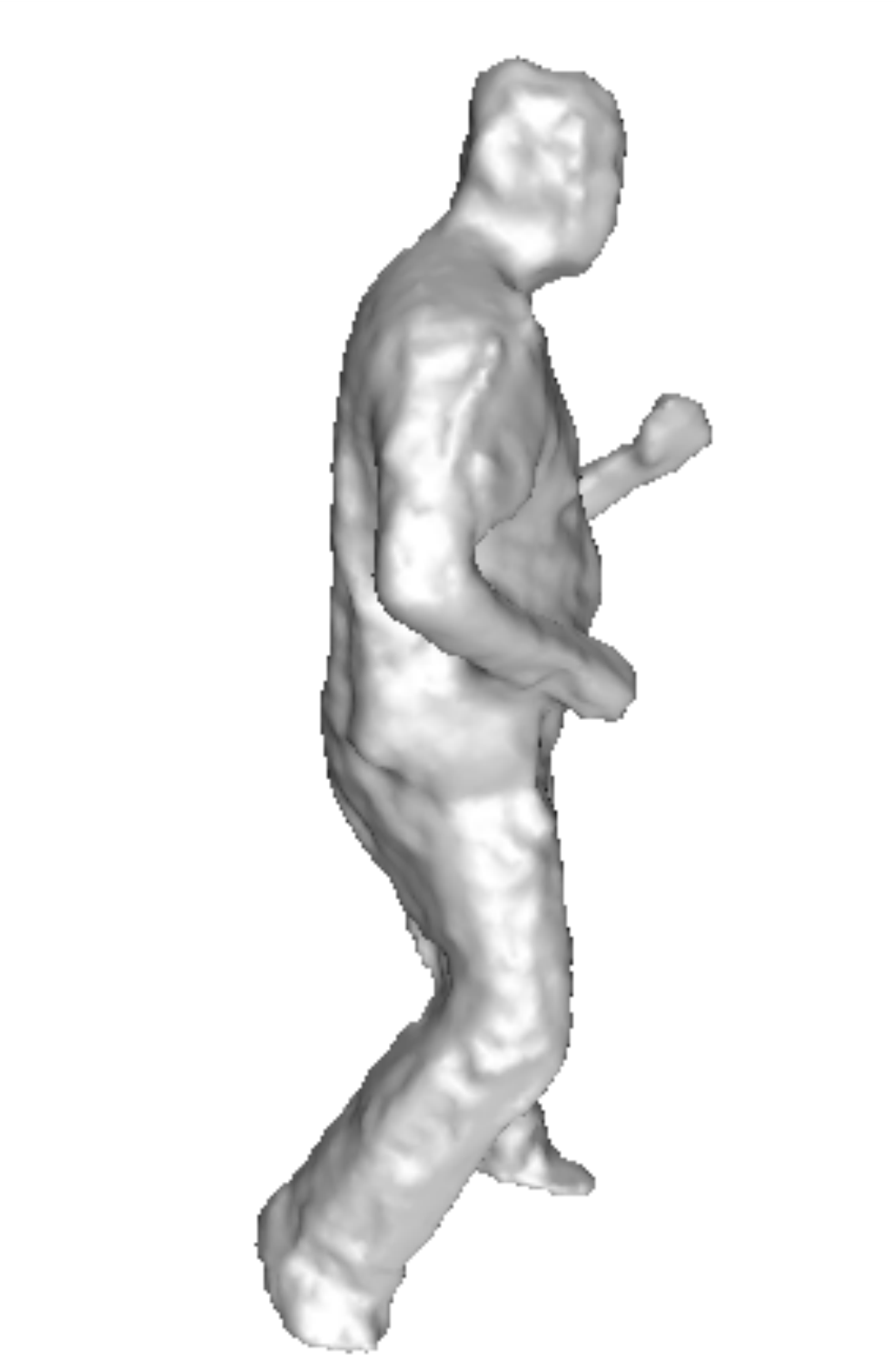} &
\includegraphicsBen{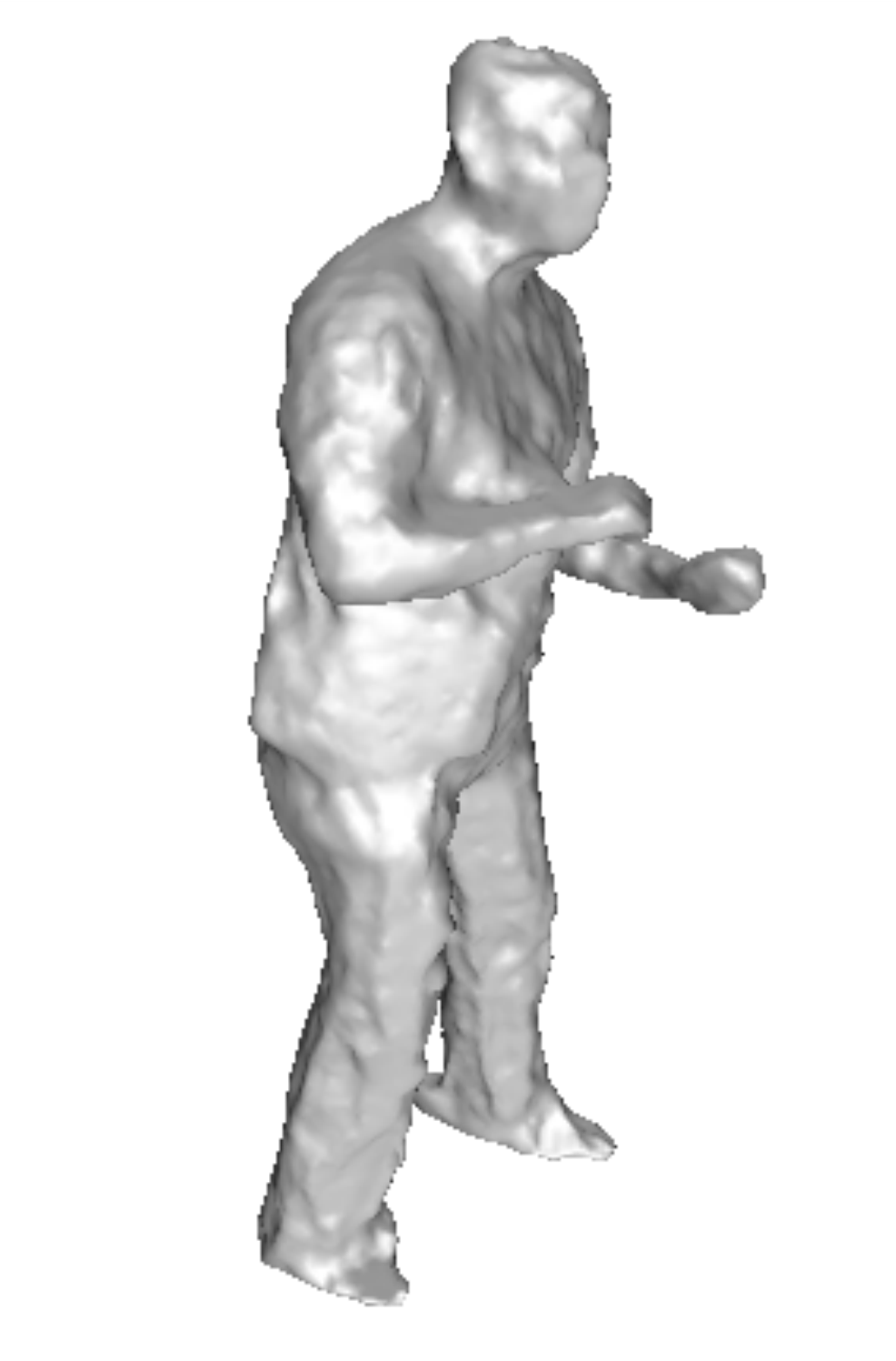} \\

\hline

\end{array}$
\end{center}
\caption{Results for the Man-dance sequence from the INRIA-PERCEPTION group. At each
  time-step the mesh is reconstructed from 34 cameras.}
\label{fig:results_ben}
\end{figure*}

\section{Conclusion}
\label{sec:conclusion}

In this paper, we proposed a geometry-driven self-intersection removal
algorithm for triangular meshes, able to handle topological changes in an
intuitive and efficient way.  We provided both a detailed description of the proposed algorithm, i.e.,  \MeshAlg{}, as well as an in-depth analysis of its convergence and performances (numerical stability and time complexity).

The \MeshAlg{} algorithm was plugged  into a generic mesh-evolution framework, thus allowing to address
two challenging problems within a topology-adaptive approach: surface morphing and multi-view image-based 3-D
reconstruction. Our main contribution with respect to the existing mesh-evolution methods is to provide a purely geometric mesh-based solution that is correct, that does not constrain meshes and that allows for facets of all sizes as well as for topological changes. In the case of surface morphing, we showed that \MeshAlg{} can deal with challenging topological
cases.

The 3-D reconstruction method that we described and which is based on
mesh evolution is extremely versatile. The method
recovers a correct discrete surface geometry starting from very coarse
approximations, such as visual hulls or sparse sets of 3-D point clouds. The
3-D reconstruction results are of comparable quality with
state-of-the-art methods recently developed by computer vision researchers.

\section*{Acknowledgments} 
We thank Jean-Philippe Pons and Renaud Keriven for providing the
source code for the gradient computation needed by the multi-view 3D
reconstruction algorithm. 
\bibliographystyle{IEEEtran}

\begin{IEEEbiography}[{\includegraphics[width=1in,
height=1.25in,clip,
keepaspectratio]{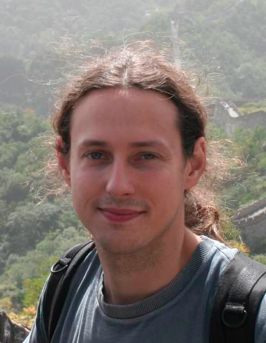}}]{Andrei Zaharescu} 
received the B.Sc. and M.Sc. degrees in computer science from York University, Toronto, Canada in 2002 and 2004, respectively. He obtained the Ph.D. degree in computer science from the Institut National Polytechnique de Grenoble, France in 2008. He is currently working in the industry in the area of computer vision, dealing with background subtraction and 2-D tracking methods. His research interests include camera calibration, background subtraction, sparse and dense 3-D reconstruction , geometric mesh processing, 2-D and 3-D tracking. He is a member of the IEEE and of the IEEE Computer Society.
\end{IEEEbiography}
\vspace*{-1cm}

\begin{IEEEbiography}[{\includegraphics[width=1in,
height=1.25in,clip,
keepaspectratio]{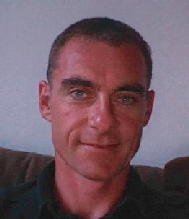}}]{Edmond Boyer} 
is associate professor at Grenoble universities (France). He obtained his PhD from the Institut National Polytechnique de Lorraine (France) in 1996. He started his professional career as a research assistant at the University of Cambridge (UK) in the Department of Engineering. Edmond Boyer joined the INRIA Grenoble in 1998. His fields of competence cover computer vision, computational geometry and virtual reality. He is co-founder of the 4D View Solution Company in the domain of spatio-temporal modeling. His current research interests are on 3D dynamic modeling from images and videos, motion capture and recognition from videos, and immersive and interactive environments.
\end{IEEEbiography}
\vspace*{-1cm}

\begin{IEEEbiography}[{\includegraphics[width=1in,
height=1.25in,clip,
keepaspectratio]{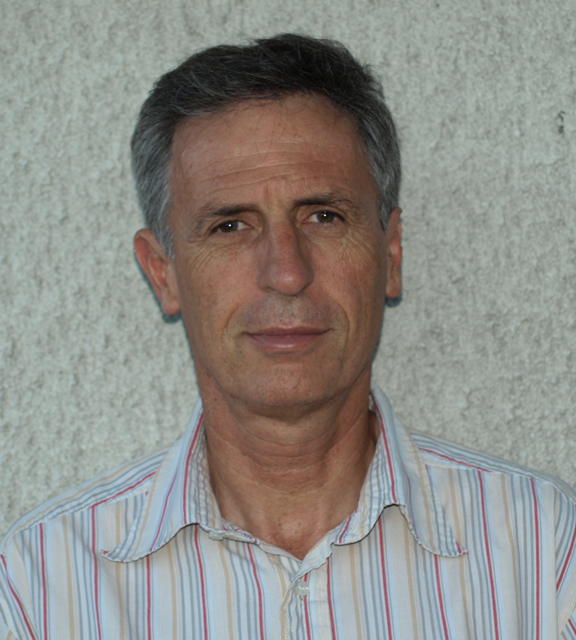}}]{Radu Horaud} 
received the B.Sc. degree in electrical engineering, the M.Sc. degree
in control engineering, and the Ph.D. degree in computer science from
the Institut National Polytechnique de Grenoble, Grenoble, France. 

He holds a position of Director of Research with the Institut National de Recherche en Informatique et
Automatique (INRIA), Grenoble Rh\^one-Alpes, Montbonnot, France, where
he is the head of the PERCEPTION team since 2006. His
research interests include computer vision, machine learning,
multisensory fusion, and robotics. He is an Area Editor of the
\textit{Elsevier Computer Vision and Image Understanding}, a member of
the advisory board of the \textit{Sage International Journal of Robotics
  Research}, and a member of the editorial board of the
\textit{Kluwer International Journal of Computer Vision}. He was a
Program Cochair of the Eighth IEEE International Conference on
Computer Vision (ICCV 2001).
\end{IEEEbiography}

\end{document}